%% file: 0main.tex
\documentclass[sigplan,10pt]{acmart}

\AtBeginDocument{%
  }

\copyrightyear{2024}
\acmYear{2024}
\setcopyright{acmlicensed}\acmConference[EuroSys '24]{Nineteenth European Conference on Computer Systems}{April 22--25, 2024}{Athens, Greece}
\acmBooktitle{Nineteenth European Conference on Computer Systems (EuroSys '24), April 22--25, 2024, Athens, Greece}
\acmDOI{10.1145/3627703.3629572}
\acmISBN{979-8-4007-0437-6/24/04}

\usepackage{tikz}
\usepackage{amsmath}

\usepackage{algorithm}
\usepackage{algpseudocode}
\usepackage{color,soul}
\usepackage{subfigure}
\usepackage[normalem]{ulem}
\usepackage{multirow}
\usepackage{makecell}
\usepackage{comment}
\usepackage{hyperref}

\usepackage{listings}
\definecolor{dkgreen}{rgb}{0,0.6,0}
\definecolor{gray}{rgb}{0.5,0.5,0.5}
\definecolor{mauve}{rgb}{0.58,0,0.82}
\usepackage{pifont}
\newcommand{\cmark}{\ding{51}}
\newcommand{\xmark}{\ding{55}}

\newcommand{\mypara}[1]{\noindent{\bf {#1}}~}

\newcommand{\ignore}[1]{}

\newcommand{\sysname}{CDMPP}
\newcommand{\problemName}{CDCM}
\input{notation.tex}

\begin{document}

\title[{\sysname}: A Device-Model Agnostic Framework for Latency Prediction ...]{{\sysname}: A Device-Model Agnostic Framework for Latency Prediction of Tensor Programs}

\author{Hanpeng Hu}
\authornote{Work done as an intern at ByteDance Inc.}
\authornote{Corresponding authors: \{hphu,jwsu\}@cs.hku.hk}
\affiliation{
    \institution{University of Hong Kong}
    \country{Hong Kong}
}
% \email{hphu@cs.hku.hk}
\author{Junwei Su}
\authornotemark[2]
\affiliation{
    \institution{University of Hong Kong}
    \country{Hong Kong}
}
\author{Juntao Zhao}
\affiliation{
    \institution{University of Hong Kong}
    \country{Hong Kong}
}
\author{Yanghua Peng}
\affiliation{
    \institution{ByteDance Inc.}
    \city{Seattle}
    \state{Washington}
    \country{USA}
}
\author{Yibo Zhu}
\affiliation{
    \institution{ByteDance Inc.}
    \city{Seattle}
    \state{Washington}
    \country{USA}
}
\author{Haibin Lin}
\affiliation{
    \institution{ByteDance Inc.}
    \city{Mountain View}
    \state{California}
    \country{USA}
}
\author{Chuan Wu}
\affiliation{
    \institution{University of Hong Kong}
    \country{Hong Kong}
}

\renewcommand{\shortauthors}{Hu et al.}

\begin{abstract}
  Deep Neural Networks (DNNs) have shown excellent performance in a wide range of machine learning %(ML) 
  applications. Knowing the latency of running a DNN model or tensor program on a specific device is useful in various tasks, such as DNN graph- or tensor-level optimization %tensor-level optimizations in Deep Learning (DL) compilers, 
  and device selection. 
  Considering the large space of DNN models and devices that impedes direct profiling of all combinations, recent efforts focus on building a predictor to model the performance of DNN models on different devices. However, none of the existing attempts %can efficiently build 
  have achieved a cost model that can accurately predict the performance of various tensor programs while supporting both training and inference accelerators.
  % existing work either fail to capture the internal structure of tensor programs or efficiently consumes highly structured features. Besides, there lack of efforts that can handle both cross-model and cross-device performance prediction. 
  %To tackle this problem, 
  We propose {\sysname}, an efficient tensor program latency prediction framework for both cross-model and cross-device prediction. We design an informative but efficient representation of tensor programs, called {\em compact ASTs}, and a pre-order-based positional encoding method, to capture the internal structure of tensor programs. 
  % Besides, we introduce a Transformer-based model to efficiently consume {\em compact ASTs} and carefully devise the training objective to handle the large cost range \jw{scale-invariant/scale-insensitive objective?}. 
  % To make the predictor learn from different domains
  We develop a domain-adaption-inspired method to learn domain-invariant representations and devise a KMeans-based sampling algorithm, for the predictor to learn from different domains (i.e., different DNN operators and devices). 
  Our extensive experiments on a diverse range of DNN models and devices demonstrate that {\sysname} significantly outperforms state-of-the-art baselines with $14.03\%$ and $10.85\%$ prediction error for cross-model and cross-device prediction, respectively, and one order of magnitude higher training efficiency. The implementation and the expanded dataset are available at \url{https://github.com/joapolarbear/cdmpp}.
\end{abstract}





\maketitle

\input{1introduction}

\input{2motivation}
\input{3overview}

\input{4design}
\input{7implementation}

\input{8evaluation}

\input{9related_work}

\section{Conclusion}
This paper presents {\sysname}, an efficient framework for accurately predicting the performance of tensor programs across different DNN models and devices. The main design highlights include 1) {\em Compact ASTs} as a concise feature format to capture internal structures of tensor programs; 2) a customized positional encoding method and a Transformer-based cost model that enables efficient learning of AST-based inputs; 3) a CMD-regularization boosted training objective to learn domain-invariant representations robust to various DNN models; 4) a KMeans-based sampling strategy for cross-device fine-tuning. Our extensive experimental results demonstrate that {\sysname} outperforms SOTA baselines in terms of prediction error and training throughput for both cross-model and cross-device performance prediction. 
% \cwu{mention we plan to open-source our system somewhere in the paper. HHP: mentioned at the end of the Abstract}

% the generalization to unseen networks or devices and achieves more efficient training.

% future work: Some Applications
% (TODO) Autotune, compare to cost model in AutoTVM/Ansor

\section*{Acknowledgement}
This work was supported by Hong Kong Innovation and Technology Commission’s Innovation and Technology Fund (Partnership Research Programme with ByteDance Limited, Award No. PRP/082/20FX), and grants from Hong Kong RGC under the contracts HKU 17208920 and C7004-22G (CRF). We would like to express our sincere gratitude to the paper reviewers and artifact evaluation reviewers for their valuable feedback and contributions to our work.

\bibliographystyle{plain}
\bibliography{0main}

\input{0appendix}
\clearpage
\input{appendix/EuroSys24_ArtifactAppendix_template}

\end{document}

%% file: notation.tex
\newcommand{\deviceSpace}[0]{\mathbb{D}}
\newcommand{\modelSpace}[0]{\mathbb{M}}
\newcommand{\randvar}[1]{\mathbf{#1}}

\newcommand{\expectation}[0]{\mathbb{E}}
\newcommand{\prob}[0]{\mathbb{P}}
\newcommand{\loss}[0]{\mathcal{L}}
\newcommand{\predictor}[0]{\mathcal{F}}

\newcommand{\dataSet}[0]{\mathcal{S}}
\newcommand{\train}[0]{\mathcal{S}^{train}}
\newcommand{\valid}[0]{\mathcal{S}^{valid}}
\newcommand{\test}[0]{\mathcal{S}^{test}}

\DeclareMathOperator*{\argmin}{arg\,min}

%% file: 1introduction.tex
%-------------------------------------------------------------------------------
\vspace{-2mm}
\section{Introduction}

The adoption of Deep Neural Networks (DNNs) in various applications has boosted the fast development of AI {\em hardware accelerators} %for AI, 
including GPUs (e.g., T4~\cite{nvidia2018T4}, V100~\cite{nvidia2017V100} and A100~\cite{nvidia2020A100}), TPUs~\cite{google2016TPU}, Huawei Ascend~\cite{huawei2018shengteng}, Habana Goya~\cite{habana2019goya}, and various IoT accelerators~\cite{lin2020mcunet, tang2017enabling, mohammadi2018deep}. %As the amount of data and model size grow, 
It is important to select proper devices~\cite{geoffrey2021habitat} and %apply 
DNN optimization techniques~\cite{hu2022dpro, jia2019optimizing, cai2018proxylessnas, chen2018tvm, dudziak2020brp, kaufman2019learned, kaufman2021learned, zhang2021nn_Meter} to accelerate DNN training or inference %on those devices.
under a specified time and cost budget.
% \jw{the relation between dnn optimization and development of hardware accelerator is not entirely clear}
 The execution latency of various DNN models or operators on different devices is essential for DNN optimization and device selection. For instance, to optimize a DNN model on different devices, Deep Learning (DL) compilers~\cite{chen2018tvm, baghdadi2021deep_tiramisu, adams2019learning} estimate or measure the performance of different tensor programs and select the best tensor programs for each computation subgraph on various devices. Another example is that automatic model-parallel training~\cite{jia2019beyond} requires querying the latency of each operator of a DNN on various devices when exploring ways to deploy %each operator
the DNN on a heterogeneous cluster.

Due to the large space of DNN models and devices and potentially %accessibility issue of devices,
limited access to certain devices, it may not be feasible to profile all DNN models on all devices~\cite{geoffrey2021habitat, liu2022nnlqp}. Many efforts have been devoted to developing cost models to estimate the performance of DNN models or operators. AutoTVM~\cite{chen2018tvm} and Ansor~\cite{zheng2020ansor} exploit XGBoost~\cite{chen2015xgboost} as a cost model to estimate the performance of tensor programs, which exploits an ensemble of decision trees and gradient boosting for supervised learning of the performance.
%cwu: this footnote should be moved somewhere else \footnote{TVM uses the term 'task' to represent the search task for a specific operator or a fused operator. One tensor program can be viewed as an implementation of an operator with a specific schedule. In this paper, we use the term ``task'' and ``operator'' interchangeably.}. 
The relative performance of a tensor program is predicted, i.e., the ratio of the processing throughput % achieved with the target schedule (i.e., a specific implementation of an operator) 
of the tensor program over the throughput of the tensor program with the smallest execution time, for the same computational subgraph in a given dataset. Similarly, TLP~\cite{zhai2023tlp} estimates the relative time of tensor programs (i.e., the speed-up over the original tensor program after some optimizations are applied)
by recursively aggregating loop and computation information of each tensor program using LSTM. The relative time may not be %lacks flexibility in adapting to 
sufficient in different use cases. Given a dataset and a cost model trained on it, if %the search space changes with
new tensor programs are introduced for each computational subgraph, the tensor program with the largest throughput for the subgraph may differ; this necessitates modifying the entire dataset to update the relative values and re-training the cost model using the entire updated dataset. Besides, it is not feasible to aggregate the relative time of subgraphs to estimate the end-to-end execution time of a DNN model.

% !!! Discuss the advantages of absolute time later
% In contrast, if we focus on the absolute time, we can simply fine-tune the cost model with the additional dataset. Besides, it's not feasible to aggregate the relative time of operators to estimate the end-to-end performance of a DNN model.

Some other studies (e.g., NNLQP~\cite{liu2022nnlqp}) represent the DNN model as a graph and exploit a graph neural network (GNN) to predict the performance of the graph. They cannot provide the latency of each specific operator, which is required by systems such as DL compilers. Besides, GNN-based approaches are relatively coarse-grained by taking a DNN (sub)graph as input, not flexible and efficient enough, %introduces redundant features even 
e.g., when two DNN models only differ on several operators. %\cwu{shiwei's work considered the following} In addition, these methods don't consider the cases where different operators are deployed on different devices, e.g., the auto-model-parallelism case we mentioned above. 
% neither to say network connections between different pairs of devices may vary.

% \Hu{Support various devices concurrently}

% ------------- Problem statement
%% --- 3. Use a performance predictor. analytical cost models or DNN-based methods. However, existing methods have some limitations
To our best knowledge, there exists no generic predictor that can accurately estimate the absolute latency of operators from various DNN models %while concurrently supporting 
on different devices. 
We consider different DNN models and devices as distinct domains and name the cross-domain learning
% We name the %studied 
problem as a \textit{Cross-Device and Cross-Model ({\problemName})} prediction problem.
The {\problemName} problem can be divided into two subproblems: (1) cross-model performance prediction (CMPP), that is, on a specific device, modeling the performance of tensor programs extracted from different DNN models and predicting the performance of {\bf unseen} tensor programs; (2) cross-device performance prediction (CDPP), predicting the performance of a tensor program on a target device, based on its performance knowledge on other devices. %(source device). 
%\Hu{Should we replace the term "cross-model" with "intra-device"? cwu: no}
It is challenging to develop an accurate and efficient performance model of tensor programs for CDCM.
% \Hu{explain what we refer to as domain here}

% Therefore, to efficiently acquire latency feedback when making DNN optimization and device selection decisions, it is essential to {\em build an accurate predictor to estimate the absolute performance of any DNN operator on any device}, abbreviated as. 

% !!! Explain what a tensor program is
% Since DNN models are often implemented with a set of {\em tensor programs}, which are often used in ML frameworks~\cite{abadi2016tensorflow, chen2018tvm} to efficiently express and manipulate large-scale muli-dimensional data (e.g., images), we mainly target to model the performance of tensor programs instead, such that the predictor can be used to model the performance of both each tensor program and the entire DNN model. 

% ------------- challenges
% It's empirically difficult to maintain a model for all ops \Hu{add related works}. . - Some recent works propose extracting tree-like features (or AST) from tensor programs,  e.g., Tiramisu, but we empirically found that it doesn't improve the prediction performance much. \Hu{add related works}
%It is challenging to develop an accurate and efficient performance model of tensor programs for CDCM. %over various devices for tensor programs from different DNN models~\cite{kaufman2019learned} in the following aspects.

%$\triangleright$ 
{\em First, how to efficiently exploit internal structure information of DNN models is the key.}
Previous studies have emphasized the importance of exploiting the internal structure of tensor programs for accurate performance modeling and proposed using Abstract Syntax Trees (ASTs) as representations of tensor programs to capture their internal structure~\cite{ryu2021metatune, baghdadi2021deep_tiramisu}. However, it is challenging to encode ASTs as inputs of DNNs due to the extremely irregular nature of ASTs. Simple solutions, like template-based padding~\cite{ryu2021metatune} and AST architecture clustering~\cite{baghdadi2021deep_tiramisu}, significantly decrease training efficiency, as they introduce significant data sparsity and small batch sizes, respectively.
It is essential to efficiently process ASTs when studying the {\problemName} problem due to the large dataset involved (we use Tenset~\cite{zheng2021tenset} with over 50 million samples, each of which is a record of a tensor program and its measured execution time on a specific device).

{\em Next, cross-model and cross-device distribution shifts are difficult to handle .}
Tensor programs from different DNN models and various devices can follow varying distributions of arithmetic features, memory access patterns and loop nesting, 
which makes it challenging to learn the universal correlations among tensor programs and their performance~\cite{kaufman2021learned}. Separately maintaining a cost model for each device or each operator type is not a scalable solution~\cite{geoffrey2021habitat}. Some prior studies~\cite{ryu2021metatune, zhao2022moses, liu2022nnlqp, zhai2023tlp} exploit transfer learning to adapt a cost model to a new device; they do not specify %provide a solution on 
how to effectively collect traces from the target device for fine-tuning. Due to the large cost of trace collection, it is essential to sample a small set of representative tensor programs that can make the cost model adjust faster to the target device, especially with limited time and monetary budgets.

% \Hu{For example, Habitat maintains a cost model for each kind of operator, so it's challenging to train a general cost model}
% \Hu{Fine-tuning without any guide to the profiling process, while profiling is time-consuming}
% $\triangleright$ {\em How to sample training data.}
% To get the training data, we will sample some schedules for each high-level op, then parse corresponding features and profile the run time. How to decide the number of sampled schedules for each op on each device?

% It requires collecting traces on the target device. However, this is not possible when the target device is not accessible.
% method fails to learn domain invariant representation that can be generalized to unseen DNN models and devices. 

% !!! comment the challenge, since it's not such important, maybe discuss this later in the design section
% $\triangleright$ {\em Handle dataset skewness.}
% In a dataset used to train a predictor for CDPP and CMPP, the latency scale of tensor programs is often diverse, ranging from microseconds to hundreds of milliseconds. Additionally, the dataset often has a skewed distribution, with a long tail in the large-latency direction. Current research on performance modeling doesn't offer a dedicated solution to this specific regression problem, leading to suboptimal prediction results for tensor programs with high latency.

We propose \sysname, an efficient framework to predict the absolute execution latency of tensor programs from different DNN models across various devices, including both training and inference accelerators. {\sysname} introduces a regular and training-friendly structure, namely {\em Compact ASTs}, to capture the internal structure of tensor programs for efficient processing. To address the distribution shift, {\sysname} learns domain-invariant representations of tensor programs
by explicitly minimizing the distribution discrepancy across DNN models and devices %\Hu{double check}
, and proposes a clustering-based sampling strategy to guide profiling on the target device. {\sysname} also utilizes a replayer to estimate end-to-end DNN performance in a bottom-up manner, with the estimated latency of each tensor program. In summary, we make the following contributions in this paper: 
% \cwu{more details in each contribution summary point}

% Secondly, we leveraged this more regular representation to design a positional encoding method based on pre-order traversal of "Compact ASTs" and incorporated a Transformer-based model architecture that can learn from large sequence datasets effectively. To make the predictor learn from various domains, we first analyze the distribution difference of different types of DNN models and devices

% based on Central Moment Discrepancy (CMD)~\cite{zellinger2019cmd} 
% and develop a domain-adaption-inspired method to learn domain-invariant representations by minimizing the distribution discrepancy across DNN models and devices. 

% Finally, to tackle the problem of a large-scale range and skewed distribution of latency, we apply Box-Cox Transformation to the dataset and propose a hybrid training objective that balances the trade-off between absolute and relative errors. \Hu{Repeated contributions}

% \Hu{to estimate the end-to-end performance, we A more flexible approach is a bottom-up method that uses operator-level cost models, which can estimate the latency of each operator and network latency separately.}

%\begin{itemize}
    %\item 
$\triangleright$ To exploit the internal structure of tensor programs efficiently, we introduce a concise yet training-friendly representation of tensor programs, namely {\em Compact ASTs}, and a pre-order-based position encoding method. {\em Compact ASTs} are regular with a small range of sequence lengths, which enables large-batch training without introducing data sparsity and any loss of loop information.

$\triangleright$ We perform theoretical and empirical analysis on domain differences presented in both CDPP and CMDD cases. Accordingly, we introduce a domain-shift-based regularization term into our training objective, to learn domain-invariant representations that are robust to various DNN models and devices. To avoid large trace collection overhead when adapting the cost model to a new device, we design a KMeans-based sampling algorithm to select representative samples on the target device for profiling. We also utilize a scale-insensitive training objective and the Box-Cox transformation to handle dataset skewness \cite{wang2023margin} that arises from  low-frequency large-cost samples.
% where the quantity of low-cost samples is significantly greater than that of high-cost samples.
    % \item To tackle the issue of skewed datasets, we develop a hybrid training objective that combines the advantages of absolute and relative errors and studies the effect on the learning performance when applying different transformations.

$\triangleright$ We expand the Tenset
dataset to include a broader range of devices, e.g., GPUs (e.g., A100, V100, P100) and inference accelerators (e.g., HL-100), and provide a larger dataset better suited for the {\problemName} problem. The expanded dataset is open-sourced to facilitate future research in this direction.

$\triangleright$ Our extensive experiments show that {\sysname} achieves $14.03\%$ and $10.85\%$ prediction error for cross-model and cross-device tensor program latency prediction, respectively. In addition, the training throughput 
of {\sysname} is one order of magnitude higher than the other DNN-based methods.

%\end{itemize}
% \Hu{(Optional)
%   3. Propose a scale-invariant loss function
% }

%% file: 2motivation.tex
\section{Background and Motivation}
% \Hu{Cut down this section since the motivation has been introduced in the introduction, see how previous EuroSys papers organize sections}
\subsection{Deep Learning Compilers}

DL compilers \cite{chen2018tvm, baghdadi2021deep_tiramisu, adams2019learning, lattner2020mlir} have recently emerged that optimize DNN models in both the graph level and the tensor level and %lowering 
convert DNN models written with different ML frameworks (e.g., TensorFlow \cite{abadi2016tensorflow}, PyTorch \cite{paszke2019pytorch}) to hardware code. %The predominant deep learning 
State-of-the-art DL compilers consist of three parts: 1) frontends that treat DL models as {\em computational graphs} and perform graph-level and tensor-level optimizations on computational graphs written in different high-level languages;
2) unified {\em Intermediate Representations} (IRs) to represent DNN models lowered from different frontends; 3) device-specific backends, each of which translates the IR to hardware code that can run on the specific device (aka code generation).

TVM~\cite{chen2018tvm} is a popular open-source DL compiler %has gained great popularity in both industry and academia. 
whose frontends decompose the high-level computational graph to a set of computational subgraphs after applying graph-level optimizations. Each subgraph corresponds to one or several operator(s) in the original DNN model. TVM performs tensor-level optimization by applying some {\em schedule primitives} on each subgraph to generate a tensor program in the form of TVM IR (TIR). With various combinations of schedule primitives, one subgraph can be converted to tens of thousands of tensor programs. 
% Fig.~\ref{xx} shows an example of a subgraph and corresponding tensor programs. \Hu{show a figure of tensor program examples}
% \cwu{give the reason and example. Hu: considering the current introduction, should we still present a concrete example of different tensor programs of the same operator?}. 
TVM's auto-tuning scheduler~\cite{zheng2020ansor} assigns a {\em task} for each subgraph to search for its optimal tensor program.
% \footnote{We use the term ``task'' and ``operator'' interchangeably in this paper. \cwu{then task and operator may not be equivalent as task may correspond to a subgraph. HHP: not strictly equivalent, will check the term used in the paper}} 
% {\em tasks} (\cwu{%the search task for 
% a specific operator or a fused operator?})
% Then, its auto-tuning framework searches for the optimal tensor-level {\em schedule} for each task and converts %lowers 
% it to tensor programs in the form of TVM IR (TIR). A tensor program is an implementation of an operator with a specific schedule and one task (operator) can correspond to multiple tensor programs, as xx 
To demonstrate the effectiveness of our proposed framework, we estimate the performance of tensor programs written as TIR on diverse devices, considering TVM's popularity in this community. The same idea can be applied to other DL compilers. %and frameworks.
% \Hu{introduce the relationship between task, operator and tensor programs}

%%%%%%%%%%%%%%%% Motivation %%%%%%%%%%%%%%%%
\begin{table*}[!t] \centering
    \begin{tabular}{c|c|c|c|c} 
        \hline
        Method & \makecell[c]{Absolute Time\\Prediction} & {Model Level Prediction} & {Op/kernel-level Prediction} & {Cross-%training-inference-
        device} \\
        \hline
        AutoTVM~\cite{chen2018tvm}(2018) & \xmark & \cmark & \cmark & \xmark \\
        Tiramisu~\cite{baghdadi2021deep_tiramisu}(2021) & \xmark & \xmark & \cmark  & \xmark \\
        Kaufman, et al.~\cite{kaufman2021learned} (2021) & \cmark & \cmark & \cmark & \xmark \\
        MetaTune~\cite{ryu2021metatune} (2021) & \cmark & CNNs only & Conv and MatMuls only & \xmark \\
        \hline
        Habitat~\cite{geoffrey2021habitat} (2021) & \cmark & \cmark & \cmark & GPUs only \\
        NNLQP~\cite{liu2022nnlqp}(2022) & \cmark & \cmark & \xmark &\cmark \\
        TLP~\cite{zhai2023tlp}(2023) & \xmark & \cmark & \cmark & \cmark \\
        \hline
        \sysname & \cmark & \cmark & \cmark & \cmark  \\
        \hline
    \end{tabular}
    \vspace{2mm}
    \caption{Prior research on DNN performance prediction. 
    % \Hu{Why not compared to NNLQP} --> it is model-level prediction
    }
    \label{table:related_work}
    % \vskip -0.15in
    % \vspace{-4mm}
\end{table*}

% \vspace{-3mm}
\subsection{Cross-Device Cross-Model (CDCM) Prediction%Why {\problemName}?
}
We aim to develop a generic cost model to estimate the absolute time of different tensor programs on diverse devices. %such that the cost model can be adapted to different use cases. In this section, we present why we study the CMPP and CDPP problems respectively.
% to accurately predict the performance of various DNN operators on various devices by showing three applications.

\mypara{Why cross-model?}
DNN optimizations, including Neural Architecture Search (NAS) \cite{pham2018efficient}, graph-level optimization (e.g., operator fusion \cite{jia2019optimizing, jia2019taso}) and tensor-level optimization (e.g., loop tiling \cite{chen2018tvm}), involve performance queries of operators from different DNN models. To handle behavior differences between different operators and distribution shifts between DNN models, one intuitive approach is to maintain a cost model for each kind of operator. Habitat \cite{geoffrey2021habitat} learns an MLP for each kind of operator that uses different GPU kernels in the source and target devices, including 2D Conv2D, LSTM, etc. This approach is not scalable due to the diversity of DNN operators, especially when the operators can be fused or partitioned. It will be very useful to design a unified cost model that can predict the performance of unseen tensor programs or DNN models. %with unseen tensor programs.

\mypara{Why cross-device?}
% \noindent\textbf{Choosing from various devices.} 
DL users often need to decide which hardware accelerators to use. For example, given a DNN model, a developer may have the following options to run model training or inference: 1) a private computer with desktop GPUs, e.g., 2080Ti\cite{nvidia20182080Ti}; 2) a cluster within an organization equipped with server-class GPUs (e.g., V100\cite{nvidia2017V100}, A100\cite{nvidia2020A100}); 3) CPUs (e.g., Intel Platinum\cite{intel2019platinum}, AMD EPYC\cite{amd2019epyc}) or inference accelerators such as Habana HL-100\cite{habana2019goya}.
% like GCP\cite{google2008gcp}, Amazon EC2\cite{amazon2006ec2}. 
These options vary vastly in performance and monetary cost. Estimating the performance of DNN models on specific devices before renting or purchasing them can significantly help users make better decisions to meet their latency and monetary budgets. It is hence critical to devise a predictor that can estimate DNN Performance on various devices, including both training and inference accelerators.
% Predicting the performance of the target device in advance can help developers to choose the proper device to meet the time budget while avoiding renting or purchasing an unnecessarily expensive device.

Another application where both CMPP and CDPP are required is to automatically search for the optimal model parallelism strategy, i.e., determining which operators to be deployed on which devices, especially when running a DNN model on a set of heterogeneous devices. 
The search algorithm needs to query the latency of an operator if the operator is unseen (i.e., CMPP) or the operator is deployed on an unseen device (i.e., CDPP). 

Table \ref{table:related_work} summarizes recent studies on DNN performance prediction grouped into two categories. The first group focuses on performance modeling on a specific device, i.e., CMPP. For instance, 
AutoTVM~\cite{chen2018tvm} utilizes XGBoost~\cite{chen2015xgboost} to predict the relative performances of tensor programs. Tiramisu~\cite{baghdadi2021deep_tiramisu} designs an LSTM-based %recursive and 
recursive cost model for performance speedup prediction on CPUs when code transformations are applied. Kaufman et al.~\cite{kaufman2021learned} exploit a GNN to predict the latency of a subgraph on Tensor Processing Units (TPUs). MetaTune~\cite{ryu2021metatune} introduces a graph template to generate uniform input features. 
The second group includes solutions specifically designed for CDPP. Habitat~\cite{geoffrey2021habitat} leverages the Roofline model~\cite{williams2009roofline} to scale the performance of an operator from one GPU to a different GPU but requires the operator to be implemented using the same kernels in the two GPUs. NNLQP~\cite{liu2022nnlqp} supports model-level latency queries on various hardware using a GNN-based feature extractor and device-specific regression layers. TLP~\cite{zhai2023tlp} suggests extracting features from schedule primitives for tensor programs to avoid heavy feature engineering but targets relative performance prediction. In summary, there exists no cost model that can predict the absolute time at the op- or tensor-program-level and supports both training and inference accelerators.

\begin{figure}[t]
\subfigure[Fused Convolution and ReLU kernel]{     
\centering
\includegraphics[width=0.45\textwidth, trim=0 0 0 0, clip]{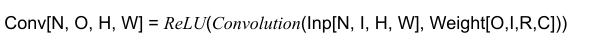}
\label{fig:feature.a}
}
\vspace{-3mm}
\subfigure[Tensor programs]{     
\centering
\includegraphics[width=0.35\textwidth, trim=0 0 0 0, clip]{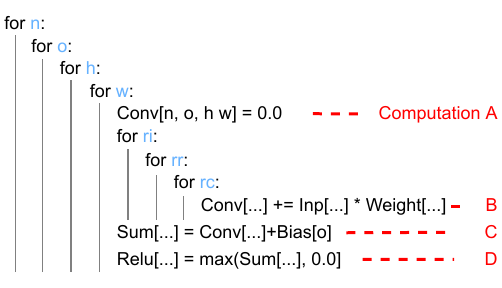}
\label{fig:feature.b}
}
\subfigure[Tiramisu's AST. ]{
\centering
\includegraphics[width=0.2\textwidth, trim=0 0 0 0, clip]{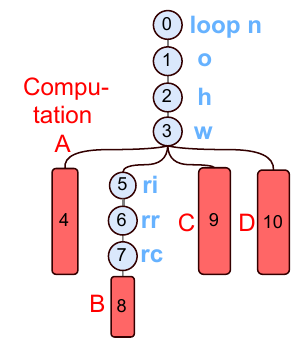}
\label{fig:feature.c}
}
\subfigure[Compact AST and Positional Encoding (PE)]{     
\centering
\includegraphics[width=0.225\textwidth, trim=0 0 0 0, clip]{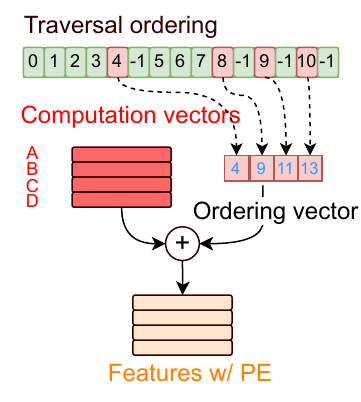}
\label{fig:feature.d}
}
% \includegraphics[width=0.9\textwidth]{fig/4design/feature.pdf}
% \centering
% \vspace{-4mm}
\caption{Example tensor program and AST. Numbers in black are used to identify nodes.%\small {\sysname} Feature Design.
}
\label{fig:feature}
% \vspace{-3mm}
\end{figure}

\begin{figure}[t]
\subfigure[]{     
\centering
\includegraphics[width=0.225\textwidth, trim=0 0 0 0, clip]{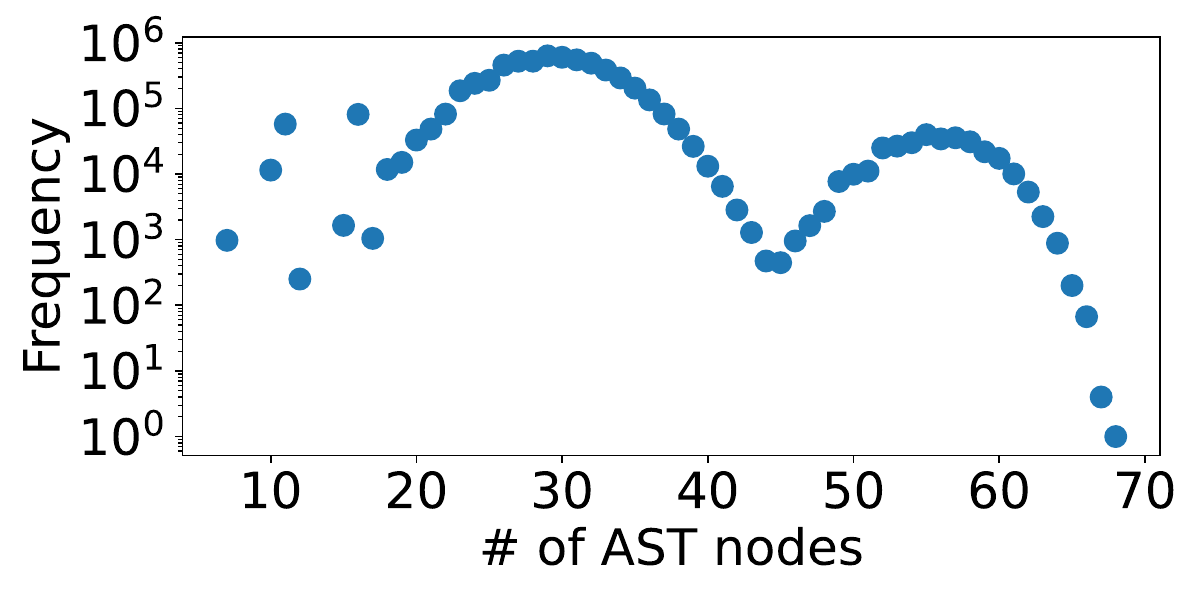}
\label{fig:ast_dist_a}
}
\subfigure[]{
\centering
\includegraphics[width=0.225\textwidth, trim=0 0 0 0, clip]{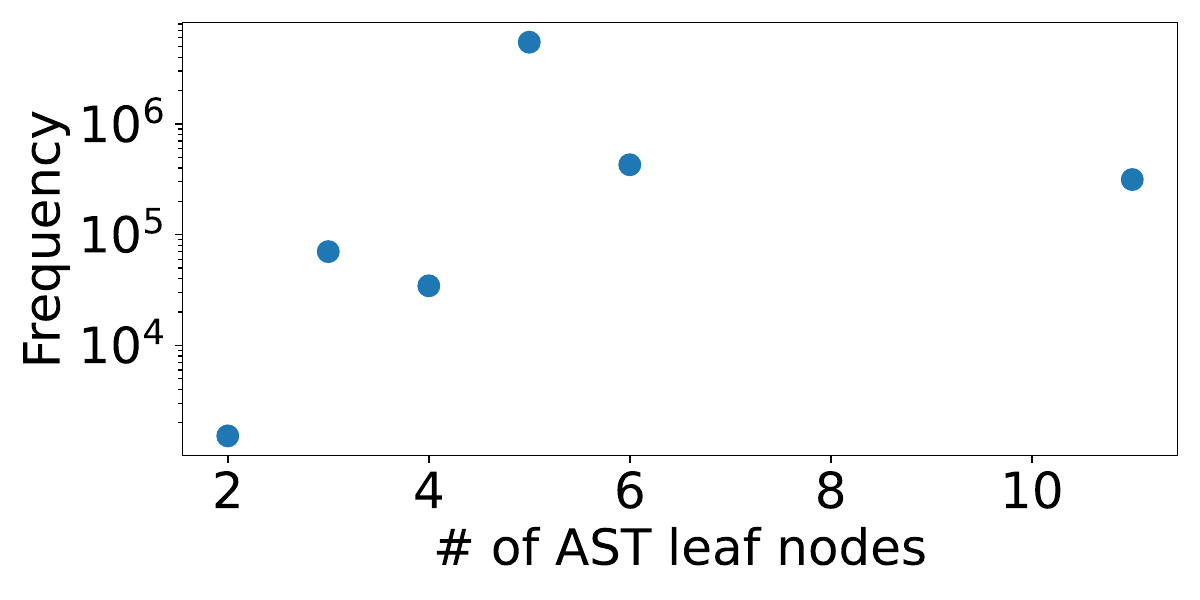}
\label{fig:ast_dist_b}
}
% \vspace{-4mm}
\caption{\small AST node number distribution in Tenset dataset: a) the distribution of node number in ASTs; b) distribution of leaf node number in ASTs.}
% from v100 to p100
\label{fig:ast_dist}
% \vspace{-5mm}
\end{figure}

\subsection{Challenge in AST Feature Extraction}
Selecting a proper representation of tensor programs is important for accurate performance modeling. Previous works \cite{baghdadi2021deep_tiramisu, ryu2021metatune} %has emphasized the importance to capture the internal structure of tensor programs for accurate performance modeling and 
have represented tensor programs as Abstract Syntax Trees (ASTs). Fig.~\ref{fig:feature} shows an example of the tensor programs of a fused Convolution and ReLU kernel (Fig.~\ref{fig:feature.b}) and its AST (Fig.~\ref{fig:feature.c}) constructed with Tiramisu \cite{baghdadi2021deep_tiramisu}. %represents a fused Convolution and ReLU kernel with AST. The original pseudocode code and tensor programs are shown in Fig.~\ref{fig:feature.a} and Fig.~\ref{fig:feature.b} respectively. 
%Each node in the AST represents a for-loop or a computation/memory expression.
There are two types of nodes in the proposed AST format in Tiramisu: 1) leaf nodes, where computation and memory access happen; 2) non-leaf nodes without node features, representing loop variables. Treating each loop variable as a node in the AST results in extremely irregular AST structures (i.e., the number of nodes and their positions vary significantly across ASTs of different tensor programs), due to the complex and diverse structures of tensor programs. Fig.~\ref{fig:ast_dist_a} plots the distribution of node numbers in ASTs in the Tenset dataset and the range of node numbers is large.
%Naive AST-based representation often leads to inefficient learning of the execution time of tensor programs due to the irregular structure of Naive ASTs, i.e., the number of nodes and their locations in Naive ASTs vary significantly across different tensor programs. 

% Do not discuss the location difference of AST nodes

% To selectively incorporate information from each AST node based on its positional relations with the other nodes 
To exploit the internal structure of ASTs, Tiramisu~\cite{baghdadi2021deep_tiramisu} proposes an LSTM-based recurrent and recursive network (RNN) to iteratively aggregate an AST by traversing all nodes in the AST. However, this traversal %-based forward 
process is highly dependent on the AST structure and only inputs with the same AST architecture can be batched together for learning. Due to irregular AST structures, batch sizes used in the training process in Tiramisu are small, resulting in low computation resource utilization and extremely low training efficiency~\cite{kosaian2021boosting}.
To extract uniform AST-based features for execution time prediction of convolutional neural networks (CNNs) at the kernel level, %in a uniform structure, 
MetaTune \cite{ryu2021metatune} augments each extracted AST to fit into a uniform super-graph template dedicated to convolution kernels. However, it is difficult to extend the template to other types of kernels due to the complexity and diversity of different kernels. Besides, given a large number of irregular AST structures, aligning small ASTs to a large template introduces significant sparsity into features, %and cost models, which 
preventing efficient and accurate learning of the cost models \cite{petrini2022learning}.

\noindent\textbf{Opportunity.} Fig.~\ref{fig:ast_dist_b} shows that the range of leaf node number in the ASTs is much more limited. We are inspired by this observation to design a regular feature structure by keeping only leaf nodes and incorporating loop information (e.g., the nesting level, and loop variable range) into the computation vector extracted for each leaf node. As a result, the AST is converted into a sequence of computation vectors of similar length. To ensure that the locations of leaf nodes in the AST are not lost, we record them and encode them into the sequence.
% by recording the memory access pattern, arithmetic features, and loop information (e.g., loop nesting level) for each leaf node and saving leaf nodes' locations in the AST.
% by binding the 
% \cwu{explain what these are (are they related to TIR of the leaf node?): loop information (e.g., loop tiling), memory access pattern, and computation statistical information} to each leaf node and recording their \cwu{leave nodes'?} locations in the AST \cwu{clarify how this can lead to a regular feature structure}.

\subsection{Challenge in Cross-Domain Prediction}
% \cwu{clarify what we are referring to as `domain'. HHP: defined in the introduction section, the same place where we define the CDCM problem}.
The distribution of tensor programs in different DNN models varies due to the difference in the types of operators used. %depending on the DNN models they are derived from. 
For instance, CNNs \cite{he2016resnet} typically have a higher proportion of convolutional operators, while RNNs tend to be dominated by recurrent units and LSTM operators. Some previous methods~\cite{geoffrey2021habitat, kaufman2021learned} maintain a unified cross-model by assigning a unique $op\_id$ to each type of operator and combining the $op\_id$ with operator-specific parameters to generate features. However, the $op\_id$s fail to reflect the correlation between different operators, making the predictor fail to generalize to new operators.  
% Operators also vary in the number of parameters 
Instead, we expect a cost model that learns a common representation that can be generalized across various DNN models with different operators. 
% \Hu{check}
% \cwu{clarify what challenge in cost model building this distribution shift brings.}

In addition, %due to %the architectural
%hardware difference between different devices, performance varies substantially when 
running the same set of tensor programs on different devices often leads to vastly different performances. The performance difference is hard to be estimated only using simple hardware parameters such as peak FLOPS and memory bandwidth, even when %both source and target 
the devices are from the same hardware vendor (e.g., different models of NVIDIA GPUs) \cite{geoffrey2021habitat}. To address the %large distribution 
performance shift between devices, previous works \cite{ryu2021metatune, zhao2022moses, liu2022nnlqp, zhai2023tlp} have suggested fine-tuning the cost model on the target device. However, they do not discuss which tensor programs to profile on the target device for cost model fine-tuning, which is very relevant when % In cases like device renting and purchasing, 
the target device is not always available, and profiling the whole DNN model is very time-consuming. For example, profiling all tensor programs in the Tenset dataset on one specific device takes days and even weeks. It is important to select tensor programs that can better represent the entire dataset to profile, for achieving better fine-tuning results with limited resources.
% \Hu{Show distribution shifts}

\noindent\textbf{Opportunity.}
TIRs provide a common representation for tensor programs of different computational subgraphs, enabling us to %extract features from TIRs to 
learn the correlations among tensor programs from different subgraphs.
%common representation among different subgraphs.
For cross-device learning, assuming that the same set of tensor programs are run on different devices, we can leverage the extracted features in the source devices to identify tensor programs that can best represent the entire set of tensor programs. 

%% file: 3overview.tex
\section{{\sysname} Overview}
\label{sec:overview}

\begin{figure}[t]
\includegraphics[width=0.48\textwidth]{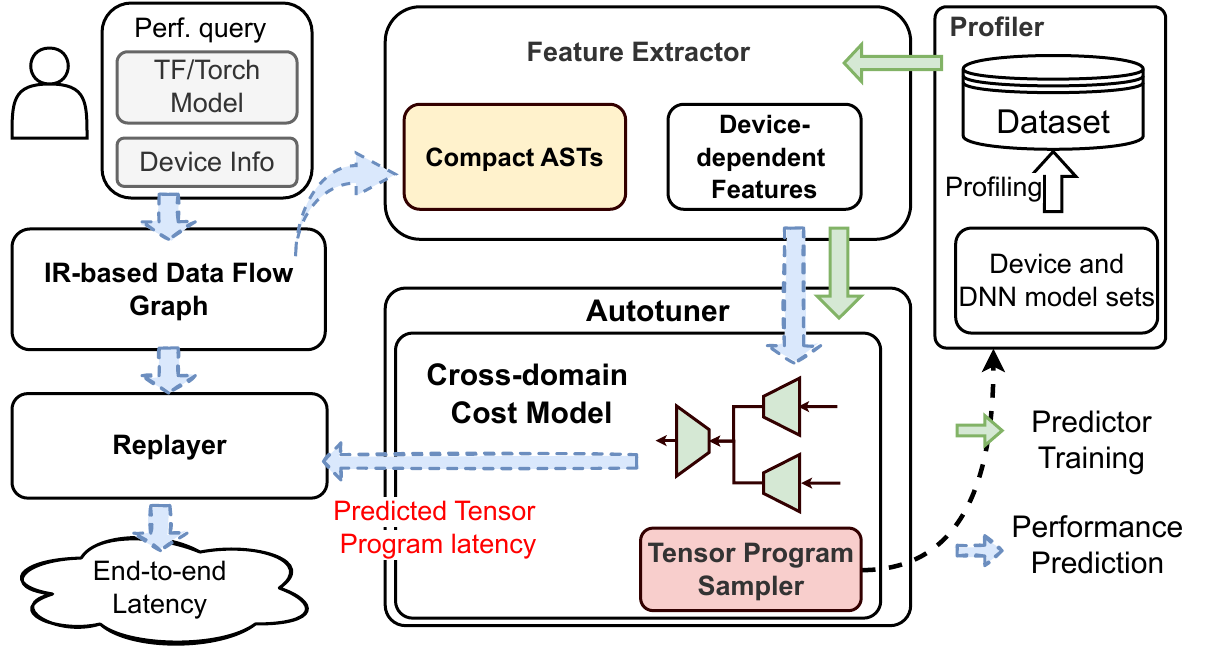}
\centering
\vspace{-6mm}
\caption{\small{{\sysname} Overview 
% \Hu{add fine-tuning; highlight compact AST}
% \cwu{update the figure to be consistent with terms in Sec. 3.2; change Perf. to Performance}
% \Hu{Tensor program Sampler --> Tensor Program Sampler; black margin for the compact AST block}
}}
\vspace{-4mm}
\label{fig:overview}
\end{figure}

\subsection{Problem Formulation} \label{sec:problem_formulation}
We aim to build a predictor for precise performance prediction of tensor programs from different DNN models on various devices, i.e., solving the {\problemName} problem. 
% \cwu{(point to the subsection defining CDCM. HHP: we define the CDCM problem in the introduction section, should we point to Section 1?)}
Formally, the goal of the predictor is to predict the %performance (
latency $\randvar{y}$ of a tensor program using the TIR presentation $\randvar{x}$ %that corresponds to
of the tensor program. 

% Cross-model
Intuitively, different DNN models $M$ would induce different %distributions on the 
sets of TIR presentations of tensor programs due to the different types of operators
%neural architectures 
they use. 
% For example, in a CNN, the main operations used in tensor programs are convolution and pooling. In contrast, an RNN is mainly composed of recurrence and activation operations.
% vs RNN... \jw{add more details on why different models would have different distributions, maybe in terms of operators}.
%This means 
We can view the %input 
TIR representation $\randvar{x}$ from model $M$ as generated from a distribution $\prob_M$, dependent on the model $M$.
% Cross-device
%Different device $D$ would influence the model performance (regression value) for the same tensor program~\cite{geoffrey2021habitat}.
% \jw{citation for works that use linear/scaling model for cross device}. 
The performance of a tensor program would condition on the device $D$ it runs~\cite{geoffrey2021habitat}, and we use $\prob(\randvar{y}|\randvar{x}, D)$ to denote the % label distribution for $\randvar{x}$ with a given device $D$.
% \cwu{clarify what the label is}.
execution time distribution for $\randvar{x}$ with a given device $D$.
Our CDCM prediction problem with model $M$ and device $D$ is to learn a cost model to predict the performance of any tensor program from model $M$ on device $D$, with input $x$ generated from distribution $\prob_{M}(\randvar{x},\randvar{y}|D) = \prob_{M}(\randvar{x}) \prob(\randvar{y}|\randvar{x},D)$.

Let $\deviceSpace$ denote the set of all devices and $\modelSpace$ be the set of all DNN models.
We use $\predictor(.)$ to represent our cost model and $\loss(·, ·)$ as a loss function to measure the difference between the predicted execution time and the ground truth $\randvar{y}$. The {\problemName} tensor program performance prediction problem can be formulated as follows:
\vspace{-1mm}
\begin{equation}\label{eq:cdcm_problem}
    \min_{\predictor} \max_{M \in \modelSpace, D \in \deviceSpace} \expectation_{(\randvar{x},\randvar{y}) \sim \prob_{M}(\randvar{x},\randvar{y}|D)}[\loss(\predictor(\randvar{x}),\randvar{y})|D].
\end{equation}
\vspace{-2mm}

In practice, we do not have thorough access to data from all possible % distributions, $\prob_{M}(\randvar{x},\randvar{y}|D), $.
% \cwu{$\forall$  what? HHP: $\forall M \in \modelSpace, D \in \deviceSpace$ make}. 
models $\modelSpace$ and devices $\deviceSpace$.
We rely on a finite dataset $\dataSet$ sampled from some finite mixtures of the distributions $\prob_{M}(\randvar{x},\randvar{y}|D)$ to support prediction modeling. 
For instance, we study the {\problemName} problem on the Tenset~\cite{zheng2021tenset} dataset, with records extracted from 120 ML models on 2 GPU models and 4 CPU models. We also supplement the dataset with 3 more GPU models and an inference accelerator. %learning. In addition, 
$\dataSet$ is partitioned into $\train$, $\valid$ and $\test$, which as the prediction model training set, validation set, and test set, respectively. 
% \Hu{CDPP and CMPP uses different data partitions}

% \subsection{Theoretical Justification}
% \subsection{Cross-Domain Learning}
We use a DNN as our prediction model. It is difficult to learn an accurate prediction model % as above 
as the observations in the training data cannot cover data from all possible devices and DNN models. We are inspired by the recent research on learning an invariant representation function~\cite{stojanov2021domain}, that allows us to effectively encode information from different devices and models into a common latent space.
% that xx \cwu{clarify how this invariant representation function is related to the problem (1)}.
Representative DNN models can often be interpreted as an encoder-decoder structure, where the encoder ($h$) %tries to 
extract useful information from the input data into a latent representation and the decoder ($f$) maps the latent to a prediction. %Let $h$ denote the encoder associated and $f$ denote the decoder.
We thus have $\predictor = f \circ h$, where $f$ is a regression function. Domain invariant learning is to learn an encoder that gives rise to an equal conditional distribution, i.e., the statistical heterogeneity from different devices and models is mitigated in the latent representation space:
% \cwu{explain the following equation using plain text}:
\begin{equation} \label{eq:equal_cond_dist} \begin{aligned} 
\prob(\randvar{y}|h(\randvar{x}),D,M) \approx \prob(\randvar{y}|h(\randvar{x}),D',M'), \\ 
\forall D, D' \in \deviceSpace, \forall M, M' \in \modelSpace.
\end{aligned} \end{equation}

\noindent Such a representation $\randvar{z}=h(\randvar{x})$ will bring a bounded generalization risk and improve the prediction results (as detailed in Sec.~\ref{sec:cmd_regular}).

\subsection{{\sysname} Architecture}
% \Hu{Describe the training workflow. Pre-training + fine-tuning?}
We propose \sysname, a system to tackle the {\problemName} problem, that efficiently learns a predictor to accurately estimate the latency of tensor programs across various DNN models and devices.
% without the need for direct access to the target device. 
Fig.~\ref{fig:overview} illustrates the architecture of \sysname. %In the following sections, we will introduce the components of {\sysname} by describing the workflow for both CMPP and CDPP at both tensor program and DNN levels.

\noindent\textbf{Feature Extractor}
%The first component of {\sysname} is a Feature Extractor, which 
extracts device-independent features $x$ in the form of {\em compact ASTs} from tensor programs, to capture the internal structure of the programs. A customized positional encoding method is used to aggregate the ordering vector and computation vectors of each compact AST.
% \Hu{clarify why we need device-independent features}. 
The Feature Extractor also extracts device-dependent features $\upsilon$, e.g., clock frequency, peak FLOPS, memory bandwidth, cache size, etc., which are used for cross-device learning. 

% The predictor is a composition $\hat{y} = \mathcal{F}_l \circ \mathcal{F}_e(p, d)$ of a {\em Feature Extractor} $x=\mathcal{F}_e(p, d)$ and a {\em Cross Domain Learner} $\hat{y} = \mathcal{F}_l(x)$.

% , consisting of two phases: 1) the training phase, where the cost model is trained using some benchmark DNN models and devices; 2) the inference phase, where users send a query about the performance of a DNN model on a target device and the system makes a prediction of the end-to-end performance based on the trained cost model.

\noindent\textbf{Cross Domain Learner}
%The {\em Cross Domain Learner}  
maintains the performance predictor $\mathcal{F}(\cdot)$. %to evaluate the performance of tensor programs.
The predictor contains a Transformer-based encoder and a linear decoder. The encoder also contains multiple linear embedding layers for different compact AST architectures, each of which is responsible for generating embeddings $z_x$ for ASTs with the same specific number of leaf nodes.
The learner is pre-trained on the training set $\train$. 
% \Hu{check} %with a training objective customized for the skewed dataset. 
To learn from different domains (aka different DNN models and devices), we incorporate a CMD-based regularization term into the training objective, such that the distribution difference among the latent representations (i.e., the output of the encoder) of tensor programs from different DNNs and devices is minimized. Here, CMD (Central Moment Discrepancy~\cite{long2016cmd}) is a metric for distribution difference.
For cross-device learning, we train the predictor with $\train$ and fine-tune it with sampled features from the target device.
% To learn from different domains (aka different DNN models and devices), \cwu{explain what the different domains refer to}),  we incorporate a CMD-based (\cwu{define CMD}) regularization term into the training objective and %fine-tune 
% train the predictor with $\train$ and sampled features from the target domain \cwu{clarify what the target domain means}, 
% such that the distribution difference among the latent representations (i.e., the output of the encoder) of tensor programs from different DNNs and devices is minimized.
% %Finally, 
The learner also contains an auto-tuner that performs an automatic search for optimal hyper-parameters and neural architecture for the predictor. 

\noindent\textbf{End-to-end Performance Prediction.}
% \noindent\textbf{DFG handler.}
To evaluate the end-to-end execution time of a DNN model, %the {\em DFG handler} first dissects
the DNN model is dissected into a set of tensor programs, and a tensor-program-based Data Flow Graph (DFG) is constructed (each node in the DFG represents a tensor program and the edges describe dependencies between tensor programs), 
by the {\em DFG handler}.
For each tensor program, the Feature Extractor parses its features and queries the predictor to obtain its execution time. A {\em replayer} then simulates the execution of the DNN model on a specific device: %in a similar manner as presented in
it decides the execution order and timestamps of each tensor program using a topological sorting algorithm% as in
~\cite{hu2022dpro},
based on the tensor-program-based DFG and predicted time of each tensor program, thus obtaining the end-to-end execution time. % \Hu{introduce how the replayer works in the appendix}

% \Hu{Delete Profiling Scheduler}
% In this paper, we take TVM IR (TIR)~\cite{chen2018tvm} as the inputs of our cost model, considering its popularity in the community and the same architecture can also be applied to other compilers or frameworks. 

%In the following sections, we will delve into the design of each component in detail. 
We next detail the design of compact AST and custom positional encoding method 
% \cwu{this method is not mentioned in the above `Feature Extractor' paragraph and should be echoed there} 
in Sec.~\ref{sec:feature_design}, describe the predictor architecture and cross-domain learning design %, and present a convergence analysis 
in Sec.~\ref{sec:cross_domain_leaner}. %The results of the evaluation will be presented in Sec.~\ref{sec:evaluation} followed by the conclusion.

% This project aims to predict DNN performance after being optimized by some deep learning compilers~\cite{chen2018tvm, lattner2020mlir, leary2017xla}, which supports both cross-device performance prediction and cross-model performance prediction. This paper takes TVM and Tensor IR (TIR) as an example of the deep learning compiler and its IR, considering their popularity in this community. 

% During the training phase, \Hu{the Profiling Scheduler automatically selects DNN models and devices to profile the time cost of IR kernels on those devices.} The profiling results, along with the features generated by the Feature Extractor, are fed into the Cross-domain Cost Model for training. \Hu{The validation results are sent back to the Profiling Scheduler and guided to profile proper DNN models and devices}. For example, if we make a prediction on different devices and the validation results show that the cost model still performs poorly on some devices, the Profiling Scheduler will profile more data related to those devices.

% During the inference phase, as receiving a performance query from users, the Profiler dissects the DNN model as a set of IR kernels. For each IR kernel, the Feature Extractor parses its features as the input and queries the cost model to get its time cost. The DFG Handler first parses IR-kernel-based DFG from the target DNN model and labels each node with its corresponding IR kernel, then uses the predicted time cost of IR kernels to label each node in the DFG. We use a replayer~\cite{dPRO} to estimate the end-to-end performance of the result DFG.

%% file: 4design.tex
\section{Feature Extraction} \label{sec:feature_design}

\subsection{Compact AST} 
%cwu: it is the third time the following was said: Some prior work~\cite{baghdadi2021deep_tiramisu, ryu2021metatune} has discussed the advantages of highly-structured data, e.g., AST, to represent tensor programs, that is, it can capture the internal structure information of tensor programs and alleviate the burden of heavy feature engineering. 
%Fig.~\ref{fig:feature.c} shows an example of how Tiramisu represents a fused Convolution and ReLU kernel with AST. The original pseudocode code and tensor programs are shown in Fig.~\ref{fig:feature.a} and Fig.~\ref{fig:feature.b} respectively. The proposed AST format in Tiramisu contains two types of nodes: 1) leaf nodes, indicating where computation and memory access happen; 2) non-leaf nodes without node features, representing loop variables. This approach of treating each loop variable as a node in the AST results in an extremely irregular AST structure due to the complex and varied structure of a large number of tensor programs. As shown in the evaluation (see Sec.~\ref{fig:tir-cm-throughput}), we observe that Tiramisu, which performs batching according to the structure of irregular ASTs, is extremely inefficient to learn from large datasets.

Feature engineering is a vital step in tensor program latency modeling. %Before introducing our feature design,
% that both captures the internal structure and achieves efficient learning, 
Device-independent factors that affect tensor program performance can be divided into three categories: 1) computation expressions, corresponding to leaf nodes in ASTs, which describe the detailed computation type and memory pattern; 2) loop information related to each computation expression, including the number of loops, lengths of loops (i.e., iteration range of loop variables) and each loop's property (e.g., whether a loop is vectorized, unrolled or parallelized); 3) the location of each computation expression in the tensor program, which may affect memory locality. Our goal is to design an AST-based feature format that encapsulates all necessary information in a compact and regular structure so that our predictor can efficiently consume features without any loss of useful information. %that may affect the performance of tensor programs.

%To this end, we carefully design an alternative representation for tensor programs, namely 
We carefully design a {\em Compact AST} to represent each tensor program. Based on the AST of a tensor program (e.g., as built with Tiramisu \cite{baghdadi2021deep_tiramisu}), we extract a {\em computation vector} with the same set of features for each leaf node of an AST as in \footnote{Ansor: \url{https://arxiv.org/abs/2006.06762}}, which consists of computation and memory access, loop information related to each computation expression, i.e., the first two categories of device-independent factors summarized above.
% \jw{maybe provide a brief argument why the feature extracted by Ansor should be sufficient?}
%Instead of keeping the entire AST, {\em Compact ASTs} only 
As illustrated in Fig.~\ref{fig:feature.d}, we also serialize the original AST based on 
% \cwu{still not clear how the pre-order traversal is carried out. HHP: the pre-order traversal is quite standard, just traversal the root first, then its left child node, and finally its right left child node. What to notice is that we use a special traversal ordering method by adding a marker to the traversal ordering vector for each leaf node, such that the ordering vector can be de-serialized to the original tree. I have plot the corresponding traversal ordering vector in Fig.1(d)} 
the pre-order traversal with a special maker, e.g. -1 in our case, appended after each leaf node, to capture locations of computation expressions (leaf nodes) and loop nesting information. We record the index of each leaf node in the traversal ordering and generate the ordering vector used for positional encoding.
% \cwu{in a uniformly-sized vector? HHP: the length of the ordering vector = # of leaf nodes in the AST. The range of # of leaf nodes is limited.}.
%The resulted ordering vector can be de-serialized to the original AST,  {\em Compact ASTs} successfully keep the loop nesting information. 
%in addition to memory access patterns and the number of various arithmetic instructions, {\em Compact ASTs} encode the loop information related to each computation expression into the computation vector of the corresponding leaf node. 
The computation vectors of leaf nodes and the ordering vector constitute a Compact AST.

Compared to original ASTs, the Compact AST design reduces the feature size while retaining all %three aforementioned
device-independent factors that may affect tensor program performance. %Additionally, this makes the features more compact and regular. 
As the range of leaf node number is small among tensor programs (Fig.~\ref{fig:ast_dist_b}), our feature size is more regular. %This observation guides us to 
We design a %clustering-based 
predictor learning framework (Sec.~\ref{sec:model_arch}) based on the leaf node number, to effectively handle the variable-length feature inputs without much overhead.

% Given an IR kernel and target device, the Feature Extractor parses a fixed length of features for IR kernels from different domains, which is composed of \textit{device features} and \textit{IR features}. 

% We collect the following metrics from the target device as device features: 1) clock frequency; 2) memory capacity; 3) FLOPS ...\Hu{add more device-related metirs}. 

% The main challenge is to collect fixed length of IR features for IR kernels from different op types since they have a different number of loop levels, loop steps, and primitive operations (e.g., add, multiply, etc.)~\cite{chen2018learning}. To achieve this, we use a sparse matrix to represent the features of one IR kernel, where each row corresponds to one primitive operation, and each column denotes the number of loop levels, loop steps, parallel steps, and unrolled steps of each primitive operation. 
% Then the sparse matrix is flattened and concatenated with device features to get the final fixed length of features for this IR kernel.

% \vspace{-2mm}
\subsection{Positional Encoding}
We encode the computation vectors of leaf nodes and the serialized AST to generate our feature for each tensor program
(Fig.~\ref{fig:feature.d}) using positional encoding (PE).
Positional encoding has been commonly used in Natural Language Processing (NLP) and Computer Vision (CV) to represent the position of an element in a sequence in the input to a neural network~\cite{attention2017, devlin2018bert, dai2019transformer}. %To enhance the learning with Compact ASTs, 
We consider the sequence of leaf nodes (expressions) in a Compact AST and %encode the location of each node in the Compact AST into the feature vector. We 
utilize the ordering vector in the Compact AST to calculate positional encoding for each leaf node, such that its position in the original AST is encoded by a unique representation. Specifically, let $N_{entry}$ be the length of the computation vector of each leaf node. %suppose we have a sequence of leaf nodes, each with a feature vector of length $N_{entry}$, 
Let $\mathcal{V}$ be the ordering vector, the position encoding of the $\xi$-th leaf node 
% of order $k$, i.e., the index in the pre-order traversal of the original AST,
% \cwu{index $k$ in the pre-order traversal of the original AST ?} 
is computed as

% \vspace{-3mm}
$$position(\xi,2\delta)=\sin{(\frac{\mathcal{V}[\xi]}{\Theta^{2\delta/N_{entry}}})}$$
% \vspace{-3mm}
$$position(\xi,2\delta+1)=\cos{(\frac{\mathcal{V}[\xi]}{\Theta^{2\delta/N_{entry}}})}$$
% \vspace{-3mm}

\noindent where $\delta \in [0, N_{entry}/2]$ denotes the entry id 
in the output positional embedding, %$k$ is given by pre-order traversal of the original AST 
and $\Theta$ is a user-defined scalar that affects how fast frequencies are decreasing along the vector dimension and is usually set to 10000 \cite{attention2017}.
% refer \url{https://kazemnejad.com/blog/transformer_architecture_positional_encoding/}
% Set to 10,000 by the authors of

% \vspace{-2mm}
\subsection{Device-dependent Features}
The Feature Extractor extracts not only %the features that represent the structure and operation of a tensor program (
Compact ASTs but also device-dependent features. These features are related to hardware %properties and characteristics
specifications of each device such as clock frequency, memory bandwidth, computation cores, the peak number of floating-point operations per second (FLOPS) in different precisions, L1/L2 cache size,  memory size, etc. They are used to model how a tensor program would perform on a specific device, %and are necessary to perform
for cross-device performance prediction.

% \subsection{Advantages of Compact AST and Decoupled Feature Design}
% Reduce the feature size, grouping based on leaf node number, maybe we should show some observations like AST with different leaf node numbers have different behaviors

% ------------------- Cross-domain learner
\section{Cross-Domain Learning} \label{sec:cross_domain_leaner}

\subsection{Predictor Model Architecture} \label{sec:model_arch}
We adopt a Transformer-based architecture to encode feature inputs to our predictor.
% Some prior work~\cite{baghdadi2021deep_tiramisu, steiner2021value} utilizes LSTM~\cite{yu2019review} to learn from sequence-based inputs. 
Recent research~\cite{attention2017} has proven that Transformer outperforms LSTMs ~\cite{baghdadi2021deep_tiramisu, steiner2021value} in tasks %Natural Language Processing (NLP) on 
processing sequence inputs. %Another advantage of the 
In addition, a Transformer architecture %compared to LSTM is that Transformer 
can be parallelized for more efficient training. 
% \Hu{Discussion: why not to use GNN} \jw{one possible argument: GNN layer assumes permutation-invariant property of the data which results in loss of ordering information from the subgraph}
We do not use a GNN as GNN layers assume permutation-invariant property of the input data, which may result in loss of ordering information in the subgraphs~\cite{meltzer2019pinet}.

\begin{figure}[t]
\centering
\includegraphics[width=0.4\textwidth, trim=0 0 0 0, clip]{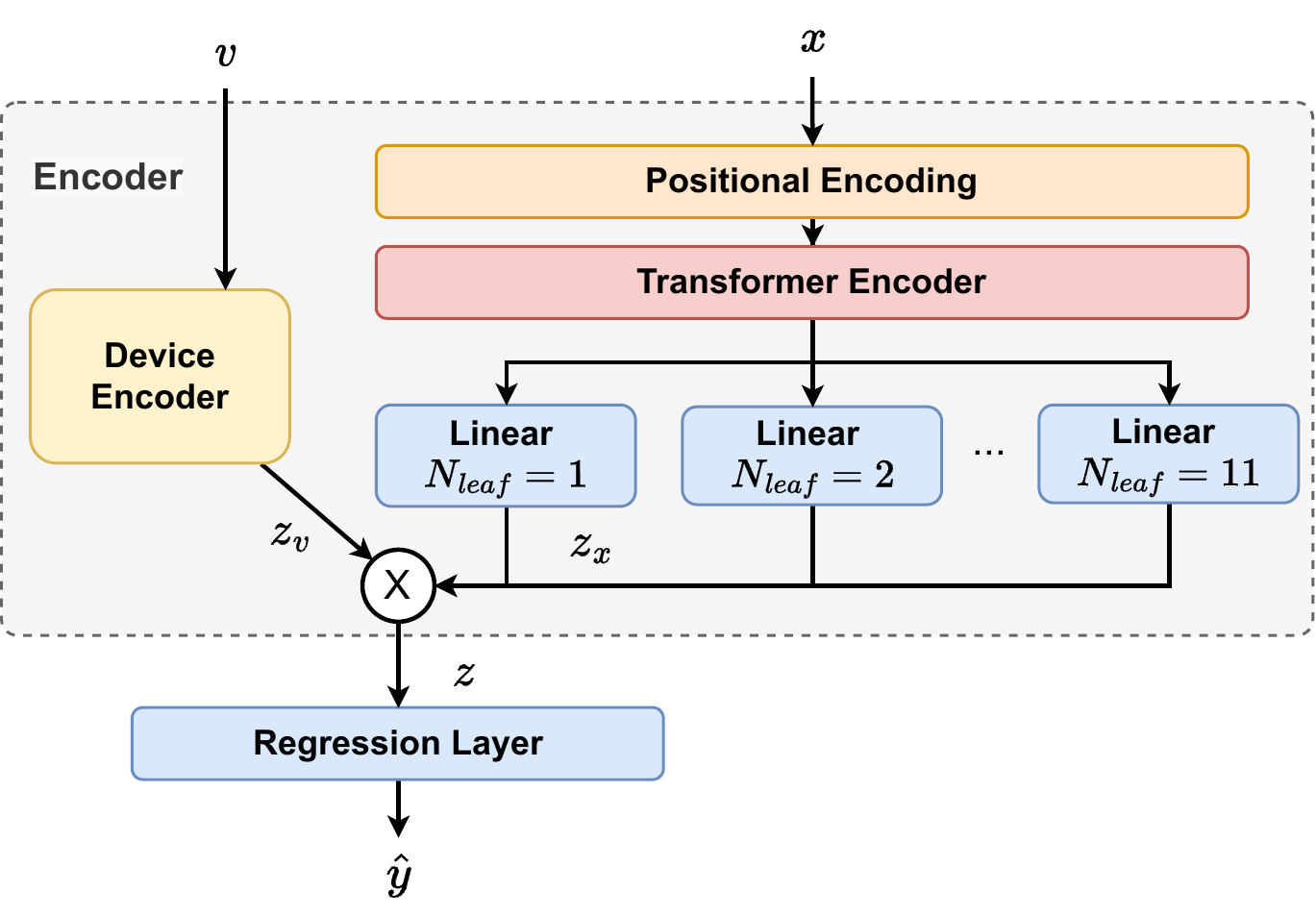}
\vspace{-3mm}
\caption{%Cost 
Predictor Model Architecture}   
\label{fig:cm_architecture}
\vspace{-5mm}
\end{figure}

%Therefore, we design a Transformer-based architecture 
As shown in Fig.~\ref{fig:cm_architecture}, we feed the device-independent Compact AST $x$ to position encoder, then a Transformer Encoder and a leaf-node-number-specific embedding layer to generate embedding $z_{x}$ of a fixed sequence length. We also feed device-dependent features $\upsilon$ to an MLP %-based side
network to compute embedding %of device-dependent features $\upsilon$, namely 
$z_{\upsilon}$. The device-dependent embedding $z_{\upsilon}$ is further aggregated with device-independent embedding $z_{x}$ to produce embedding $z= z_{x} \bigotimes z_{\upsilon}$, which is then fed into the %clustering-based
decoder to generate the time prediction $\hat{y}$.
% an MLP-based regression layer to generate the estimated time $\hat{y}$.
% https://medium.com/saarthi-ai/transformers-attention-based-seq2seq-machine-translation-a28940aaa4fe

%\mypara{Clustering-based Learning.} 
%Considering the Compact ASTs of different tensor programs can have different numbers of leaf nodes, we need to address 
To address the variable-length input due to different leaf node numbers in the Compact ASTs,  %Benefiting from the regular structure of Compact ASTs, we further partition the dataset according to the number of leaf nodes in ASTs. As Fig.~\ref{fig:cm_architecture} shows, 
we use different linear layers to process outputs of the Transformer encoder according to the respective %values of $N_{leaf}$ 
leaf node number ($N_{leaf}$), i.e., outputs corresponding to Compact ASTs of the same leaf node number are processed by the same linear layer. The linear layers produce embedding $z_x$ of the same length. Compared to the common padding approach that adds zero paddings to force uniform input sequences~\cite{ryu2021metatune}, this leaf node number-based method guarantees %that each partition has 
uniform input shapes to the decoder without introducing additional sparsity and computation, leading to more efficient predictor model learning. Given the limited range of leaf node numbers, the memory overhead of keeping multiple linear layers is limited.

% The decoder maintains an MLP-based regression layer for ?? \cwu{explain more about the decoder - it is said in Sec. 3.2 that there are `multiple decoders for different compact AST architectures'. HHP: previous description about the decoder in Sec. 3.2 is incorrect. There is only one decoder. The N-leaf-specific linear layers have been introduced in the previous paragraph. Have corrected Sec. 3.2 accordingly.} 

% \Hu{Discuss how we use fine-tuning to train cost models with different leaf node no}
% \Hu{Why not represent 1 leaf node compact AST by “sparsified” 11 leaf node compact AST? As discussed in https://www.analyticsvidhya.com/blog/2022/10/dealing-with-sparse-datasets-in-machine-learning/, sparse datasets will introduce a series of issues. We indeed observed worse training efficiency and larger prediction errors with sparse datasets. Although the clustering-based method indeed introduces some memory overhead, it’s limited to the maximum number of leaf nodes.}

% \vspace{-2mm}
\subsection{Pre-training with Scale-insensitive Training Objective}
% \jw{scale-invariant/ scale-insensitive objective?}. 
A machine learning-based cost model is often trained to minimize the
mean square error (MSE)~\cite{ryu2021metatune}. However, when the range of %tensor program latency 
prediction values (i.e., tensor program latencies in our case) differs substantially, MSE often leads to a model whose predictions are about the mean of the performance distribution, underestimating high-latencies %tensor programs 
and over-estimating low-latencies. Mean Absolute Percentage Error (MAPE) measures the relative error (i.e., the average absolute percentage difference between predicted values and actual values). 
When minimizing MAPE as the training objective, overestimation poses a risk of significantly large MAPE error ($>> 1$), whereas underestimation ensures that the MAPE error remains $\leq 1$. In datasets with significant skewness, where a large portion of samples have small values, the cost model tends to produce small predicted values to prevent overestimation for samples with small actual values and achieve a low MAPE error. However, this results in large absolute errors for samples with large actual values. 
To balance between the absolute and relative errors, we use a scale-insensitive hybrid training objective, which minimizes MSE and MAPE concurrently. Specifically, the loss function adopted in our prediction model pre-training (on the training set $\train$) is as follows: 

% \vspace{-3mm}
\begin{equation} \begin{aligned} 
\loss_{pre\_train} = & \frac{1}{|\train|} \times \left( \sum_{i \in \train} (\hat{\randvar{y}}_{i} - \randvar{y}_{i})^2 + \right.\\
& \left. \lambda  \sum_{i \in \train} |\hat{\randvar{y}}_{i} - \randvar{y}_{i}|/\randvar{y}_{i} \right)
\end{aligned} \end{equation}
where $\lambda$ is a coefficient to ensure the same order of magnitude of MSE and MAPE terms. We empirically find
$\lambda=10^{-3}$ performs well in our experiments.

% \vspace{-2mm}
\subsection{Fine-tuning with CMD-based Regularization}\label{sec:cmd_regular}

For better prediction performance on a target domain, we fine-tune the prediction model with ${\train}$ and only the input features in the target domain.
% \Hu{check}. 
% \cwu{briefly tell what data are used for fine-tuning}.
Fine-tuning aims to minimize both the hybrid errors and the distribution difference between the latent representations (output of the encoder) from source domains and the target domain. %, as discussed in Sec.~\ref{sec:cmd}. 

%\subsubsection{Distribution Difference Metric} %\label{sec:cmd}

% \jw{added a new paragraph to further solidify the domain invariant learning}
% \jw{newly added begin}\
\vspace{1mm}
\noindent\textbf{CMD to measure distribution difference.} Our goal is for our fine-tuned predictor to perform well on different DNN models and devices.
One of the most important theoretical results in domain invariant learning is that the generalization risk of the model (i.e., the difference in the average error of a cost model evaluated on model $M$ and device $D$ versus that on model $M'$ and device $D'$) can be mitigated by reducing the distance among different domains in the latent space~\cite{redko2020survey}.
To put this in our context, recall that $h$ is our encoder that maps input features to the latent space. With a suitable distribution difference metric $\Delta(\cdot)$, we have the following bound on the generalization risk for $D, D'\sim \deviceSpace$ and $M, M' \sim \modelSpace$~\cite{redko2020survey}:

% \vspace{-3mm}
\begin{equation} \begin{aligned}\label{eq:domain_adaptation_guarantee}
    & \expectation_{(\randvar{x},\randvar{y}) \sim \prob_{M}(\randvar{x},\randvar{y}|D)}
     [\loss(\predictor(\randvar{x}),\randvar{y})|D]   \\
    & \leq \expectation_{(\randvar{x},\randvar{y}) \sim \prob_{M'}(\randvar{x},\randvar{y}|D')}[\loss(\predictor(\randvar{x}),\randvar{y})|D'] \\
    & + \Delta(\prob(h(\randvar{x})|D,M),\prob(h(\randvar{x})|D',M'))
\end{aligned}\end{equation}
% \vspace{-3mm}

\noindent which says the difference in the expected performance of the system between any pair of $\prob_{M'}(\randvar{x},\randvar{y}|D)$ and $\prob_{M}(\randvar{x},\randvar{y}|D)$ is bounded by the distance between their induced representation distributions in the latent space.

% \jw{add some transition to improve the transition between these two paragraph}
Based on this distribution-difference-based bound, we can minimize the following objective in our model fine-tuning %introduce a method for learning domain-invariant representations through domain divergence minimization
~\cite{abhinav2020domain}, %find a representation space where 
to project representations of samples from different domains %are projected 
close together in the representation space: %Then, the domain divergence minimization principle amounts to the following learning objective,

% \vspace{-3mm}
\begin{equation}\begin{aligned}\label{eq:modified_learning_objective}
     \min_{h,f} & \underbrace{\quad \loss_{pre\_train}(f(h(\dataSet)))}_{\text{standard training loss}} + \\
                & \underbrace{\expectation_{D,D'\sim \deviceSpace || M,M' \sim \modelSpace} [\Delta(\prob(h(\randvar{x})|D,M),\prob(h(\randvar{x})|D',M'))]}_{\text{regularization for minimizing representation difference}}
\end{aligned}\end{equation}
% \vspace{-3mm}

One practical divergence measure for $\Delta(.)$ is the Central Moment Discrepancy~\cite{zellinger2019cmd}, which is theoretically grounded, efficient to implement and compute, and has shown superior empirical success in learning domain invariant representations~\cite{long2016cmd,qi2021gnncmd}. Given two distributions $\prob_1,\prob_2$, the CMD distance can be defined as:

% \vspace{-3mm}
\begin{equation}\begin{aligned}
    \text{CMD}(\prob_1,\prob_2) = \frac{1}{|b-a|}\|\expectation(\prob_1)-\expectation(\prob_2)\|_2  + \\ \sum_{j=2}^{\infty}\frac{1}{|b-a|^j}\|\Omega_j(\prob_1)-\Omega_j(\prob_2)\|_2
\end{aligned}\end{equation}
% \vspace{-3mm}

\noindent where $a, b$ are the joint distribution support of the distributions $\prob_1$ and $\prob_2$, respectively, and $\Omega_j(\prob_1) = \expectation(\prob_1 - \expectation(\prob_1))^j$ 
is the $j$-th order moment. 
In practice, a limited number of moments are usually needed (e.g., $j\le 5$) \cite{long2016cmd}.

We then obtain the training objective for our predictor fine-tuning as follows:
\begin{equation} \label{eq:cmd_loss} \begin{aligned} 
\loss_{fine\_tune} =\loss_{pre\_train} + \alpha \times \text{CMD}(z_{s},z_{t})
\end{aligned} \end{equation}
where $z_{s}$ and $z_{t}$ are latent representations of the source domain (e.g., a set of devices for CDPP) and the target domain (e.g., a set of devices disjoint to source devices for CDPP), respectively. $\alpha$ is a coefficient decided by the auto-tuner.
% \cwu{$\alpha$ has been used earlier - change the symbol} 

%\subsubsection
\vspace{1mm}
\noindent\textbf{Sampling Strategy for Fine-tuning on Target Device.}
To achieve fast fine-tuning for accurate performance prediction on a new device, we should select tensor programs that best represent the entire dataset to profile on the target device. %To profile the execution time of a tensor program, it is necessary to specify the operator and the optimizations applied to it \cwu{not clear why necessary}. However, it can be difficult to retrieve which optimizations have been applied from a tensor program alone, making it impractical to directly specify a tensor program for profiling. 
As tensor programs for different devices may not be exactly the same (e.g., a tensor program for GPU cannot be directly run on CPU), %we cannot directly sample representative tensor programs and profile them on different devices; instead, 
we sample representative tasks (instead of tensor programs) and profile the respective tensor programs of the tasks on the target device. We %assume tensor programs are derived from 
consider the same set of $\mathbb{T}$ tasks 
% (operator) %and predict performance of their tensor programs 
on different devices. %We select representative tasks and profile corresponding tensor programs on the target device.
%Suppose for each device, all tensor programs are derived from $T$ tasks, 
Each task $\tau$ in set $\mathbb{T}$ has a set of device-independent features $X_\tau$ (including features of its tensor programs) and corresponding latent %hidden 
representations of its tensor programs $Z_\tau$. Let $Z = \bigcup_{\tau=1}^{|\mathbb{T}|} Z_\tau$ denote the set of all %the hidden 
latent representations of tasks in $\mathbb{T}$. %on each device. 
Our goal is to decide a subset $\mathcal{Q}$ of $\kappa$ tasks to profile on the target device, such that the distribution difference between the %hidden 
latent representations of the selected tasks and those of all tasks %$z$ 
is minimized.
\begin{comment}
$$
\underset{\mathcal{Q}}{\mathrm{argmin}} ~\Delta(Z, \bigcup_{j=1}^\kappa Z_{\mathcal{Q}_j})
$$
\noindent where $Z_{\mathcal{Q}_j}$ is the set of latent representations of tasks in $\mathcal{Q}_j$.
\end{comment}
% where $\upsilon$ is a function to measure the distribution difference between two distributions.
% \jw{need to change the above formulation}

% \Hu{We can have some theoretical bound here}
% \jw{newly added, check for notation consistency}
% \mypara{Convergence Bound Analysis.} 
Recall that $X$ is the feature space of all possible tensor programs and $\mathcal{H}$ is the latent/embedding space. %and $\loss(.,.)$ is the loss function.
Let $\mathcal{C}=\bigcup_{j=1}^\kappa X_{\mathcal{Q}_j}$ denote the feature set of tasks %tensor programs 
used for fine-tuning.
Let $c_x= \argmin_{c \in \mathcal{C}}\Delta_{\mathcal{H}}(h(x),h(c))$ denote the closest $c \in \mathcal{C}$ to $x$ in the latent space, and $\epsilon := \max_{x \in X} \min_{c \in \mathcal{C}}\Delta_{\mathcal{H}}(h(x),h(c))$, which captures the maximal distance between any tensor program in the input space and the closest tensor program among the finetuning samples in the latent space.
%When the loss function $\mathcal{L}$ is smooth with respect to the representation/embedding of the tensor program, 
%As our fine-tuning loss function in xx \cwu{point to the formula and check if the following true} is smooth with respect to the representations of the tensor program,
We derive the following bound for the generalization risk of the fine-tuned model with respect to the sampling strategy (see Appendix~\ref{appendix:cluster_guarantee} for the detailed proof):

% \vspace{-3mm}
\begin{equation}\begin{aligned}\label{eq:cluster_guarantee}
     \mathcal{L}_{\mathrm{pre\_train}}(x) \leq \mathcal{L}_{\mathrm{pre\_train}}(c_x) + \mathcal{O}(\epsilon) 
\end{aligned}\end{equation}
% \vspace{-3mm}

% \Hu{Is the loss function here the same as the loss function we used for training, which is mentioned in Section 5.1? $\mathcal{L}(x)$ or $\loss(\predictor(x),y)$}
% \jw{does not really matter so long as the l-lips property holds.}
% ### Solution 1: Greedy
%Based on the above analysis/results, 

We propose a clustering-based sampling strategy that minimizes $\epsilon$ and hence lowers the generation risk, as shown in Algorighm~\ref{algo:sampling}. We first perform K-means clustering to divide all tensor program features in $X$ into $\kappa$ clusters, $\mathbb{G}_e, e=1, 2, ..., \kappa$, which are sorted according to the cluster size (line 1-2). 
Then we calculate a table $\Psi$, with entry $\Psi[e, \tau] = (\sum_{j=1}^{|X_\tau|} ||\mathbb{G}_e- X_\tau[j]||_2)/|X_\tau|$ denoting the average distance of task $\tau$'s features %samples
$X_\tau$ to the center of cluster $e$ (line 6). We pick one task for each cluster to profile starting from the cluster with the largest cluster size and % label the task % \cwu{clarify what `label the task' means} once it is selected
remove the task from the candidate task set once it is selected. %for each cluster one by one. For each cluster $i$, we select 
Specifically, for cluster $e, e=1, 2, ..., \kappa$, we select the task with the smallest $\Psi[e, \tau]$ value in the candidate set (line 9-15).

\begin{algorithm}[!t]
\caption{Clustering-based sampling strategy}
\label{algo:sampling}
\small
\begin{algorithmic}[1]
\Require all tensor program features $X$, number of tasks to select $\kappa$, all tasks $\mathbb{T}$
\Ensure Selected tasks $\mathbb{T}^*$
\State $\mathbb{G} \leftarrow \textsc{KMeans}(X, \kappa)$
\State Sort $\mathbb{G}$ in descending order according to the cluster size
\State $\mathbb{T}^* \leftarrow \emptyset $
\For{$e = 1 \rightarrow \kappa$}
    \For{$\tau = 1 \rightarrow |\mathbb{T}|$}
       \State $\Psi[e, \tau] = (\sum_{j=1}^{|X_\tau|} ||\mathbb{G}_e- X_\tau[j]||_2)/|X_\tau|$
    \EndFor
    \State $d \leftarrow$ sorted index of $\Psi[e, 1:|\mathbb{T}|]$ in ascending order
    \For{$\tau$ in $d$}
        \If{$\tau$ in $\mathbb{T}$}
            \State $\mathbb{T}^* \leftarrow \mathbb{T}^* \cup \{\tau\}$;$\mathbb{T} \leftarrow \mathbb{T}$\textbackslash $\tau$; break
        \EndIf
    \EndFor
\EndFor
\State \Return $\mathbb{T}^*$
\end{algorithmic}
\end{algorithm}

\noindent\textbf{NAS and Automatic hyper-parameter tuning.}
We employ an auto-tuner to automatically optimize the model architecture and hyper-parameters in our cost model to minimize the prediction error. For model architecture, our focus is primarily on determining the values for the number of transformer encoder layers, the number of MLP layers in the decoder, and the intermediate dimension. As for the hyper-parameters in the cost model, we conduct a search for variables such as $\alpha$ in Eqn.~\ref{eq:cmd_loss}, learning rate, weight decay, optimizer (Adam or SGD), learning rate scheduler, and batch size.
To implement the auto-tuner, we utilize Optuna~\cite{optuna_2019}, a hyper-parameter optimization framework equipped with state-of-the-art exploring algorithms. We use Optuna to explore the optimal combinations of the aforementioned variables to minimize the evaluation error. Since NAS (Neural Architecture Search) and hyper-parameter search are not the primary focus of this paper, instead of exhaustively searching for the optimal combination, we terminate the auto-tuner after testing approximately 1000 configurations and choose the best one among them, which we find performs well across all our experiments. For detailed values of variables mentioned above, please refer to Appendix \ref{appendix:auto-tune}.

\begin{table*}[!t] \centering
    \begin{tabular}{c|c|c|c|c|c|c} 
        \hline
        Taxonomy & Device & Clock (MHz) & Mem. (GB) & \makecell[c]{Mem. band-\\width(Gbps)} & Cores & $\#$ of Samples\\
        \hline
        \multirow{5}*{\makecell[c]{NVIDIA GPUs}} & T4~\cite{nvidia2018T4} & 1590 & 16 & 320 & 40 & 9M\\ 
        ~ & K80~\cite{nvidia2014K80} & 824 & 12 & 240.6 & 26 & 9M \\ 
        ~ & P100~\cite{nvidia2016P100} & 1329 & 16 & 732.2 & 56 & 9M\\ 
        ~ & V100~\cite{nvidia2017V100} & 1530 & 32 & 900 & 80 & 2M\\ 
        ~ & A100~\cite{nvidia2020A100} & 1410 & 40 & 1555 & 108 & 2M \\ 
        \hline
        \multirow{1}*{\makecell[c]{Inference Accelerators}} & HL-100~\cite{habana2019goya} & 1575 & 8 & 40 & 11 & 4K \\ 
        \hline
        \multirow{3}*{\makecell[c]{CPUs}} & Intel E5-2673~\cite{intel2016e5}  & 2300 & 2048 & 572.24 & 8  & 9M \\
        ~                                 & AMD EPYC 7452~\cite{amd2019epyc}  & 2350 & 2048 & 1525.6 & 4  & 9M \\
        ~                                 & ARM Graviton2~\cite{amazon2010graviton} & 2500 & 32   & 4.75  & 32 & 9M \\

        \hline
    \end{tabular}
    \vspace{2mm}
    \caption{GPU and non-GPU devices used in evaluation}
    \label{table:devices}
    \vspace{-2mm}
    % \vskip -0.15in
\end{table*}

\begin{figure}[t]
\subfigure[Original Y]{
\centering
\label{fig:yDist-origin}
\includegraphics[width=0.225\textwidth, trim=0 0 0 0, clip]{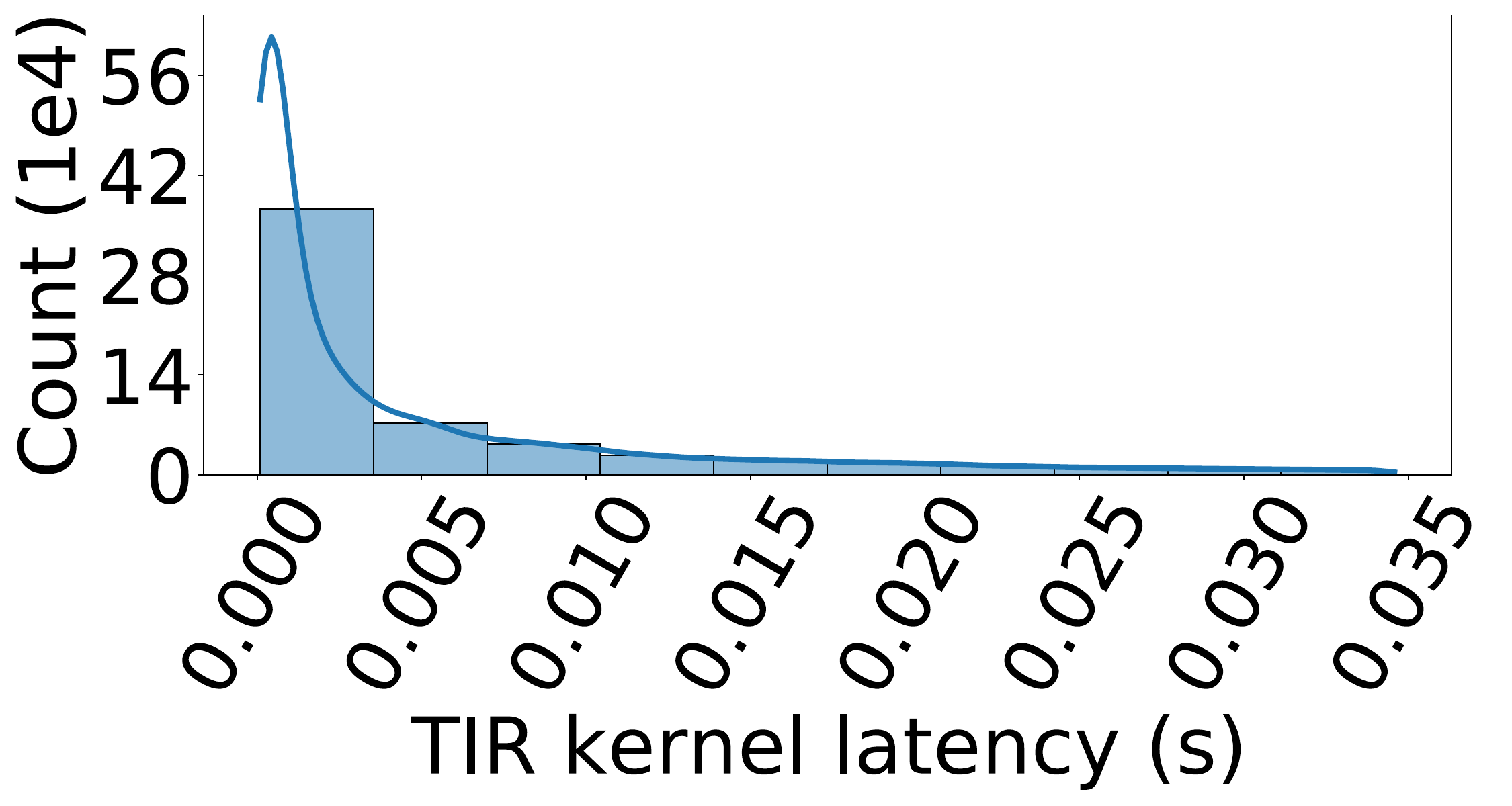}
}
\subfigure[Y with Box-Cox Transformation]{
\centering
\label{fig:yDist-box}
\includegraphics[width=0.225\textwidth, trim=0 0 0 0, clip]{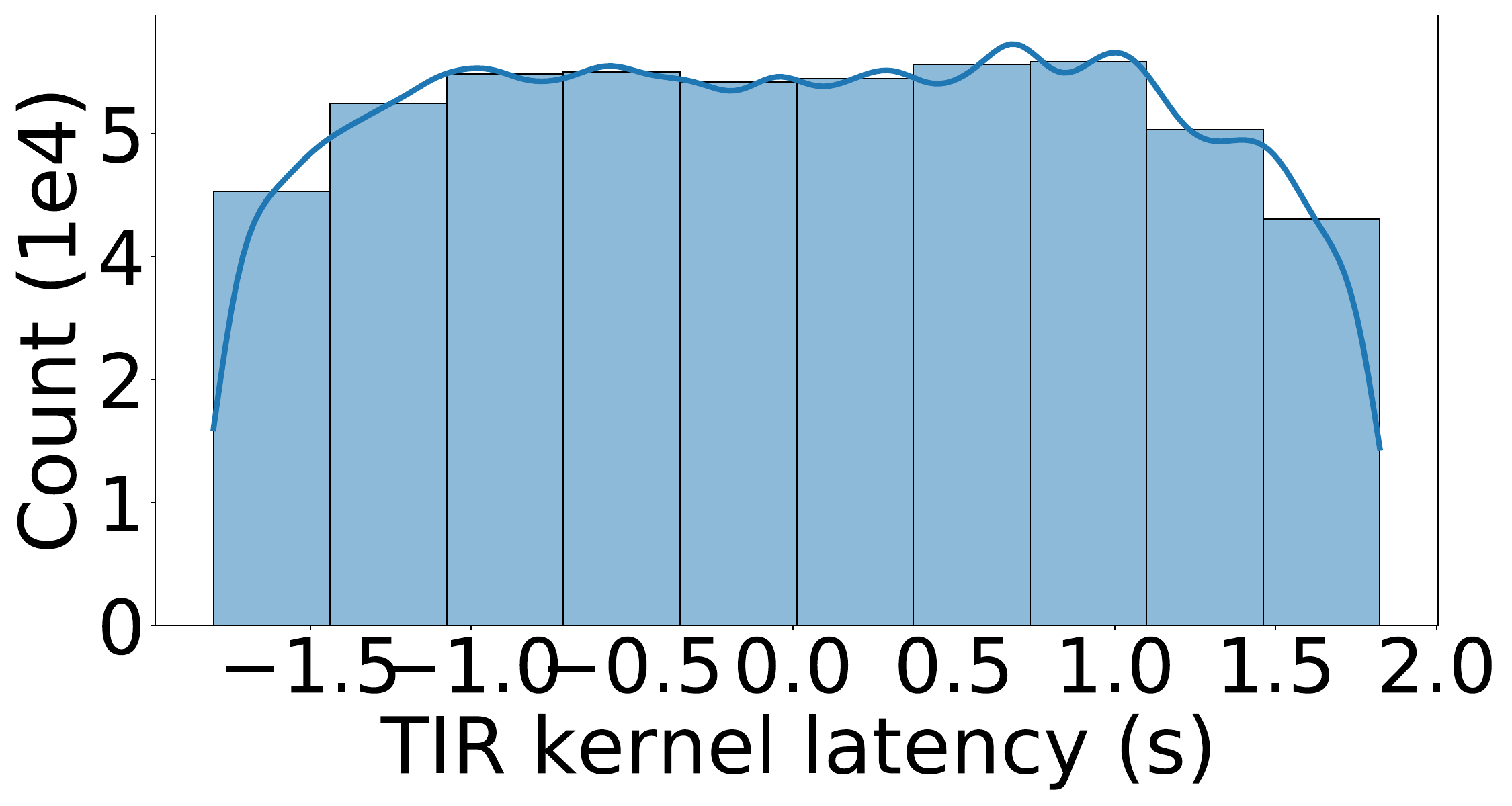}
}
\subfigure[Y with Yeo-Johnson Transformation]{
\centering
\label{fig:yDist-yeo}
\includegraphics[width=0.225\textwidth, trim=0 0 0 0, clip]{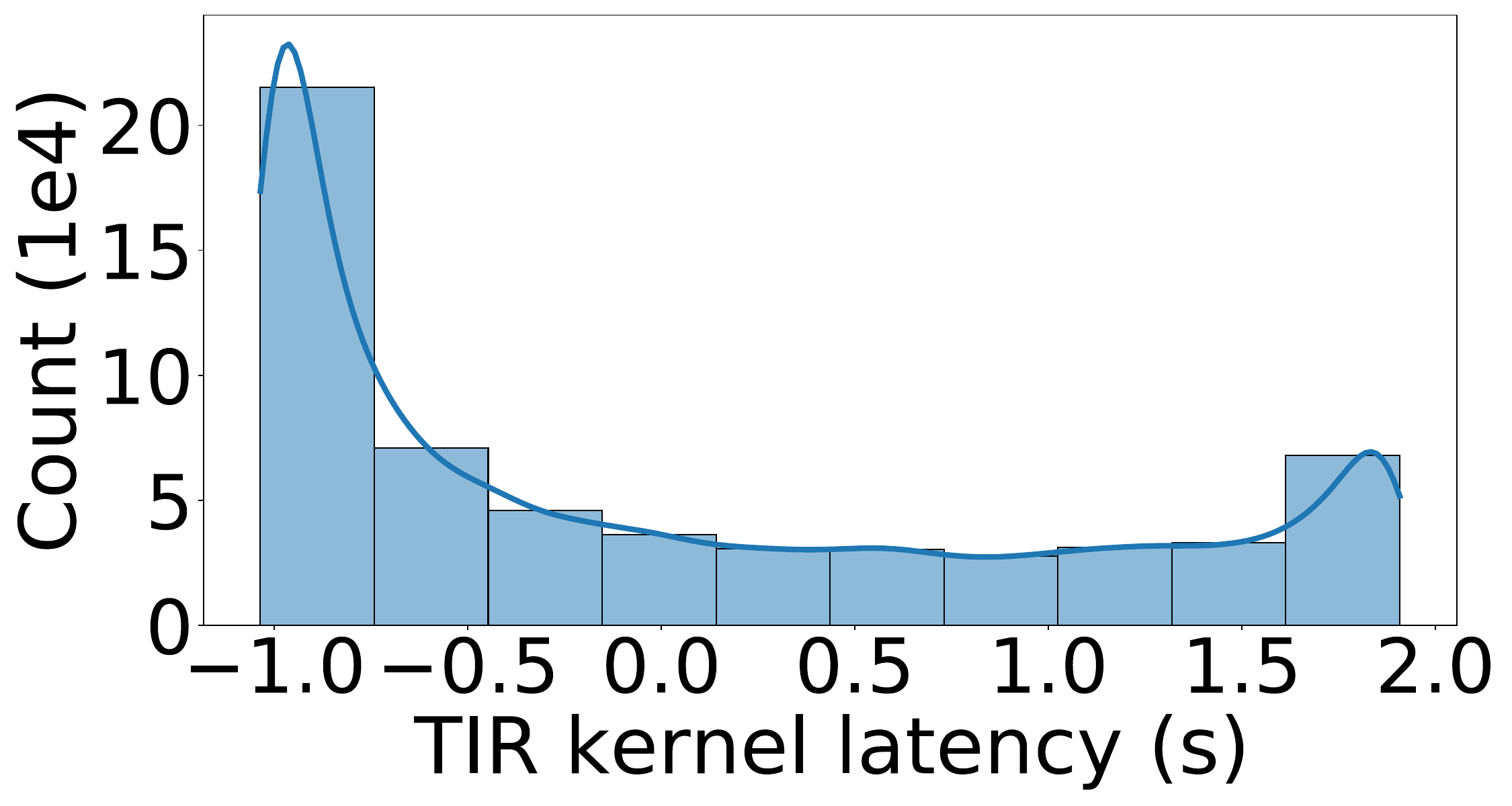}
}
\subfigure[Y with Quantile Transformation]{
\centering
\label{fig:yDist-quantile}
\includegraphics[width=0.225\textwidth, trim=0 0 0 0, clip]{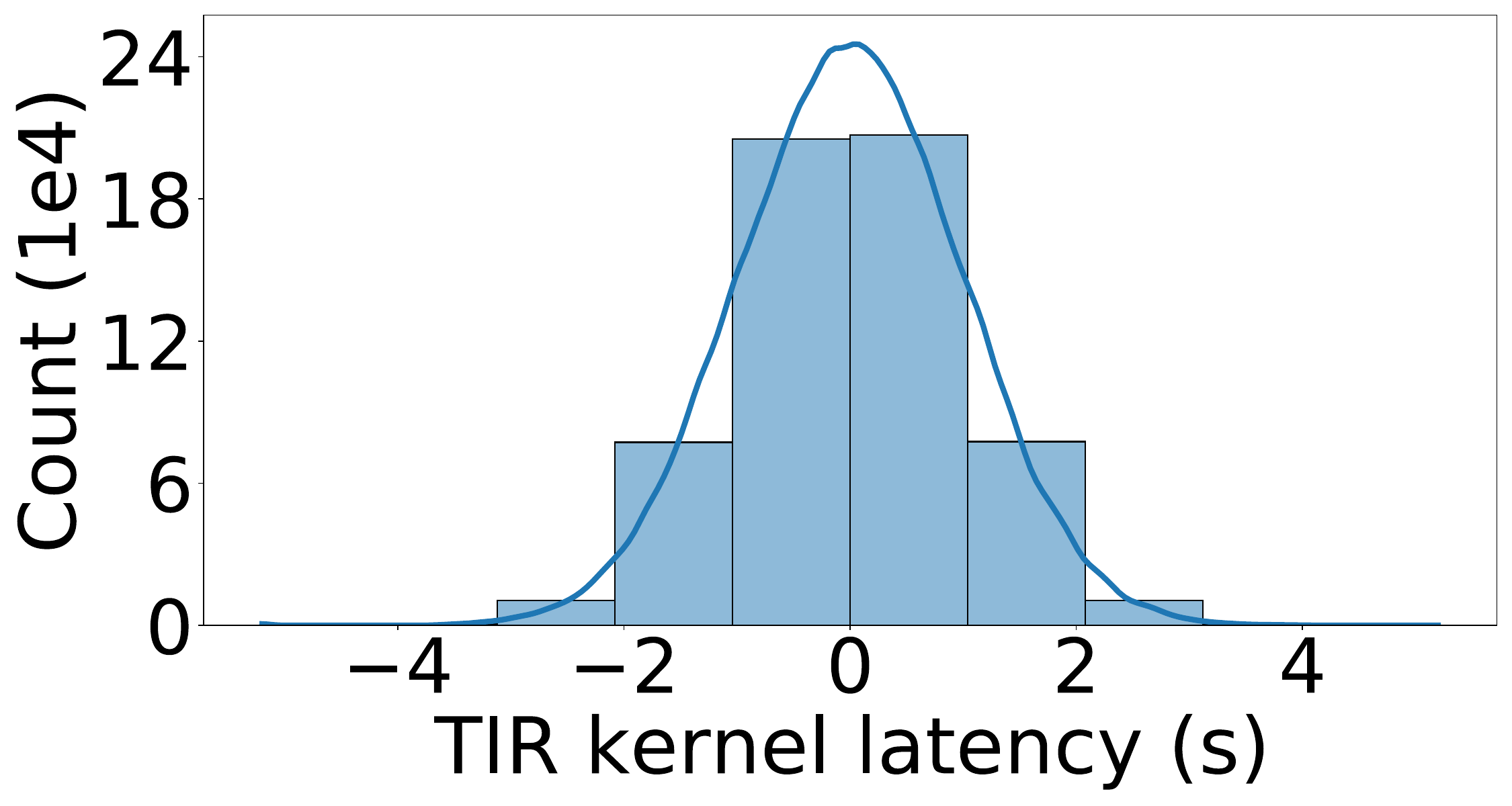}
}
\vspace{-3mm}
\caption{The distribution of tensor program latency (Y) under various normalization methods. The blue line represents the kernel density estimate to smooth the distribution. 
% \cwu{change cost to latency in the figures}
}
\label{fig:yDist}
\vspace{-5mm}
\end{figure}

\vspace{-2mm}
\subsection{Handling Dataset Skewness}

% Fig.~\ref{fig:yDist} shows the distribution of tensor program cost in the Tenset dataset.
% \Hu{Metric for data skewness \url{https://study.com/academy/lesson/skewness-in-statistics-definition-formula-example.html}}
Most ML algorithms exhibit better performance when the input features and output predictions follow standard distributions such as a Gaussian (normal) 
distribution or the uniform distribution ~\cite{dasgupta2011probability}. %Our dataset used to model the performance of tensor programs 
The dataset we use, which contains Tenset and our own profiled records, is generated by randomly sampling schedules for tasks in various DNN models and collecting their performance on different devices. 
Fig.~\ref{fig:yDist-origin} exhibits a long tail distribution of tensor program latencies in the dataset. %has long tails in the direction of large cost. 
Such skewness may significantly hinder an accurate prediction model~\cite{hsieh2020non}. 

Power transformation~\cite{weisberg2001yeo} is a technique to map a non-normal probability distribution more Gaussian-like, %with learnable parameters, 
which can be used to alleviate the effect of outliers. One example of a power transformation is the Box–Cox transformation~\cite{box1964analysis}, which fits an optimal parameter for the mapping through maximum likelihood estimation. Another example is Yeo-Johnson transformation, which %. The main difference between the Yeo-Johnson and Box-Cox transformations is that the former 
can handle negative values and zeros. Quantile transformation %is another class of technique used to 
transforms variables to a standard distribution, including uniform and normal distributions, and is non-parametric.
% Note that logarithmic transformation, which is widely used to solve the long-tailed problem is a special case of Box–Cox transformation. 
% \Hu{remove outliers in the data set}

We choose among the representative normalization methods to make our tensor program latency data more standard, %for the dataset of tensor program costs, we 
by evaluating the distribution of tensor program latency %(Y) 
after applying each method. Fig.~\ref{fig:yDist} shows that the Box-Cox transformation generates a more normal and symmetric distribution with fewer outliers. %when compared to the Yeo-Johnson Transformation and Quantile Transformation. 
Therefore, to rectify the effect of data skewness on our prediction model, %in the regression problem, 
we estimate the optimal parameter of the Box–Cox transformation based on the training dataset using an out-of-the-box library \cite{sklearn_api} and apply inverse Box-Cox transformation to convert the latency back to the original space for error measurement.

\subsection{End-to-end Performance Prediction}
We have developed a replayer that leverages our cost model as a backbone to predict the end-to-end latency of a DNN model. To achieve this, we begin by constructing a TIR-based Data Flow Graph (DFG) for the given DNN model. The DFG is created by establishing connections between dependent TIR functions based on their dependencies. For each node in the DFG, we extract the corresponding TIR kernel or TIR tensor program for feature extraction. Subsequently, we utilize our cost model to estimate the latency of each node on a specific device. It is important to note that multiple TIR functions may be implemented using the same TIR kernel. In such cases, we only perform the cost model inference once for these TIR functions, optimizing computational efficiency. 
The replayer takes the TIR-based DFG and the execution time of each node as inputs. It decides the execution order of nodes on specific devices using a topological sorting algorithm, following the methodology outlined in~\cite{hu2022dpro}, and takes the end time of the last scheduled node as the estimated iteration time. Please refer to Appendix~\ref{appendix:replay} for the detailed simulation algorithm.
% We perform end-to-end performance prediction of a DNN model following~\cite{hu2022dpro}, with the following modifications. 
% We construct the tensor-program-based DFG by connecting dependent TIR functions. 
% Multiple TIR functions may be implemented with the same TIR kernel. Therefore, we map the cost of a TIR kernel to all corresponding TIR functions. 
The replay process to simulate the execution of a DFG also takes device-specific characteristics into consideration. For example, % on 
Habana HL-100 chips %, there are two kinds of Processing Engines (PEs), including 
contain 3 GEMM cores %, which are responsible 
for GEMM and convolution operations, and 8 Tensor Processor Cores (TPC) for SIMD vector computation.
Taking a convolution operator with an estimated execution time of $\hat{y}$ for instance, we replace the corresponding node in DFG with 3 sub-operators that can run in parallel and each sub-operator's execution time is $\hat{y}/3$.

% decide how each TIR function is partitioned \cwu{tell how} and which cores each partition is deployed on \cwu{how we decide it}. 
% We also consider the pipeline between operators deployed on different cores \cwu{explain how we consider the pipeline. HHP: the ``pipeline'' between operators has previously supported by dPRO's replayer and is not new to this paper, thus ignored}.

% \Hu{Open source?}

% Two continuous OPs running on two different PEs can be pipelined.

% % \Hu{Should we consider multiple processes. (8 Cores, with SIMD for single-core running and SPMD parallel for multi-core running). Two continuous OPs running on two different PEs can be pipelined}

%% file: 7implementation.tex
\vspace{-3mm}
\section{Implementation}

%In this paper, we extract device-independent features from TVM intermediate representations (TIRs) considering the popularity of TVM in the DL community. 
We extract device-independent features from TIRs using TVM v0.9.dev0 and build our prediction model using PyTorch~\cite{paszke2017automatic} v1.11.0+cu102. The {\sysname} framework is implemented using Python with 15,331 %lines of code (LoC)
LoC. Users can call the command as follows to query the latency of a network: 
\vspace{-3mm}
\begin{lstlisting}
$ cdmpp <network> <batch_size> <device>
\end{lstlisting}
\vspace{-3mm}
% The detailed model architecture is shown in Table.

% \cwu{explain how a developer can use our framework, e.g., user APIs. HHP: design and show APIs later}

%% file: 8evaluation.tex
\begin{figure}[!t]
\subfigure[GPUs]{     
\centering
\includegraphics[width=0.225\textwidth, trim=8 0 8 0, clip]{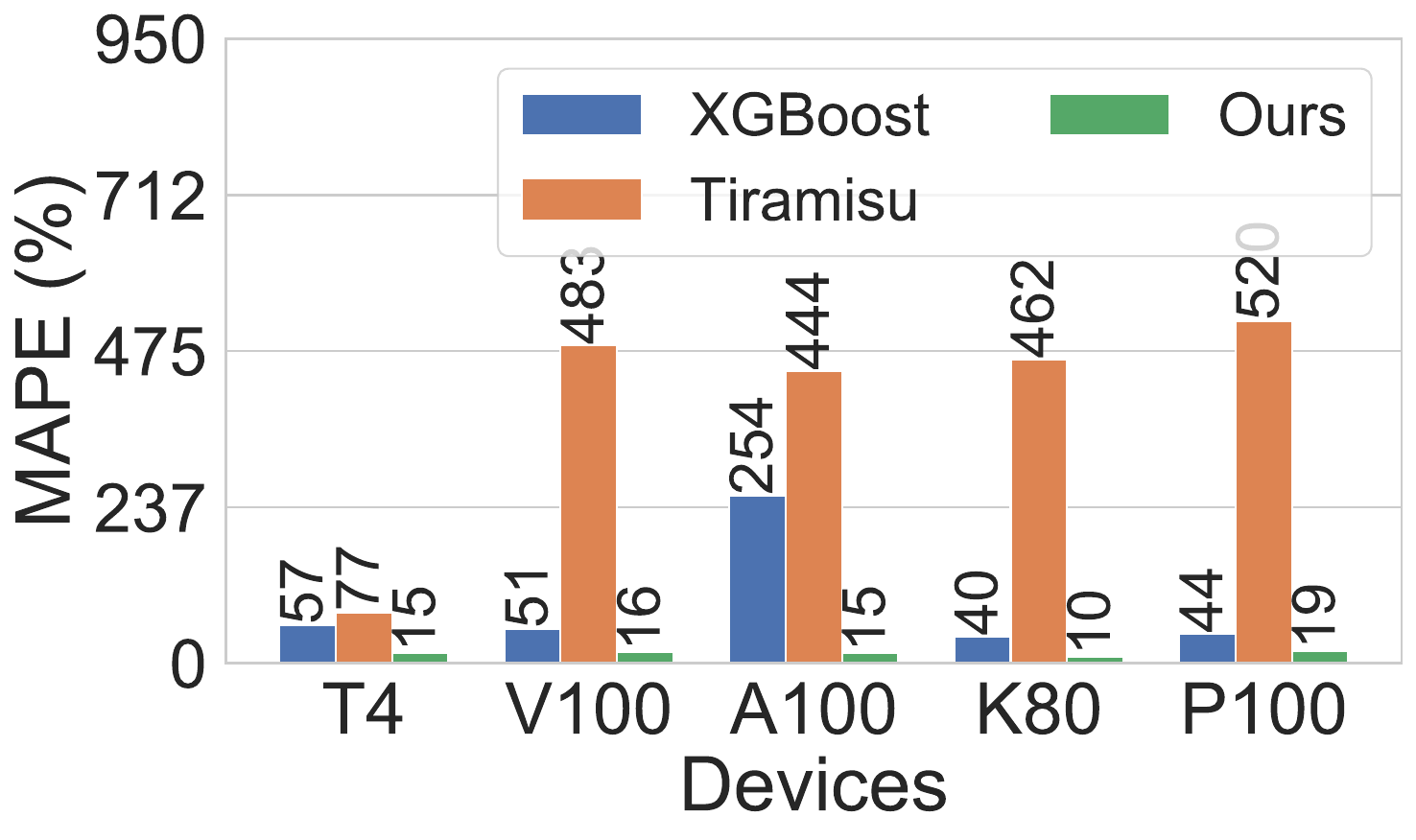}
}
\subfigure[Inference accelerators and CPUs]{     
\centering
\includegraphics[width=0.225\textwidth, trim=8 0 8 0, clip]{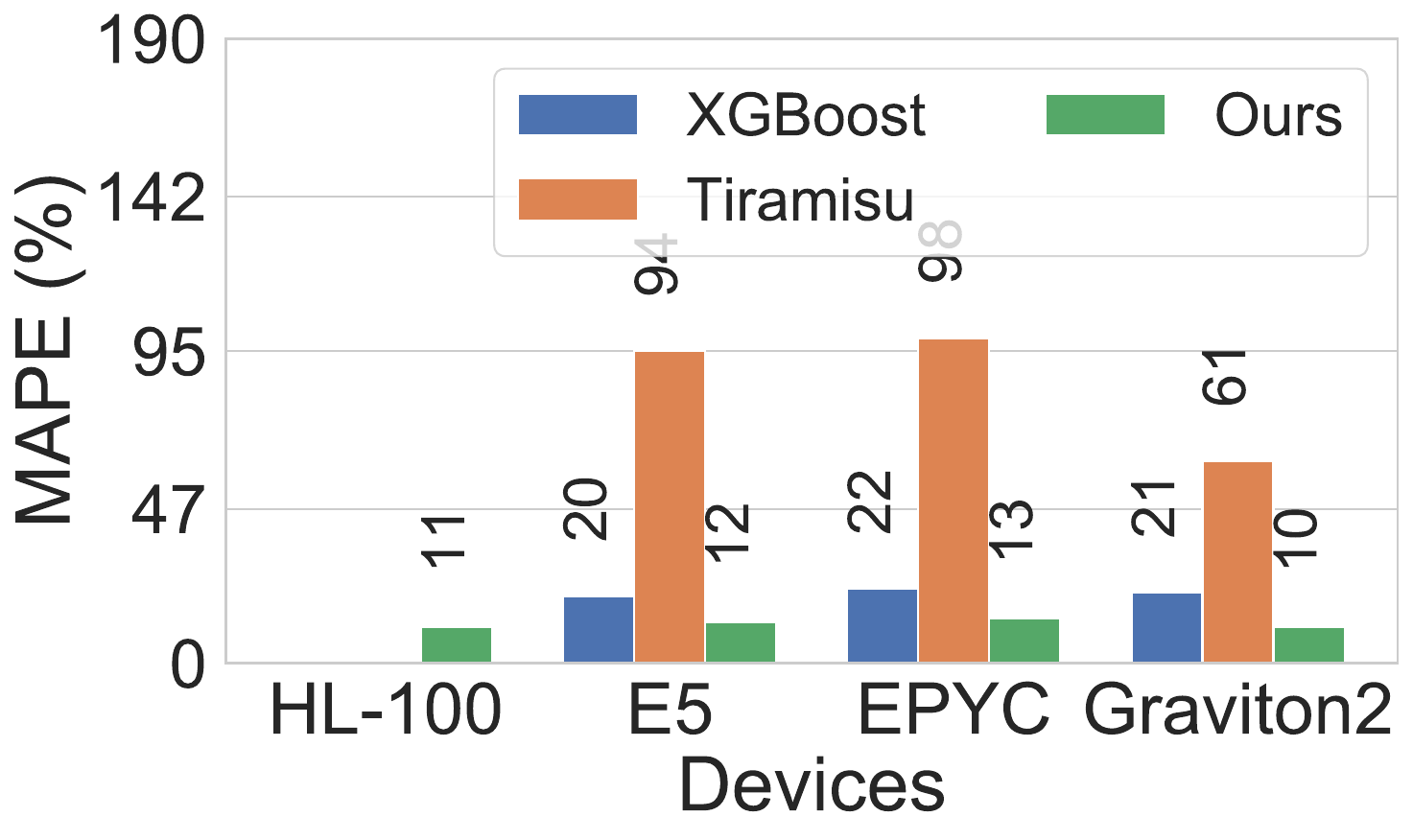}
}
\vspace{-2mm}
\caption{Comparison of prediction errors at the TIR level. The number on top of each bar is the exact MAPE value. 
% \Hu{Increase the font size of figures in this section.}
}
\label{fig:tir-cm}
\vspace{-3mm}
\end{figure}

\section{Evaluation} \label{sec:evaluation}
\subsection{Experimental Setup}
\noindent \textbf{Testbed.} We profile the ground truth of DNN models and tensor programs on devices listed in Table~\ref{table:devices}, including both GPU and non-GPU devices. To train and test the predictor and baselines, we use devices equipped with NVIDIA Tesla V100 32GB GPUs. We use CUDA v11.0~\cite{cuda} and cuDNN v7.6.5~\cite{cudnn} in our experiments.

\noindent \textbf{Dataset.} We %model tensor program performance by 
train our cost model on a large multi-platform dataset (combining Tenset and our own profiling results), which includes tensor program performance records for 120 DL models (ResNet50 ~\cite{he2016resnet}, VGG16~\cite{simonyan2014very}, BERT Base~\cite{devlin2018bert}, etc.) on 5 GPU models (Nvidia K80, P100, T4, V100, A100), three CPU models (Intel E5-2673~\cite{intel2016e5}, AMD EPYC 7452~\cite{amd2019epyc}, Graviton2~\cite{amazon2010graviton}), and one inference accelerator (Habana HL-100). Table~\ref{table:devices} gives the detailed device features we use for cross-device learning and the dataset size collected from each device. For T4, K80 and CPUs, we use the data from Tenset~\cite{zheng2021tenset}; %which is a TVM-based tensor program performance dataset containing 52 million program performance records collected from 6 hardware platforms and 120 networks. 
for the remaining devices, %that Tenset doesn not support, 
we profile tensor program performance on them. %on those more advanced devices. 
We will release the dataset to the community. %for further study of performance modeling. 
For cross-model learning, we %hold out 
use a test set $\mathcal{S}_{hold}$ of 3 DNN models (ResNet-50, MobileNet-V2, and BERT-tiny) for each device. 
% with batch size 1. 
We randomly split the remaining dataset into training, validation, and test sets ($\train$, $\valid$, and $\test$) at a ratio of 8:1:1 for pre-training.
% For cross-model learning, we select one DNN model (e.g., BERT-Base) as the target model and sample two datasets from it as the validation and test set respectively.
% To test the performance of cross-model learning, for the dataset from each device, we randomly split it into a training set, a validation set, and a test set, and the ratio of dataset size of those datasets is $8:1:1$.
For cross-device learning, we pre-train the cost model on $\train$ from source devices and sample tensor programs from $\train$ of the target device for fine-tuning and then evaluate the predictor on $\test$ of the target device.
% where the dataset size ratio is also $8:1:1$.
% For all cases, the ratio of the size of train, validation and test set is $8:1:1$.
We repeat each experiment 3 times and obtain the average results.

% !!! \Hu{Consider using cross-validation instead of random splits to guarantee the results aren’t biased by data partitioning.}

% \noindent {\em Neural Networks.} Numerous DNN models have been developed for various ML applications, e.g., ResNet50~\cite{he2016resnet}, VGG16~\cite{xxx} and InceptionV3~\cite{xxx} for image recognition~\cite{xx}, BERT~\cite{xxx} and GPT-3~\cite{xxx} for natural language processing, Graph Convolution Networks (GCN)~\cite{xxx} and Graph Attention Networks (GAN)~\cite{xxx} for graph-based applications. 

\noindent \textbf{Baselines.} For cross-model learning, we compare {\sysname} with two %state-of-the-art 
SOTA predictors: XGBoost~\cite{chen2015xgboost}, a representative rule-based ML method, and Tiramisu, which also utilizes AST-based features. We modify Tiramisu (https://github.com/ Tiramisu-Compiler/tiramisu) to enable it to support TVM 
% the same idea as in to generate AST-style features and implement the LSTM-based recurrent and recursive architecture on TIR-based AST inputs,
and use the default settings in Tiramisu, e.g., taking MAPE as the learning objective, using cycle learning rate scheduler, etc. For cross-device learning, we use Habitat and TLP
% \Hu{theoretical results (calculated based on the device peak FLOPS and the total number of computations) and } 
as baselines.

\begin{figure}[!t]
\subfigure[Cross-model learning on T4]{     
\centering
\includegraphics[width=0.225\textwidth, trim=8 0 8 0, clip]{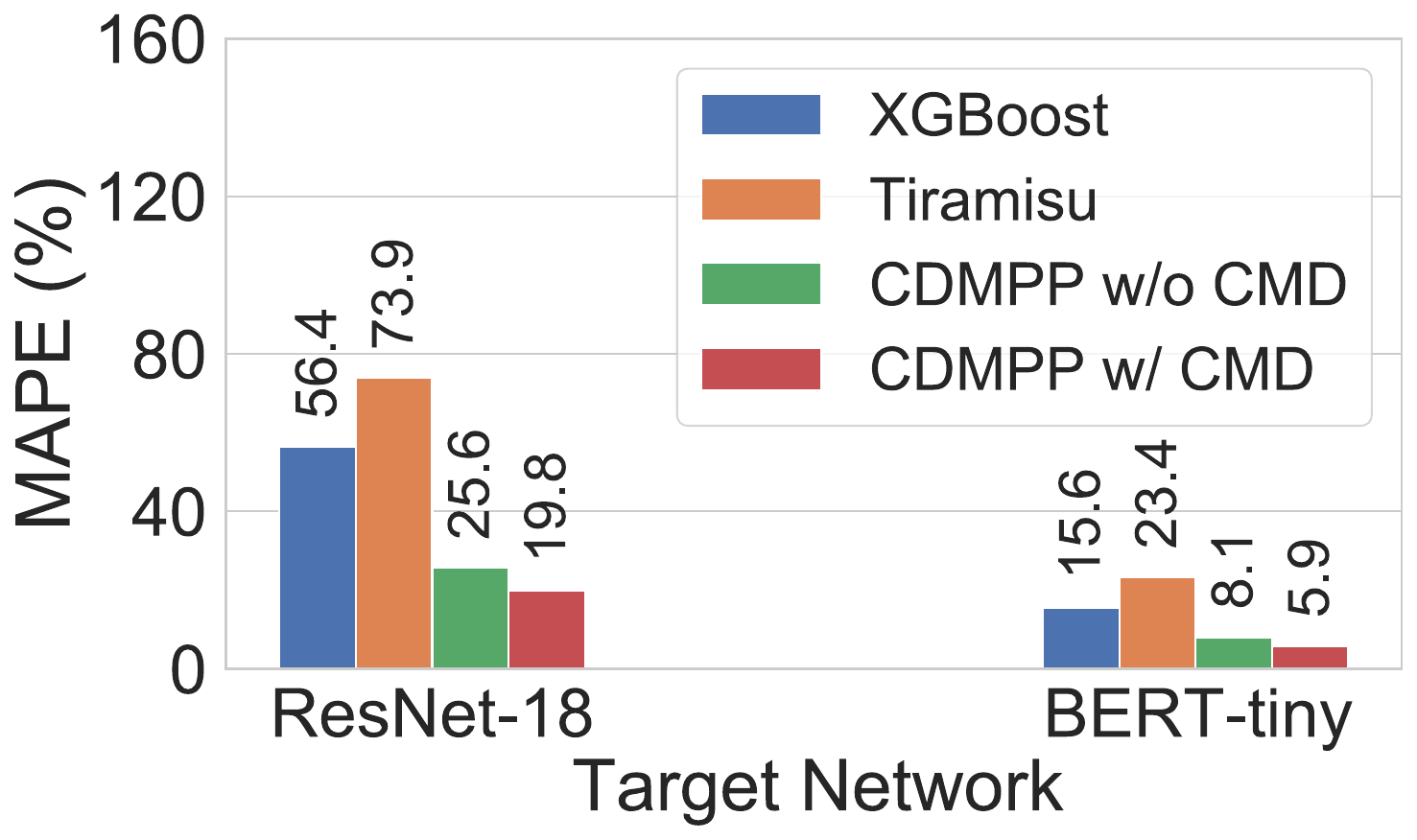}
}
\subfigure[Cross-model learning on EPYC]{     
\centering
\includegraphics[width=0.225\textwidth, trim=8 0 8 0, clip]{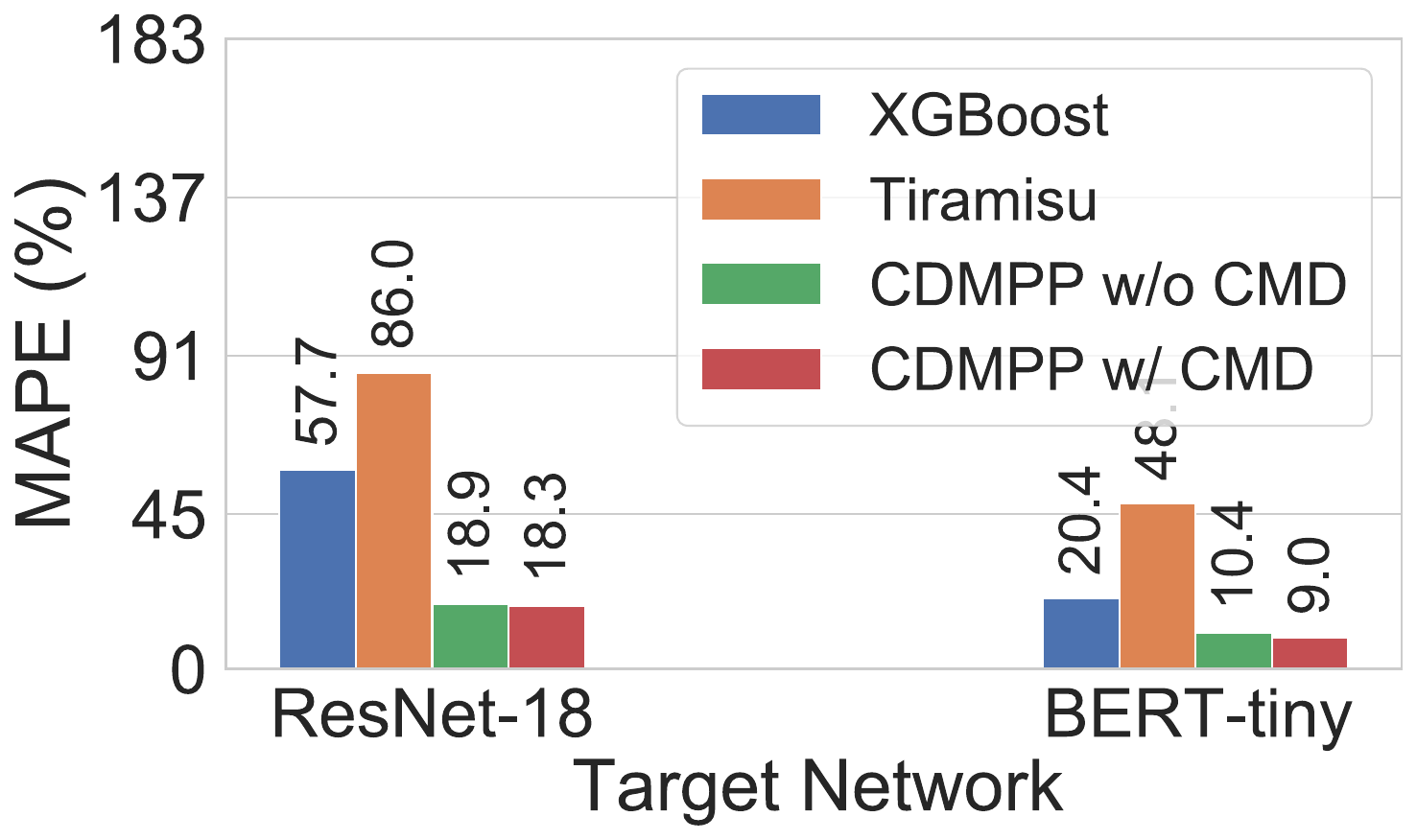}
}
\vspace{-2mm}
\caption{Comparison of cross-model prediction errors. The number on top of each bar is the exact MAPE value.}
\label{fig:cm-finetune}
\vspace{-3mm}
\end{figure}

\begin{figure}[!t]
\subfigure[w/o CMD]{     
\centering
\includegraphics[width=0.225\textwidth, trim=8 0 8 0, clip]{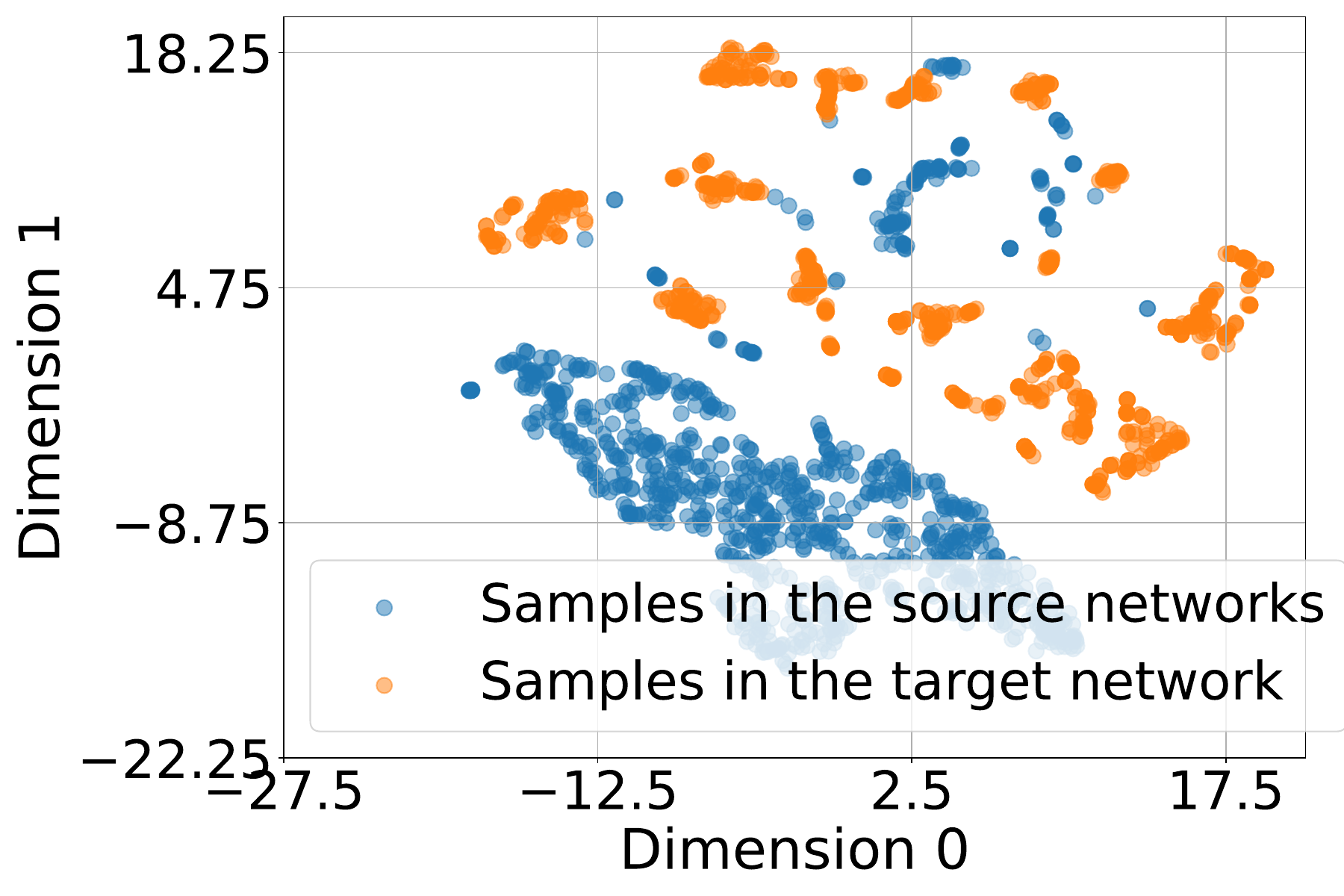}
\label{fig:cm-latent-analysis.a}
}
\subfigure[w/ CMD]{     
\centering
\includegraphics[width=0.225\textwidth, trim=8 0 8 0, clip]{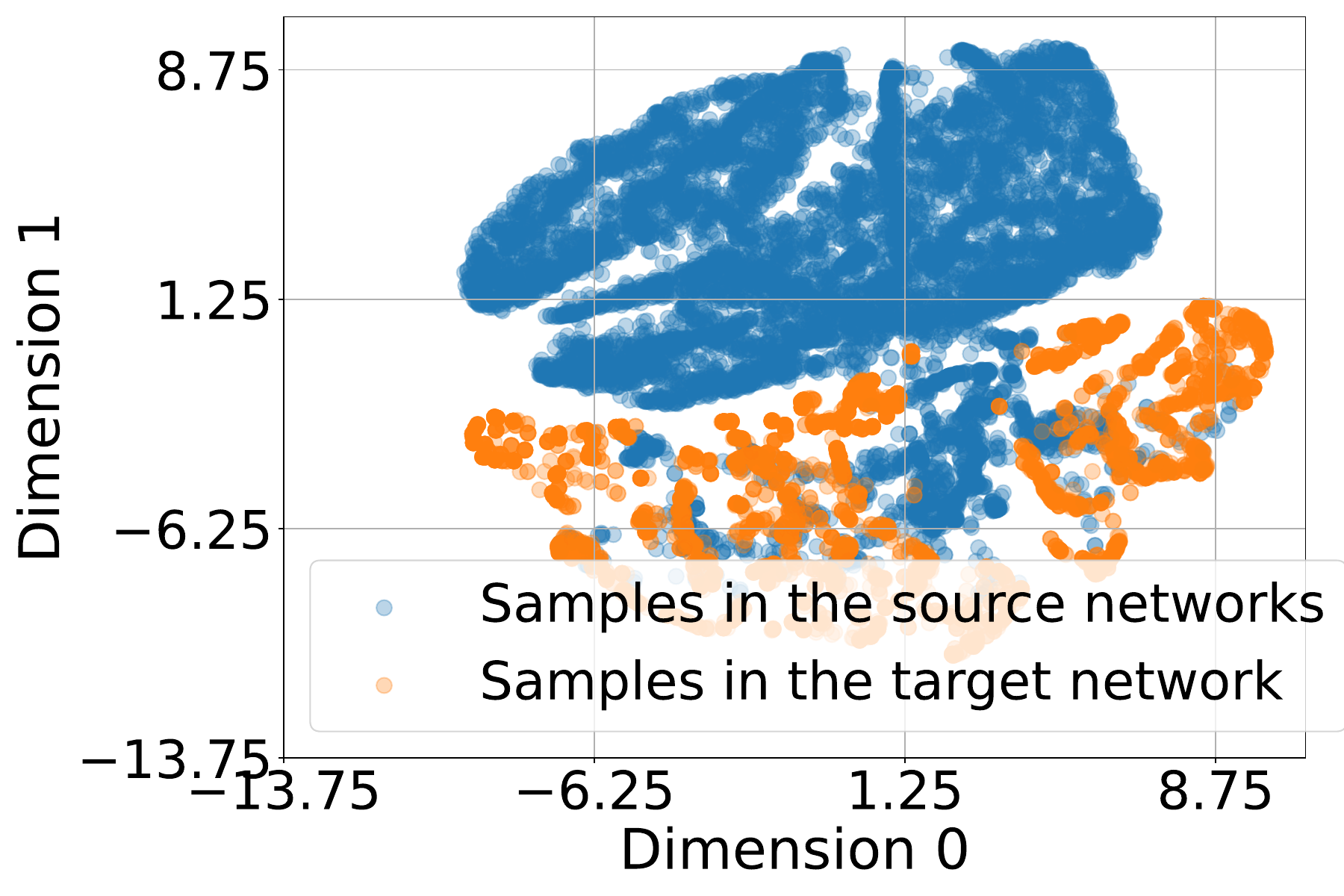}
\label{fig:cm-latent-analysis.b}
}
% \subfigure[MobileNet-V2, w/o CMD]{     
% \centering
% \includegraphics[width=0.225\textwidth, trim=8 0 8 0, clip]{fig/8evaluation/0cmpp_finetune/t4_1_200_mobilenet_v2_dist_visual.pdf}
% \label{fig:cm-latent-analysis.c}
% }
% \subfigure[MobileNet-V2, w/ CMD]{     
% \centering
% \includegraphics[width=0.225\textwidth, trim=8 0 8 0, clip]{fig/8evaluation/0cmpp_finetune/t4_1_200_mobilenet_v2_cmd_dist_visual.pdf}
% \label{fig:cm-latent-analysis.d}
% }
\vspace{-2mm}
\caption{Hidden representation comparison when taking BERT-tiny as the target network.}
\label{fig:cm-latent-analysis}
\vspace{-3mm}
\end{figure}

\subsection{Cross-Model Performance Prediction}

%We first evaluate the prediction error for cross-model performance prediction.

\noindent\textbf{Pre-training performance.}
 % and Fig.~\ref{fig:tir-cm-throughput} and training throughput 
Fig.~\ref{fig:tir-cm} compares the prediction error of our pre-trained predictor with baselines at the TIR level on different devices. %As shown in Fig.\ref{fig:tir-cm}, 
Our predictor %consistently 
achieves a prediction error %of less than 
$<16 \%$ on most devices
% The exception is HL-100, where the prediction error is higher due to the smaller dataset size as indicated in Table \ref{table:devices}. 
and outperforms the baselines %in terms of prediction error
across all devices. Tiramisu~\cite{baghdadi2021deep_tiramisu} exhibits large errors significantly larger than the values ($16\%$) claimed in their paper. The reason is as follows: 1) the recursive LSTM in Tiramisu requires samples in a batch to have the same original AST structure, while the structure of ASTs in our dataset is extremely irregular, as shown in Fig.~\ref{fig:ast_dist_a}, resulting in small batch sizes and large gradient randomness; 2) Tiramisu is primarily designed to estimate the performance speedup when some transformations are applied to a program thus may not perform optimally when estimating the absolute value of tensor programs in our dataset that encompass values across a large range; 3) As stated in their paper, Tiramisu exhibits an exponential increase in prediction error for speedups that deviate far from 1. In detail, the error is larger than $40\%$ when the speedup is smaller than 0.1, while our dataset encompasses a wide range of values, spanning from hundreds of microseconds to tens of milliseconds. This further reinforces the conclusion that Tiramisu's performance is compromised when working with our skewed dataset that contains values across a wide range.
In terms of training efficiency, our measurements of average throughput over all devices for the three methods reveal that {\sysname} (14241 samples/s) improves the training throughput by 1 order of magnitude over Tiramisu (1870 samples/s), because Tiramisu's LSTM %-based architecture
requires recursive computation of loop embeddings according to the structure of input ASTs, %which results in smaller batch sizes and
with multiple forward passes through the LSTM layers in each iteration. Our predictor utilizes a simpler, more regular feature structure that enables large-batch training. The training throughput with XGBoost (644588 samples/s) is larger than ours because it ensembles simple decision trees and has a smaller prediction model size. 
The end-to-end training cost for the CDMPP is 1~2 hours on V100, much smaller than Tiramisu’s 9 hours.
% Furthermore, we utilize the Transformer architecture, which has been proven to be able to achieve higher training throughput than LSTM-based models, due to the parallelizable structure, large batch size, and lower computation complexity of the Transformer~\cite{raffel2020exploring, dai2019transformer}.

% \Hu{Explain why the proposed method can achieve better results than the xgb}

\begin{figure}[!t]
\centering
\includegraphics[width=0.49\textwidth, trim=20 50 100 10, clip]{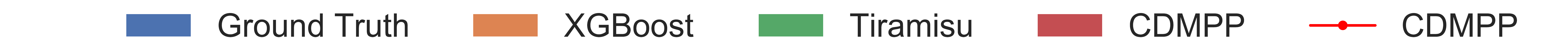}
\subfigure[Predicted onto ResNet50, BS=1]{     
\centering
\includegraphics[width=0.225\textwidth, trim=8 0 8 0, clip]{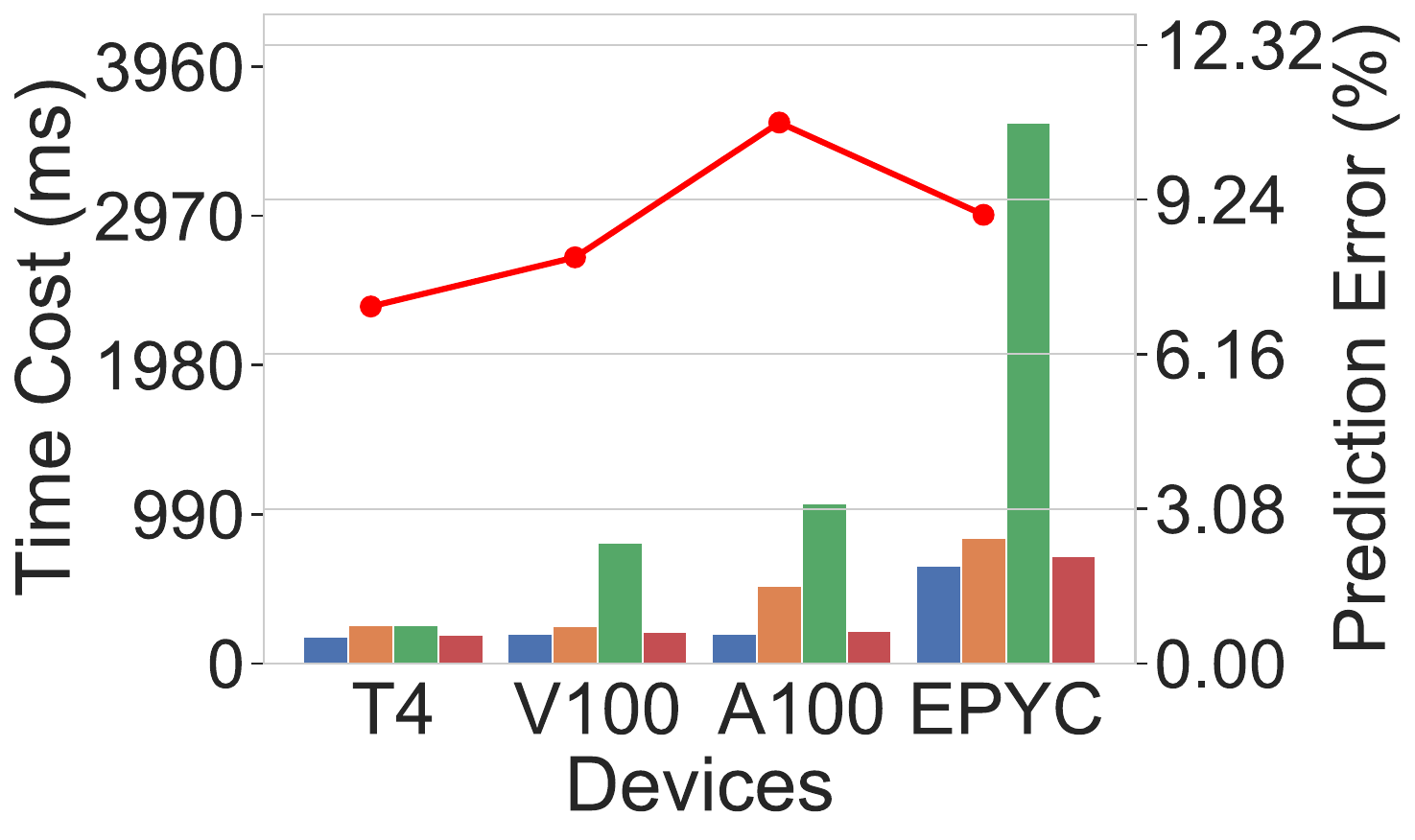}
}
% \subfigure[Predicted onto ResNet50, BS=4]{     
% \centering
% \includegraphics[width=0.225\textwidth, trim=8 0 8 0, clip]{fig/8evaluation/end2end-cm/ResNet-50_4.pdf}
% }
% \subfigure[Predicted onto ResNet50, BS=8]{     
% \centering
% \includegraphics[width=0.225\textwidth, trim=8 0 8 0, clip]{fig/8evaluation/end2end-cm/ResNet-50_8.pdf}
% }
% \subfigure[Predicted onto InceptionV3, BS=1]{     
% \centering
% \includegraphics[width=0.225\textwidth, trim=8 0 8 0, clip]{fig/8evaluation/end2end-cm/InceptionV3_1.pdf}
% }
\subfigure[Predicted onto BERT Base, BS=1]{     
\centering
\includegraphics[width=0.225\textwidth, trim=8 0 8 0, clip]{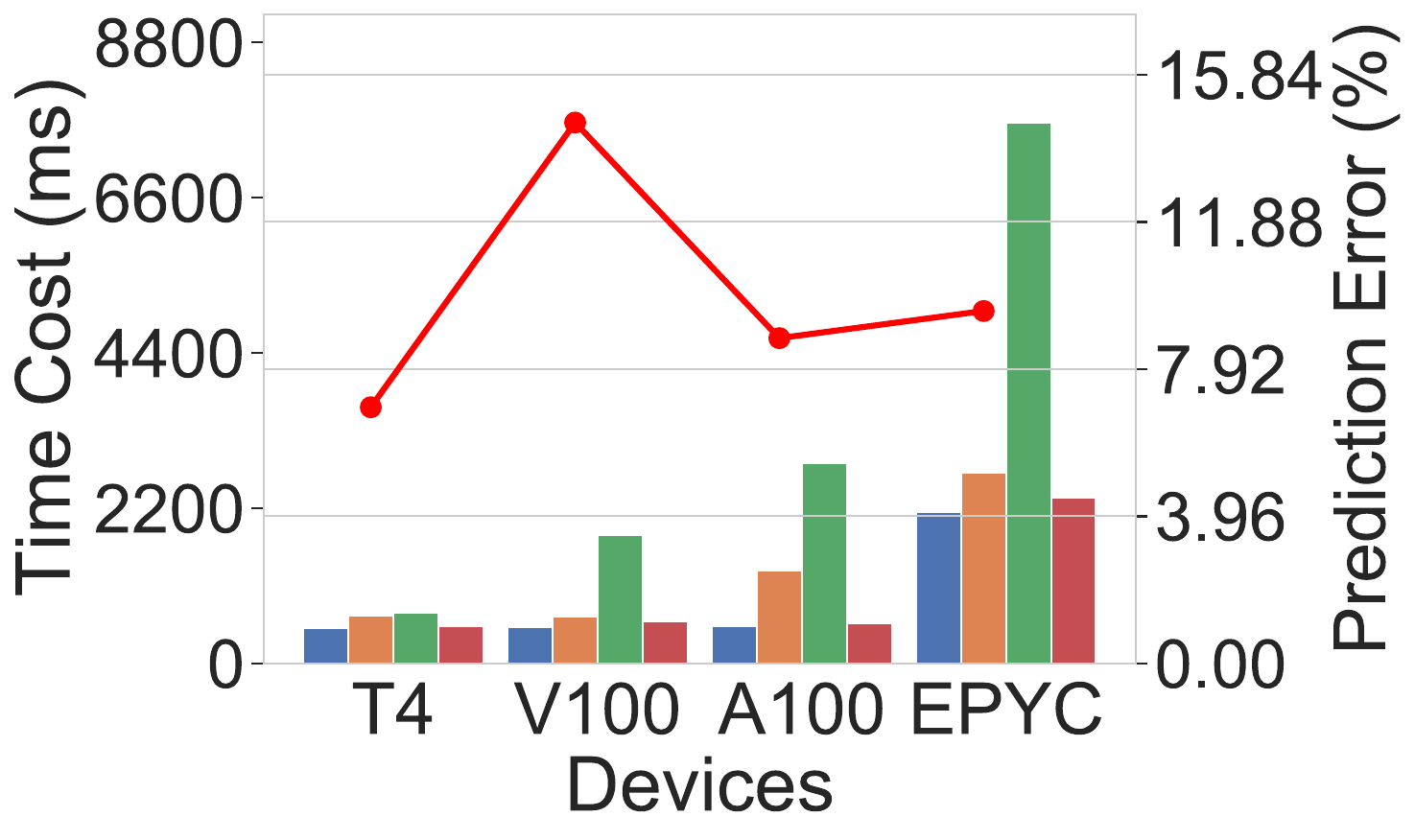}
}
% \subfigure[Predicted onto BERT Base, BS=4]{     
% \centering
% \includegraphics[width=0.225\textwidth, trim=8 0 8 0, clip]{fig/8evaluation/end2end-cm/BERT_Base_4.pdf}
% }
%%% Habana
\subfigure[End-to-end performance prediction on HL-100. ResNet-50 (1) denotes ResNet-50 with batch size=1. ]{     
\centering
\includegraphics[width=0.4\textwidth, trim=8 0 8 0, clip]{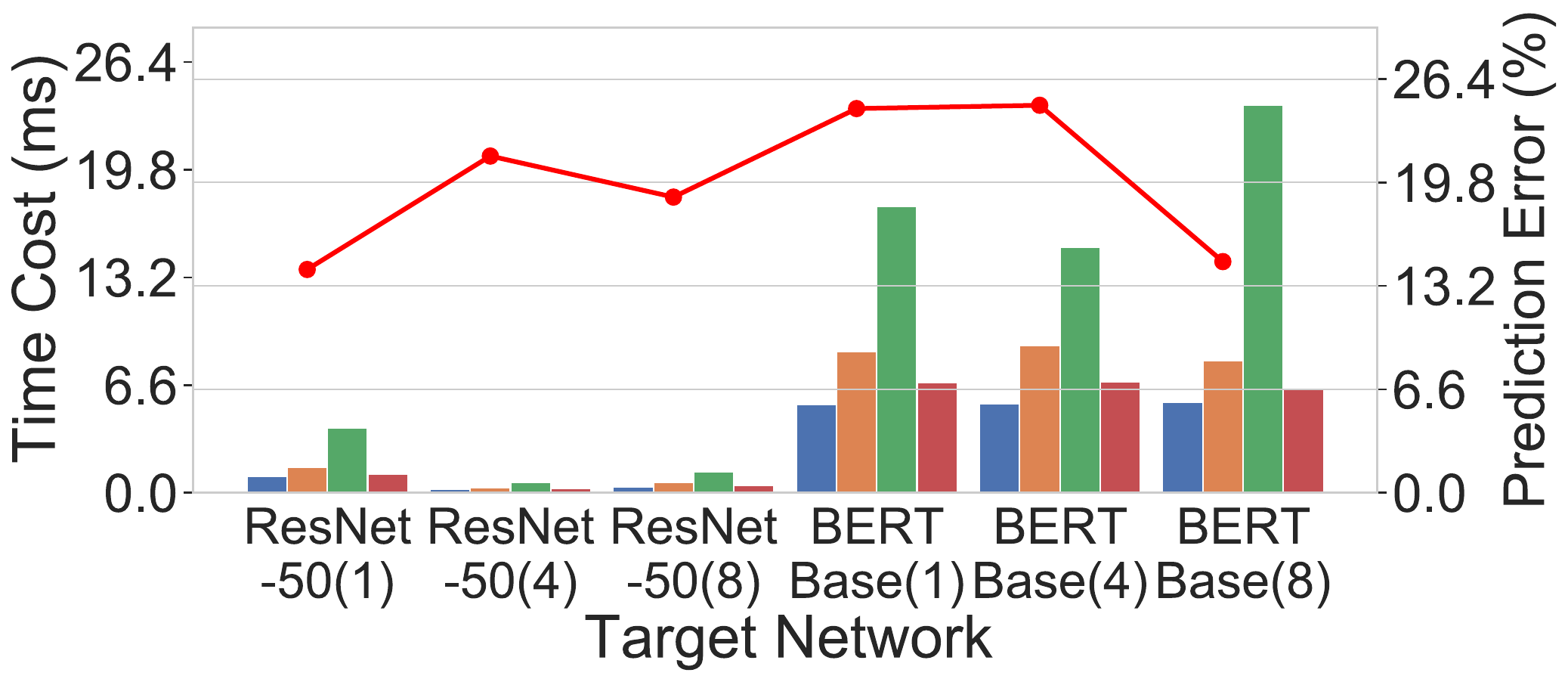}
\label{fig:end2end-cm-habana}
}
\vspace{-2mm}
\caption{End-to-end performance prediction for cross-model learning. 
% \Hu{Increase the font size}
}
\label{fig:end2end-cm}
\vspace{-3mm}
\end{figure}

% \begin{figure}[!t]
% \centering
% \includegraphics[width=0.45\textwidth, trim=8 0 8 0, clip]{fig/8evaluation/end2end-cm/habana.pdf}
% \vspace{-3mm}
% \caption{End-to-end performance prediction on HL-100. ResNet-50 (1) denotes ResNet-50 with batch size=1. \Hu{add legend or merge this with the previous figure}}
% \label{fig:end2end-cm-habana}
% \vspace{-5mm}
% \end{figure}

\noindent\textbf{Cross-model prediction performance.}
We then evaluate the generalizability of our cost model to unseen DNN models by fine-tuning it with input features sampled from $S_{hold}$ for each of the three target networks. %using the training objective as in Eqn.~\ref{eq:cmd_loss}.
Fig.~\ref{fig:cm-finetune} plots the cross-model learning results on T4 and EPYC, which show that {\sysname} always achieves the lowest prediction error. We further explore why {\sysname} performs best by analyzing the hidden representations of source networks and that of the target network in Fig.~\ref{fig:cm-latent-analysis}, where t-SNE~\cite{belkina2019automated} is applied to reduce the representation dimension. The results show that our CMD-based regularization reduces the distribution discrepancy between latent representations from different networks and allows the predictor to generalize better to the target network. 
% However, in some cases, e.g., when taking MobileNet-V2 as the target network, the prediction error is large for all methods, because the distribution shift between the source and target domain is too large for domain adaption technique to mitigate, as shown in Fig.~\ref{fig:cm-latent-analysis.c}. In this case, we recommend to augment the training set to cover more samples.

% \Hu{it would have been useful and interesting to analyze latent space representations of different tensor programs (from different devices and models) to study how "close" they are. I believe t-SNE or UMAP plots would be useful in this case.}

\noindent\textbf{End-to-end performance prediction.}
%In order to measure or estimate the overall performance of DNN networks,
We break each DNN model down into a set of tasks and randomly sample a schedule for each task. 
% Since even for one task, the costs of tensor programs of different schedules can vary greatly, from hundreds of microseconds to tens of milliseconds, we repeat the sampling and evaluation process three times for each network and take the average value.
Fig.~\ref{fig:end2end-cm} compare prediction error of end-to-end model performance %of our proposed method {\sysname}, the widely used XGBoost and Tiramisu, 
against the actual measurements %taken in real-time 
on corresponding devices. %The results show that 
{\sysname} incurs an average prediction error of $12.4 \%$, substantially outperforming XGBoost and Tiramisu, whose average error rates are $63.8 \%$ and $293.6 \%$, respectively. Fig.~\ref{fig:end2end-cm-habana} further demonstrate that {\sysname} can accurately model the performance of HL-100. %This comparison demonstrates the effectiveness of our proposed method {\sysname} in predicting end-to-end performance with high accuracy. 

% ----------------- Cross-device -----------------
\subsection{Cross-Device Performance Prediction}
\begin{figure}[!t]
\subfigure[Prediction onto GPU]{     
\centering
\includegraphics[width=0.225\textwidth, trim=8 0 8 0, clip]{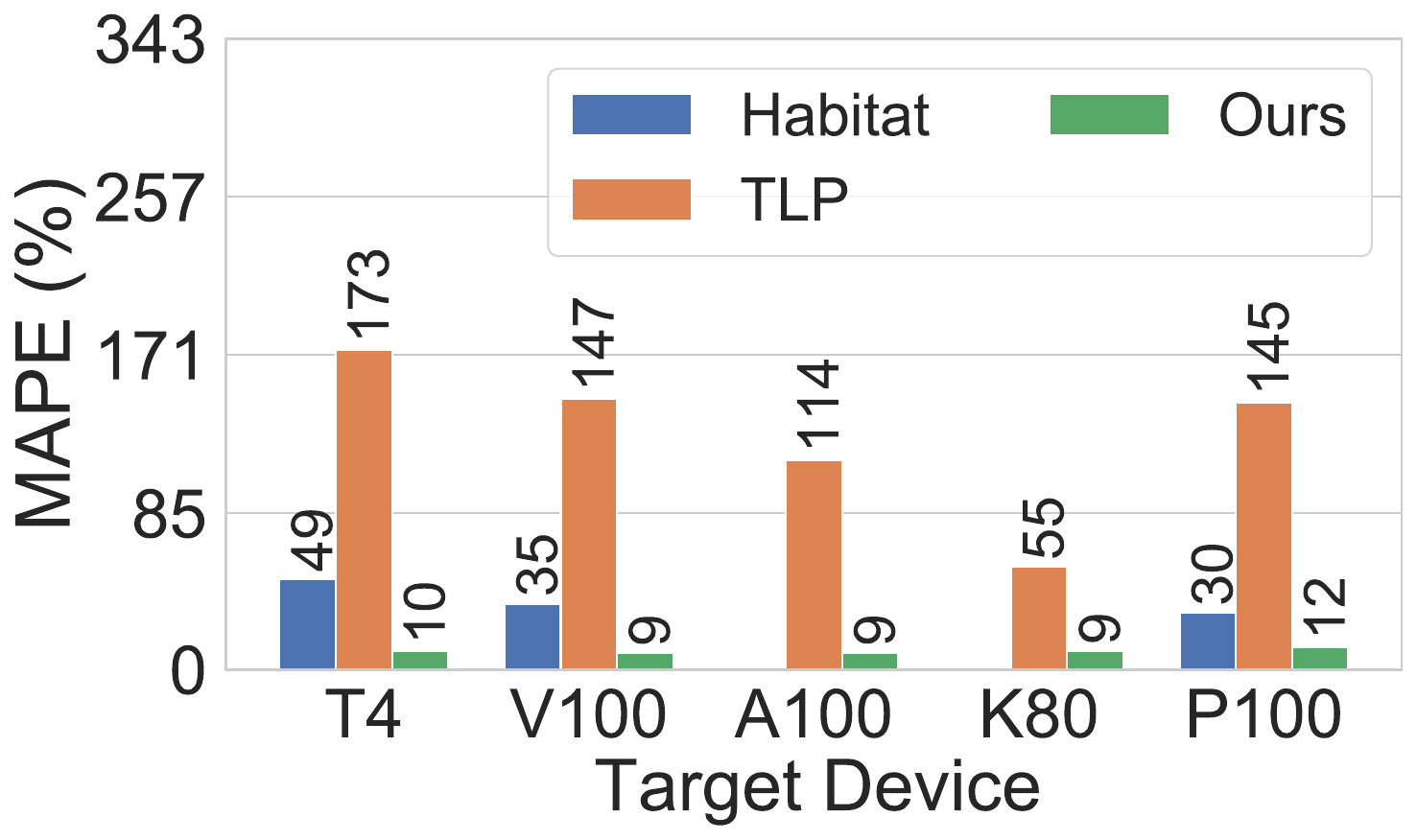}
\label{fig:tir-cd.a}
}
\subfigure[Prediction onto non-GPU device]{     
\centering
\includegraphics[width=0.225\textwidth, trim=8 0 8 0, clip]{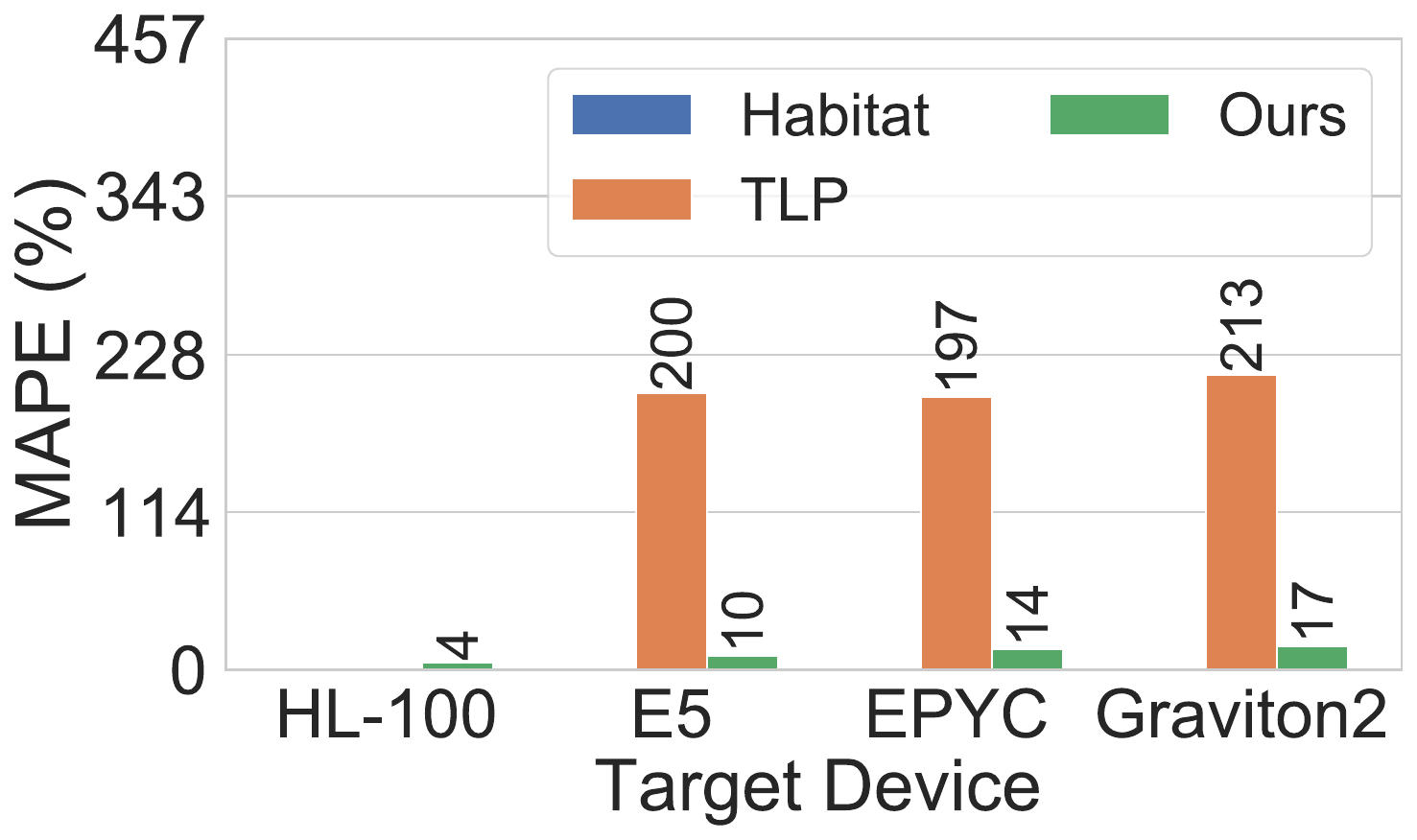}
\label{fig:tir-cd.b}
}
\vspace{-2mm}
\caption{Comparison of cross-device prediction errors at the TIR level. The number on top of each bar is the exact MAPE value. 
% \Hu{check the large errors of TLP} \cwu{change Target Devices to Target Device}
}
\label{fig:tir-cd}
\vspace{-3mm}
\end{figure}

\begin{figure}[!t]
\subfigure[Before finetuning]{     
\centering
\includegraphics[width=0.225\textwidth, trim=8 0 8 0, clip]{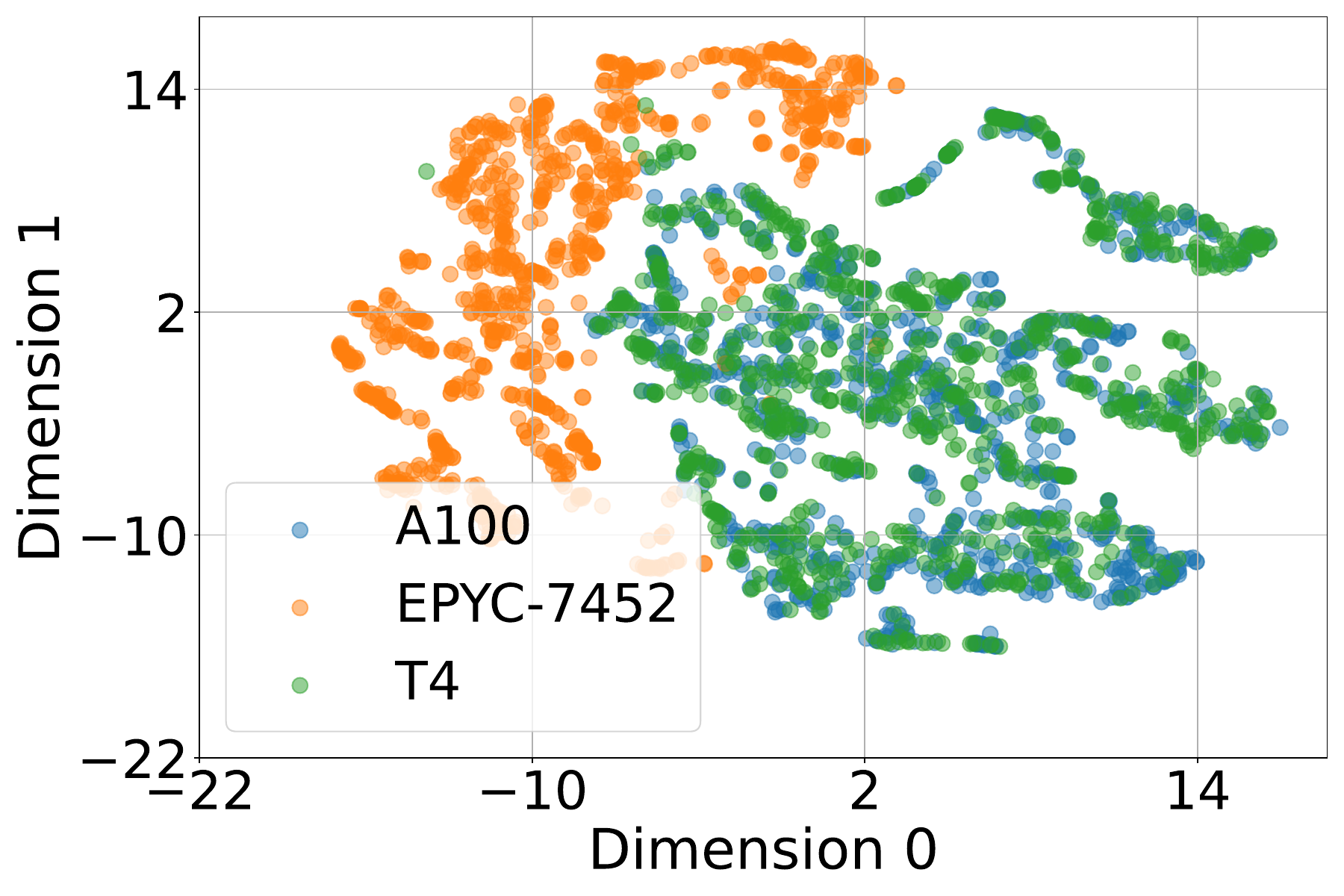}
\label{fig:tir-cd-dist.a}
}
\subfigure[After finetuning]{     
\centering
\includegraphics[width=0.225\textwidth, trim=8 0 8 0, clip]{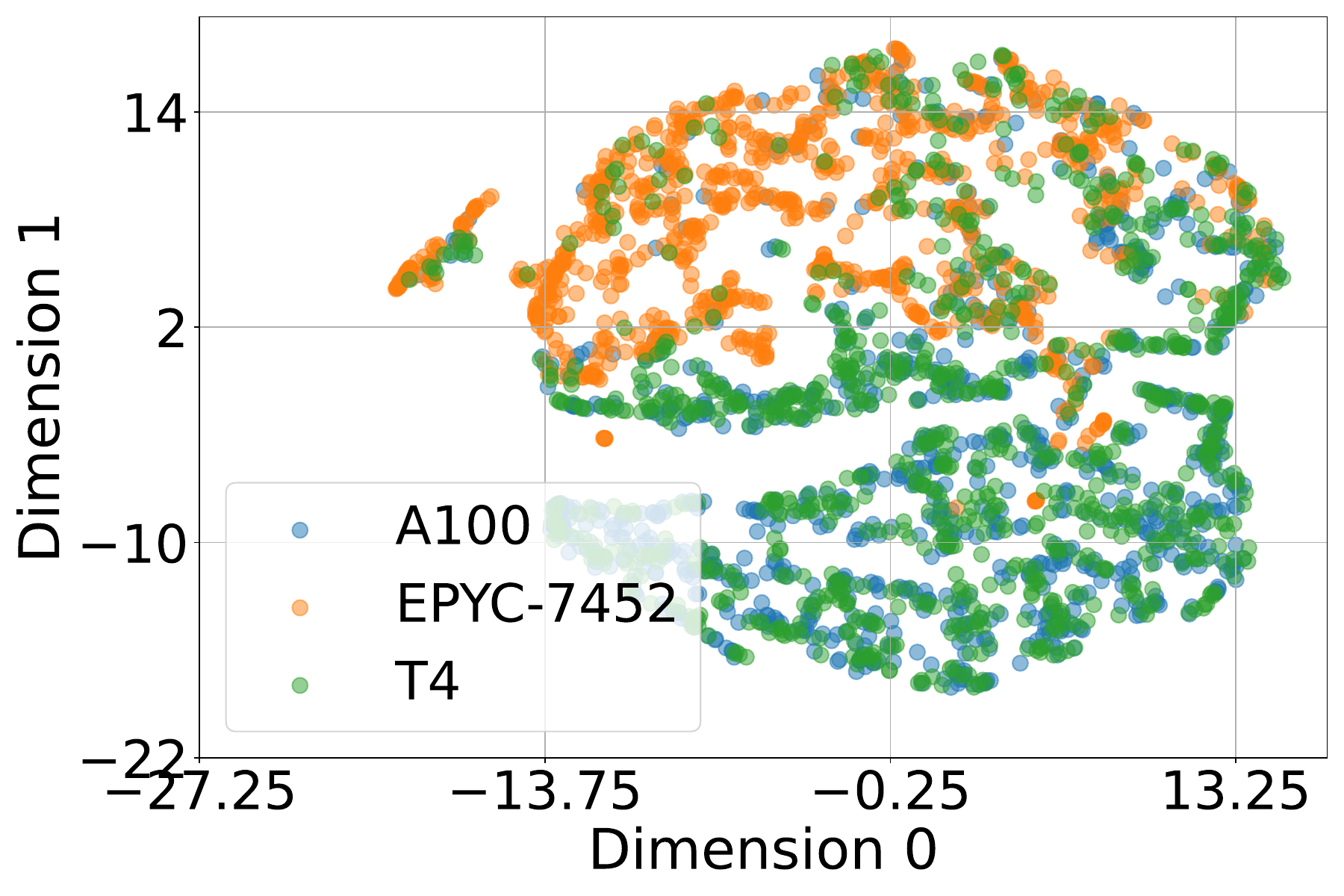}
}
\vspace{-2mm}
\caption{Hidden representation comparison before and after fine-tuning for CDPP. Target device: EPYC.
% \Hu{check the large errors of TLP} \cwu{change Target Devices to Target Device}
\label{fig:tir-cd-dist.b}
}
\label{fig:tir-cd-dist}
\vspace{-3mm}
\end{figure}

We evaluate the cross-device prediction performance %of our predictor 
under 3 different combinations of source and target devices: 1) one GPU as target device and the remaining GPUs as source devices (from GPUs %(\cwu{clarify if the pre-training is done on all other GPUs}) 
to a GPU); 2) one CPU as the target device and the remaining GPUs and CPUs as source devices (from GPUs and CPUs % (\cwu{clarify if pretraining is done on all GPUs and other CPUs}) 
to a CPU; 3) the inference accelerator as the target device and all GPUs as source devices (from GPUs %(\cwu{clarify if pretraining is done on all GPUs}) 
to the inference accelerator). Fig.~\ref{fig:tir-cd} shows that our fine-tuning-based method achieves the lowest prediction error, $10.85\%$ on average. TLP exhibits a large prediction error on absolute time prediction (it focuses on relative time prediction). Habitat uses simple MLPs to predict the performance of the most ``important'' operators (conv2d, lstm, bmm and linear) with operator-level features, while in our case, one operator with a specific shape may have different tensor programs when different scheduling is applied. Habitat's MLP-based cost model struggles to differentiate between these distinct tensor programs lowered from the same operators and has limited generalization to different operator types.
In contrast, we exploit the internal structure of tensor programs to boost the learning of common representations across different tensor programs derived from different operators. % \cwu{briefly what we achieve beyond Habitat}.
% taking two distinct devices as the source and target device. First, we pre-trained the predictor using data from the source device. Then, we fine-tuned it using input features from the target device, 
% without the need for direct access to the target device. 
% Finally, we use the predictor to estimate the latency of tensor programs in a DNN model on the target device
% and feed this information into the replayer to evaluate the end-to-end performance of the DNN model on the target device.
%Fig.~\ref{fig:end2end-cd-habana} \Hu{missing fig} 
Fig.~\ref{fig:tir-cd.b} exhibits that %generalizability of 
our predictor can generalize well from GPU devices to non-GPU devices. We do not have the results of Habitat here as it only supports GPU devices. %The results indicate that our predictor is able to generalize to non-GPU devices from GPU devices.
% albeit with a relatively large average error of $61.24 \%$.  Predicting performance from GPU devices to unseen non-GPU devices is still challenging, particularly when we do not have access to those devices for profiling. This is due to the fundamental differences in hardware architecture between GPUs and non-GPUs devices. We plan to investigate ways to further reduce the prediction error across devices with varying architectures in future work.
Taking the prediction experiment onto EPYC for instance, Fig.~\ref{fig:tir-cd-dist} shows the latent representation of different devices before and after finetuning, where t-SNE is also applied. The results indicate that our method effectively reduces the distribution shift between two GPUs, as well as between GPUs and CPUs.  

\begin{figure}[!t]
\centering
\includegraphics[width=0.45\textwidth, trim=0 50 100 0, clip]{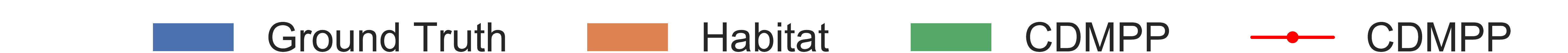}
\subfigure[Prediction onto P100]{     
\centering
\includegraphics[width=0.225\textwidth, trim=8 0 8 0, clip]{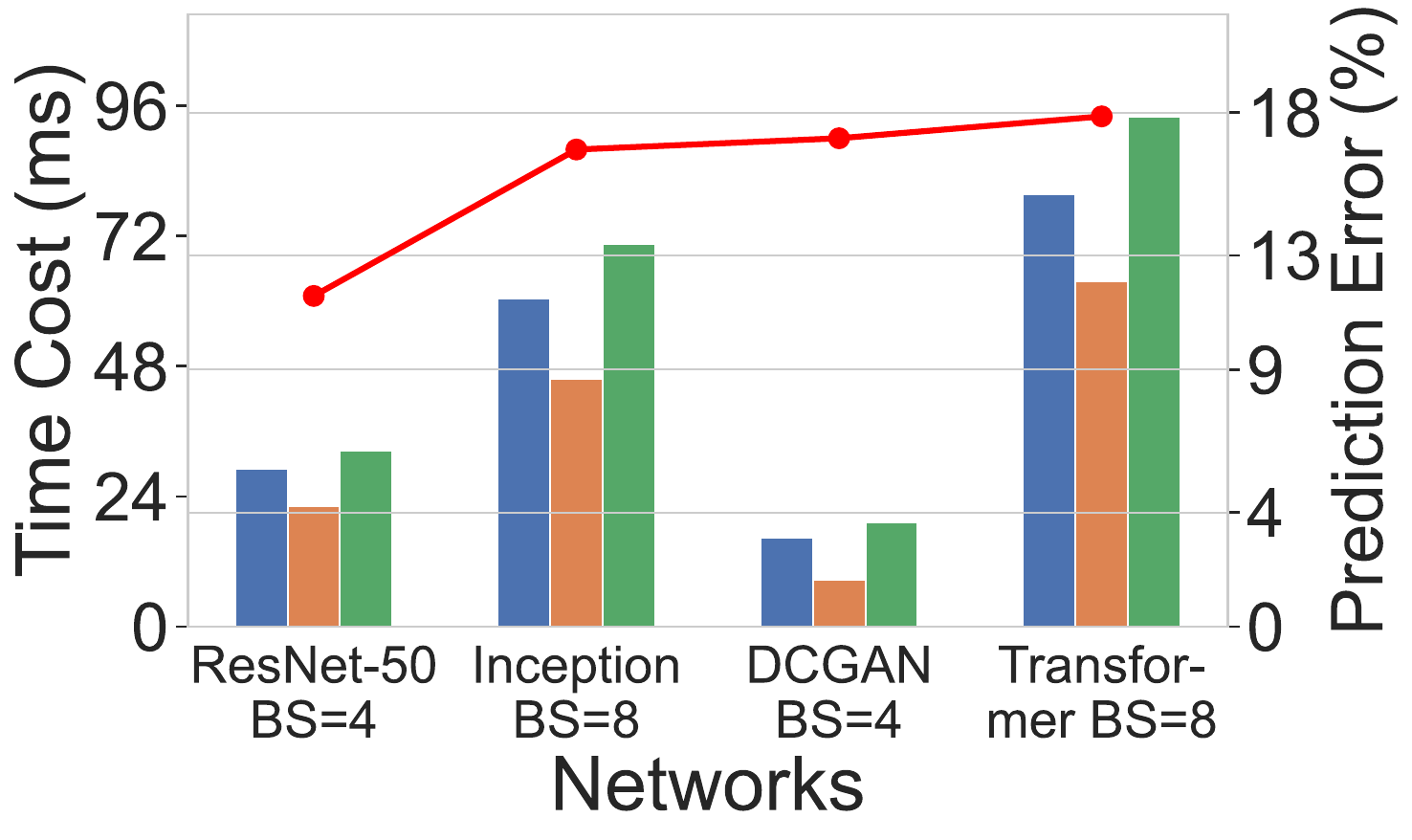}
}
\subfigure[Prediction onto V100]{     
\centering
\includegraphics[width=0.225\textwidth, trim=8 0 8 0, clip]{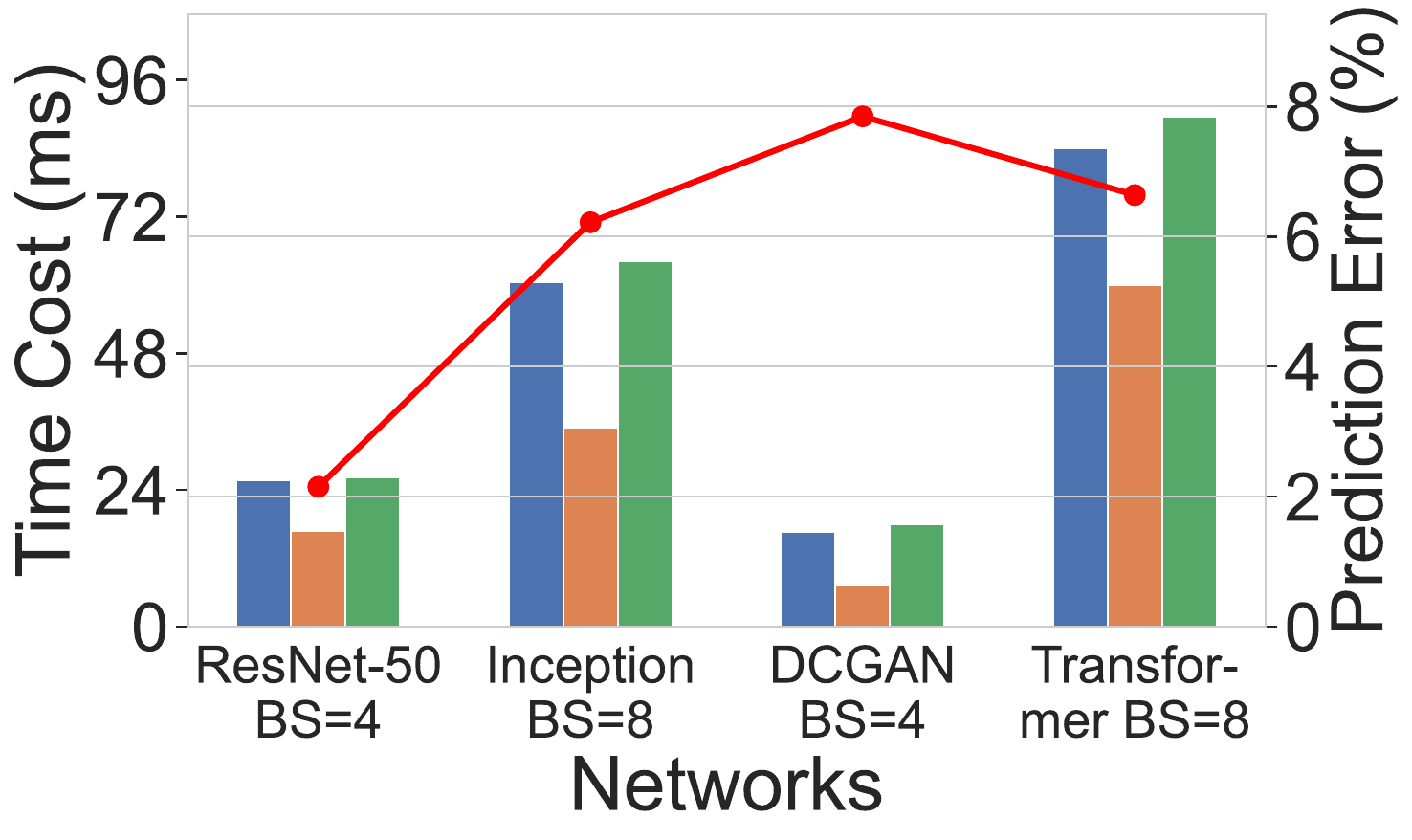}
}
% \subfigure[ $\Rightarrow$ P100]{     
% \centering
% \includegraphics[width=0.225\textwidth, trim=8 0 8 0, clip]{fig/8evaluation/end2end-cd/Source_V100-Target_P100.pdf}
% }
% \subfigure[Prediction onto HL-100]{     
% \centering
% \includegraphics[width=0.225\textwidth, trim=8 0 8 0, clip]{fig/8evaluation/end2end-cd/habana.pdf}
% \label{fig:end2end-cd-habana}
% }
\vspace{-2mm}
\caption{Cross-device end-to-end performance prediction. 
% \Hu{Increase the font size;move to the appendix} 
% \cwu{(a) and (c) are duplicate? HHP: removed one}
} 
\vspace{-3mm}
\label{fig:end2end-cd}
\end{figure}

%To demonstrate the effectiveness of our proposed cross-device learning method on end-to-end performance prediction, 
We further compare cross-device end-to-end model performance prediction among {\sysname}, the ground truth and Habitat in Fig.~\ref{fig:end2end-cd}. %and Fig.~\ref{fig:end2end-cd-habana}. 
Here we do not compare with TLP, since it predicts relative performance of each tensor program which cannot be accumulated as the end-to-end performance. %Fig.~\ref{fig:end2end-cd} shows that 
Our method, {\sysname}, consistently outperforms Habitat with smaller prediction errors in all cases. On average, the prediction error of {\sysname} is $15.72 \%$, and $28.01 \%$ with Habitat. %while the average error of Habitat is $28.01 \%$. 

% \Hu{Explain the large difference between the average error of Habitat claimed by the Habitat paper and that reported in this paper (11.8% VS 28.01%).
% We use the trained cost model from https://github.com/geoffxy/habitat and follow their instructions on different platforms. But we fail to reproduce the results. We are also in touch with the authors to discuss the reproducibility of Habitat.} 

% \Hu{Explain why the proposed method can achieve better results than the baselines. For cross-device learning, }

% The results indicate that by explicitly minimizing the distribution discrepancy between the latent representations from source and target devices and 
% , the predictor can effectively generalize to unseen devices.

% \noindent\textbf{TIR-level cost model.}

% \noindent\textbf{End2end performance prediction.}
% Compared to Habitat, 
% % NNLP (https://xhplus.github.io/publication/2022-icpp-nnlqp/2022-icpp-nnlqp.pdf)
% Estimate the end2end performance of SAME networks on DIFFERENT devices (e.g., T4)

\begin{figure}[!t]
\centering
\includegraphics[width=0.33\textwidth, trim=8 0 8 0, clip]{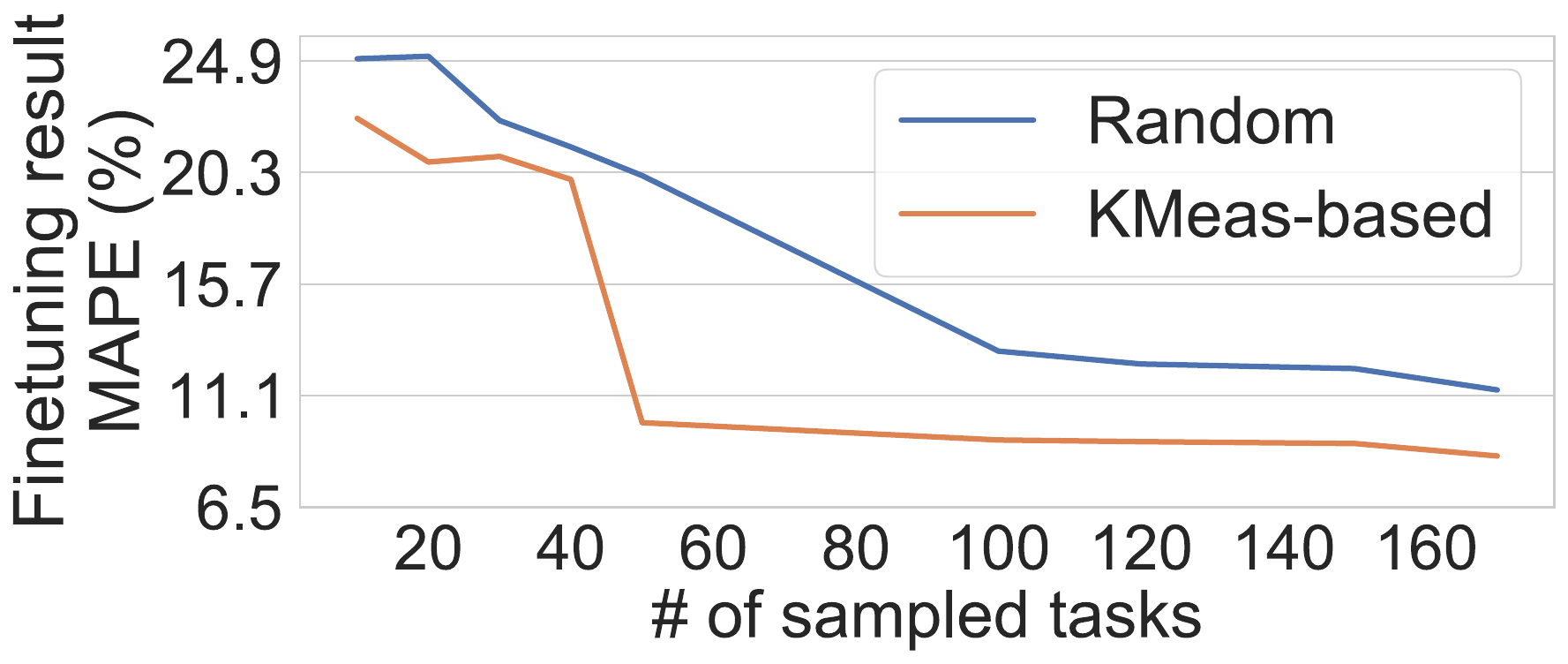}
\vspace{-2mm}
\caption{Effect of sampling strategy for CDPP.
% \cwu{remove (s) from x axis}
}
\label{fig:finetune}
\vspace{-3mm}
\end{figure}

\begin{table}[!t]
    \centering
    \begin{tabular}{l|cccc}
        \toprule
        Device & Box-Cox & Yeo-Johnson & Quantile & original Y \\
        \midrule
        T4 & 15.18 & 49.30 & 17.88 & 72.55 \\
        A100 & 17.53 & 20.09 & 17.38 & 68.77 \\
        K80 & 14.79 & 24.88 & 15.37 & 71.34 \\
        \bottomrule
    \end{tabular}
    \vspace{2mm}
    \caption{MAPE ($\%$) with different normalization  methods.}
    \label{tab:norm-methods}
    % \vspace{-4mm}
\end{table}

% \vspace{-1mm}
\subsection{Effect of sampling strategy in fine-tuning}

%To study the effect of our sampling strategy, we evaluate the fine-tuning results for the GPUs-to-GPU case take T4 as the target device. We vary the number of sampled tasks and 
We compare our KMeans-based sampling strategy to random sampling when fine-tuning the cost model (trained on all other GPUs) on T4. %To remove the experimental bias, we for each experiment, 
We repeat each experiment with random sampling 10 times and report the average results. Fig.~\ref{fig:finetune} shows that when the number of fine-tuning tasks is the same, our sampling method leads to smaller prediction errors, indicating that our KMeans-based sampling strategy can find tasks that better represent the dataset. The error does not decrease % \cwu{why does it slightly increase? HHP: measurement error, re-test the error and update the figure} 
as we increase the number of sampled tasks beyond 50. 50 sampled tasks contain around 100k tensor programs % \cwu{clarify the relation between the 100k tensor programs and 50 sampled tasks}
, which is much smaller than the size of the fine-tuning dataset used in TLP (500K).

% \vspace{-1mm}
\subsection{Ablation Study}
%We then evaluate the impact of each component of {\sysname} on the learning performance with the following ablation experiments. 

% We present these results in Table.~\ref{table:ablation}, which includes the MAPE, training throughput, as well as MSE evaluated on the test dataset collected from T4 for cross-model learning. The first row shows the results for our default setting with all components enabled, and the following rows show the results when individual components are removed in addition to the previous one(s), while the remaining configuration remains the same. 

% Finally, we also observe that applying those components has little effect on training efficiency.

\noindent{\bf %Effect of the 
Normalization method.} %\cwu{the term normalization is not used earlier where we discuss the transformations. HHP: discussed in Sec 5.4} 
Table~\ref{tab:norm-methods} compares %the effect on the 
performance of the pre-trained cost model for cross-model learning when different normalization methods are applied to the target labels. We perform the experiment on 3 devices.
% \cwu{clarify what the devices in the table means, the target device?}. %Overall speaking, 
Box-Cox transformation leads to the smallest test errors, by making the obviously skewed data distribution in our dataset more normal. %We also observe 
Without any normalization, the predictor tends to output a value around the mean cost for any inputs, leading to large prediction errors.

\begin{table}[!t]
\centering
\begin{tabular}{c|c|c|c|c}
\toprule
Device & MSE      & MAPE     & MSPE       & MSE+MAPE  \\
\midrule
T4     & 20.69 & 30.74 & 49.47 & 15.18  \\
A100   & 20.47 & 25.15 & 49.44 & 17.53  \\
K80    & 17.63 & 28.96 & 212.81 & 14.79  \\
 \bottomrule
\end{tabular}
\vspace{2mm}
\caption{MAPE ($\%$) with different loss functions}
\label{tab:lossfuncs_mape}
% \vspace{-5mm}
\end{table}

% \begin{table}[h]
% \centering
% \begin{tabular}{c|c|c}
% \toprule       
% Device &      w/ PE  &   w/o PE \\
% \midrule
% T4   &    15.18 & 18.25 \\
% A100 &    17.53 & 18.82 \\
% K80  &   14.79 & 15.71 \\
% \bottomrule
% \end{tabular}
% \caption{MAPE ($\%$) w/ and w/o PE}
% \label{tab:pe}
% \vspace{-5mm}
% \end{table}

% \begin{table}[h]
% \centering
% \begin{tabular}{c|c|c|c}
% \toprule       
% Batch Size & XGB  & TLP & Ours \\
% \midrule
% % 1   &    0.54 & 0.254 & 0.39 \\
% 4 &   0.74  & 1.005 & 0.57 \\
% 8  &  1.08 & 1.97 & 0.74 \\
% \bottomrule
% \end{tabular}
% \caption{End-to-end latency (ms) after schedule search with different cost models for BERT-tiny on T4.}
% % \Hu{check the resutls of BS=1; more models and devices}
% \label{tab:sche_search}
% \vspace{-5mm}
% \end{table}
% % \subsection{Some Applications}
% % (TODO) Autotune, compare to cost model in AutoTVM/Ansor

\vspace{1mm}
%\subsubsection
\noindent{\bf %Effect of 
Scale-insensitive loss function.} 
We then evaluate the effect of different loss functions by running cross-model learning tasks on different devices. We measure %the error in terms of 
both MAPE (Table~\ref{tab:lossfuncs_mape}) and RMSE (Table~\ref{tab:lossfuncs_rmse}).
% \cwu{tell settings of this experiment}. 
% Here, the only difference between 
MSPE (Mean Square Percentage Error) % and MAPE is that MSPE 
also measures the relative error by summing up the square of the relative error.
We empirically observe that 1) taking MSE as the objective function tends to produce values close to the mean of the real values, leading to a large relative error (MAPE) for samples far away from the mean; 2) taking MAPE and MSPE as the objective function makes the predictor prefer relatively small predictions since under-estimating large values will not incur significantly large error as over-estimating small values (larger than $100\%$).
In contrast, as shown in Table.~\ref{tab:lossfuncs_mape} and Table.~\ref{tab:lossfuncs_rmse},
the scale-insensitive loss function, which explicitly optimizes the absolute and relative error, outperforms any other methods in terms of both MAPE and RMSE.
% , boosting the learning even with large range of dataet
% \Hu{discuss why our method is better}

\begin{figure}[!t]
% \vspace{-4mm}
\subfigure[]{     
\centering
\includegraphics[width=0.225\textwidth, trim=8 0 8 0, clip]{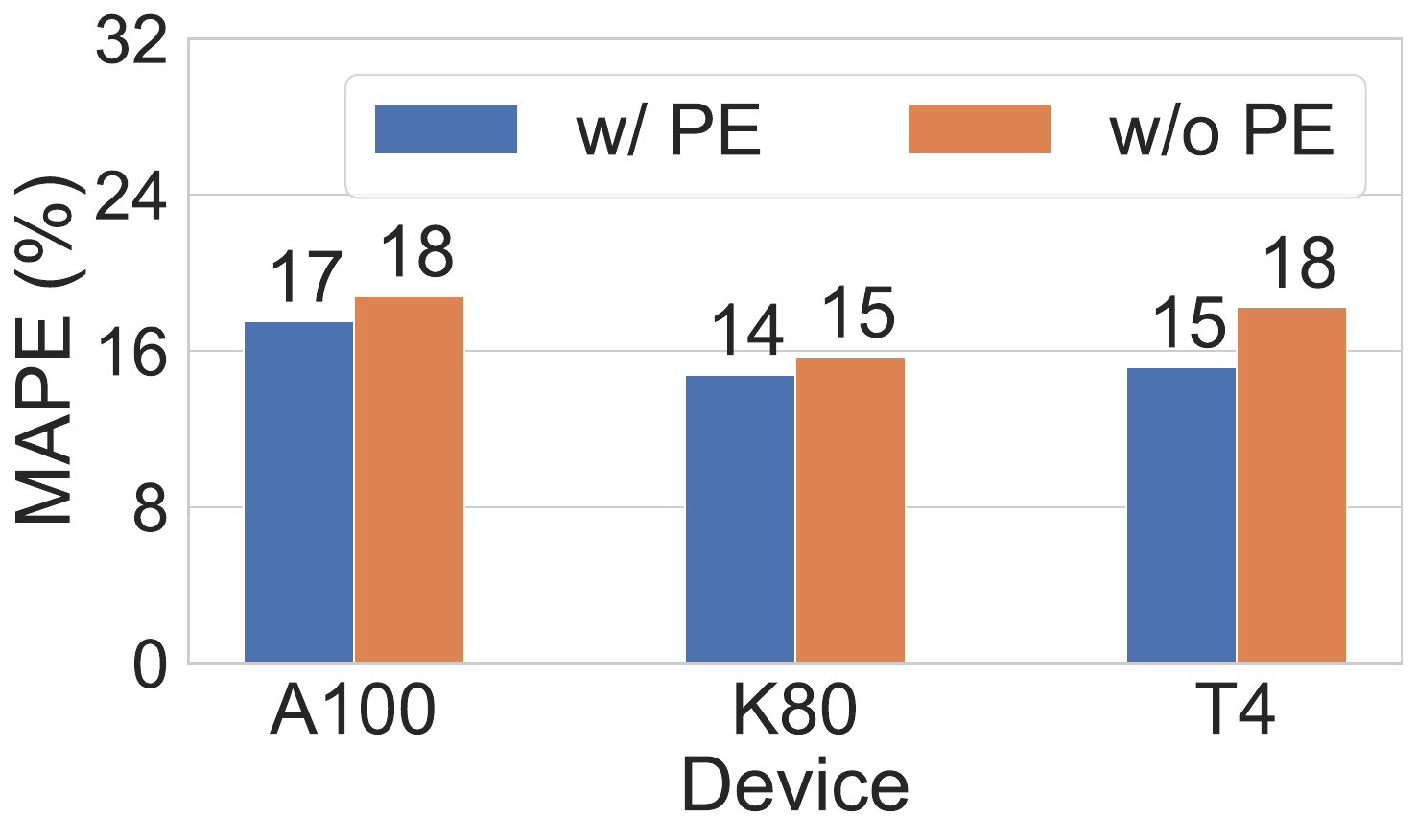}
\label{fig:pe-effect}
}
\subfigure[]{     
\centering
\includegraphics[width=0.225\textwidth, trim=8 0 8 0, clip]{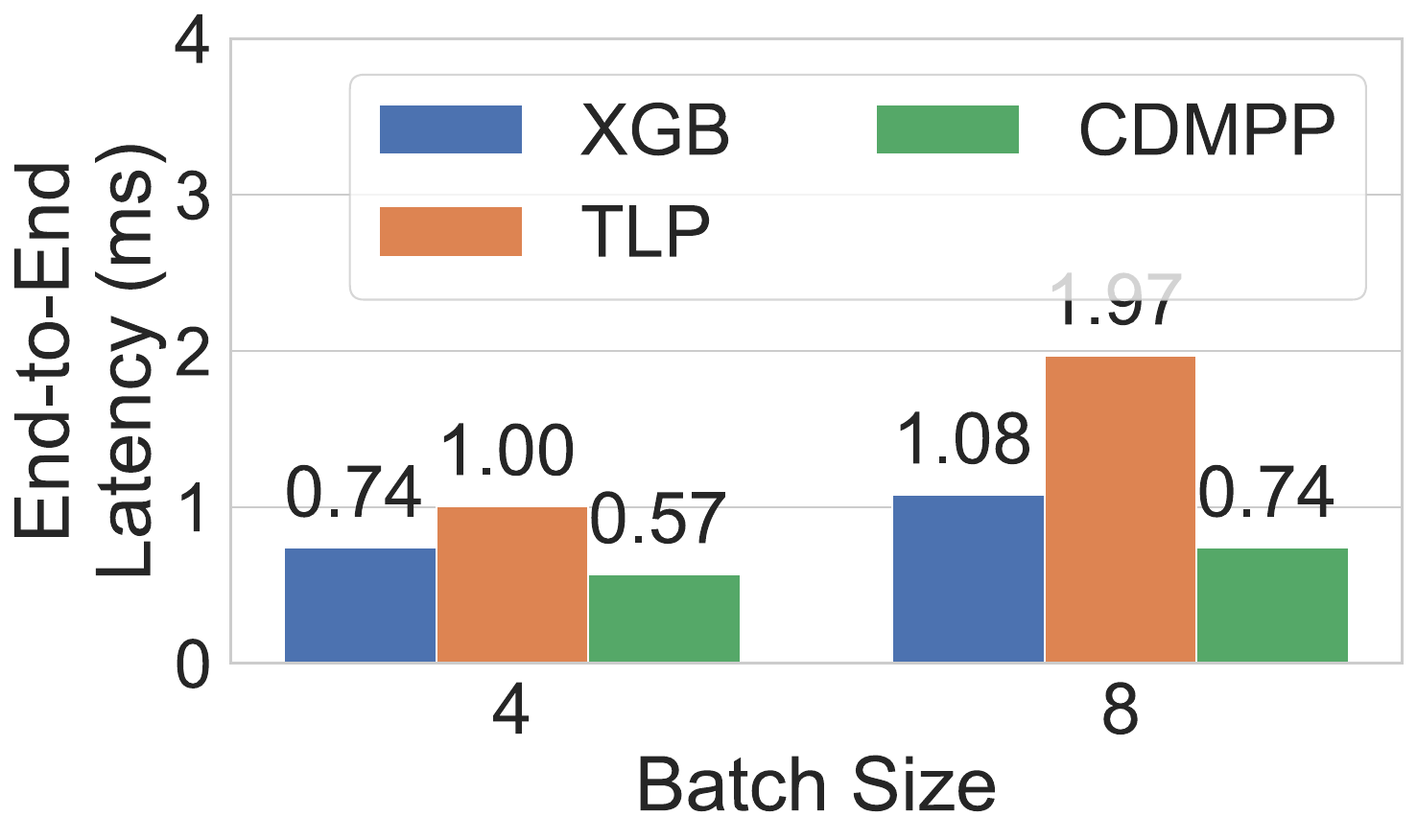}
\label{fig:sche_search-effect}
}
\vspace{-2mm}
\caption{\small (a) MAPE ($\%$) w/ and w/o PE; (b) Schedule search results with different cost models for BERT-tiny on T4. 
% \Hu{Mention the cost to achieve the final latency}
}
\vspace{-3mm}
\end{figure}

\vspace{1mm}
%\subsubsection
\noindent{\bf%Effect of 
Positional encoding.}
Fig.~\ref{fig:pe-effect} compares the prediction errors with and without our customized positional encoding for feature generation. The results prove that positional encoding can reduce the prediction error, revealing that encoding location information of leaf nodes can indeed help the predictor capture the relationship between tensor programs and their performance better.

\vspace{1mm}
%\subsubsection
\noindent{\bf Schedule search.}
We also integrate our cost model into the auto-tuning framework in Ansor to evaluate whether it can identify better schedules. We tune a  DNN network % \cwu{what DNN}
, BERT-tiny, for 2000 search rounds, and the cost model is used to prune the search space in each search round. %We compare the search results in 
Fig.~\ref{fig:sche_search-effect} shows that using our cost model can help find better schedules.
{\sysname}’s inference time on V100 is ~8 ms across batch sizes from 1 to 400, higher than XGBoost’s 0.2 ms. However, our schedule search experiment reveals that {\sysname} can find better schedules while the time ratio for completing 2000 search rounds between {\sysname} and XGB varies from 1.5:1 to 2:1. This smaller gap is primarily because Ansor’s algorithm requires performance measuring of selected candidates (e.g., 10 candidates per search round) on real devices, which incurs significant overhead, alleviating the impact of cost model latency.

\begin{table}[!t]
\centering
\begin{tabular}{c|c|c|c|c}
\toprule
Device & MSE      & MAPE     & MSPE       & MSE+MAPE  \\
\midrule
T4     & 0.31 & 0.48 & 0.62 & 0.34  \\
A100   & 0.30 & 0.40 & 0.41 & 0.27  \\
K80    & 0.87 & 0.89 & 1.5 & 0.59 \\
 \bottomrule
\end{tabular}
\vspace{2mm}
\caption{RMSE ($ms$) with different loss functions}
\label{tab:lossfuncs_rmse}
% \vspace{-5mm}
\end{table}

% \Hu{show that our method can find a better schedule faster. x-axis: search time, y-axis: latency}

% \Hu{To further investigate why our method performs better, compare the relative metric, e.g., top-k accuracy}
% \vspace{-2mm}
\subsection{Discussion}
\noindent \textbf{Extend {\sysname} to different DNN models.} {\sysname} is capable of accurately predicting the latency of various DNN models. When presented with a new DNN model, we employ a feature extraction process from its corresponding tensor programs and leverage the pre-trained cost model, which has been trained on datasets collected from multiple DNN models, for performance prediction. To enhance the generalization capability to new DNN models, we employ a cross-model fine-tuning method, as depicted in Equation ~\ref{eq:cmd_loss}, by taking advantage of the dataset distribution specific to the target DNN model. This fine-tuning process aids in achieving enhanced generalization and adaptability when dealing with previously unseen DNN models.

\vspace{1mm}
\noindent \textbf{Extend {\sysname} to more devices.} {\sysname} is not limited to cross-GPUs only. Given a cost model pre-trained on the GPU dataset, our fine-tuning approach will utilize the clustering results on the available dataset to guide trace collection on target devices, including non-GPU architectures, and adapt the cost model to the new device fast in 10-20 minutes. It never requires re-training from scratch. Fig.~\ref{fig:tir-cd.b} shows examples of prediction from GPUs to Habana HL-100 and CPUs of different brands (e.g., Intel, ARM, AMD).

%% file: 9related_work.tex
%%%%%%%%%%%%%%%%%%%%%%%%% Related Work %%%%%%%%%%%%%%%%%%%%%%%%%
\vspace{-1mm}
\section{Related Work}

\mypara{Cross-device Performance Prediction.}
%There exists some prior work on cross-device performance modeling. 
% For example, Paleo~\cite{qi2016paleo} estimates the performance of DNN operators (ops) with an analytical model based on the number of floating points (FLOPs) in ops. However, this approach is limited because
Habitat~\cite{geoffrey2021habitat} proposes a roofline-model-based~\cite{williams2009roofline} scaling method and MLP-based model to predict performance of ops across different GPUs. %according to whether ops are implemented with the same CUDA kernel on the source and target devices. However, it only focuses on GPUs, while more and more devices from various vendors (e.g., Habana Goya~\cite{habana2019goya} ) are used for DL jobs in the industry, especially for inference jobs. 
% Second, the approach requires users to actually have access to the target GPU to profile what kind of kernel is used on the target device, which is not applicable
% does not does not describe how to predict the CUDA kernel used in the target device
% when the target device is not available.
TLP~\cite{zhai2023tlp} extracts features from schedule primitives, instead of tensor programs, and maintains a prediction head for each device. %However, TLP focus on predicting the speedup of a tensor program against the one without any schedules applied. 
nn-Meter~\cite{zhang2021nn_Meter} builds a kernel-level latency predictor for model inference on diverse edge devices, but only focuses on CNNs. %and the proposed adaptive data sampling requires data profiling on the target device.
%Moses~\cite{zhao2022moses} achieves efficient cross-device adaption by automatically identifying the transferable model parameters across devices, but it relies on fine-tuning on the target device.
NNLQP~\cite{liu2022nnlqp} %directly 
estimates the iteration time of DNN models with a device-independent GNN-based encoder and device-specific prediction heads. %Both Moses and NNLQP doesn't provide a solution for trace collection to adapt to a new device.

\vspace{1mm}
\mypara{Cross-model Performance Prediction.}
%DNN optimizations often need to query the performance improvement of optimization on a specific device~\cite{hu2022dpro, jia2019optimizing, cai2018proxylessnas, chen2018tvm, dudziak2020brp, kaufman2019learned, kaufman2021learned, zhang2021nn_Meter}. Therefore, extensive research has been developed to perform cross-model performance prediction. 
Many works~\cite{chen2018learning, chen2018tvm, zheng2020ansor} maintain a cost model for each kind of DNN op (Conv2d, Matmul, etc.) or kernel (a subgraph of DNN) separately, which is not scalable considering the variety of DNN models and tensor programs. %Although 
AutoTVM~\cite{chen2018tvm} and Ansor~\cite{zheng2020ansor} exploit transfer learning to avoid training cost models for each kernel from scratch, while the fine-tuning process for each kernel is still time-consuming. %and makes developers torn between better search results and smaller optimization time. 
Besides, they only predict the relative order among optimization candidates and do not provide accurate cost estimation. 
Kaufman et al.~\cite{kaufman2021learned} decompose a DNN model into computation subgraphs and use a GNN to predict the performance of each subgraph on TPUs~\cite{you2019fast_tpu}.
% But it only focuses on XLA programs on Tensor Processing Units (TPUs), having no ability for cross-device prediction
Steiner et al.~\cite{steiner2021value} predict %the expected 
performance of a partial schedule using an LSTM. %over carefully engineered features.
%Both %of them 
%focus on subgraph level predication and are not open-sourced. 
%Both methods focus on directly predicting the performance of a subgraph, where each node represents an operator. Still limited on utilizing the internal structure of tensor programs for each operator.
MetaTune~\cite{ryu2021metatune} and Tiramisu~\cite{baghdadi2021deep_tiramisu} propose AST-based representations of tensor programs to exploit the internal structure of tensor programs. We did not compare with MetaTune since it is not open-sourced.
% \cwu{briefly why we did not compare with MetaTune.} %but fail to provide an efficient learning method to consume large datasets in a feasible time.

% Tiramisu~\cite{baghdadi2021deep_tiramisu} proposes AST (Abstract Syntax Tree) representation of programs. But it only focuses on performance prediction on CPUs. 

%\mypara{Summary.} In summary, none of the previous predictors achieve fine-grained accurate cross-model and cross-device performance prediction concurrently. 
% \Hu{Among all the previous work, NNLQP~\cite{liu2022nnlqp} is the one that most close to our target, but it can not estimate the performance of any tensor program and still requires that the target device is accessible}. 
%In contrast, our proposed method learns domain-invariant representations that can be generalized to various devices and DNN models.

% It is empirically difficult to maintain a model for all ops \Hu{add related works}. . - Some recent works propose extracting tree-like features (or AST) from tensor programs,  e.g., Tiramisu, but we empirically found that it doesn not improve the prediction performance much. \Hu{add related works}

%%%%%%%%%%%%%%%%%%%%%%%%%

% Developing an accurate and efficient performance model over various devices for tensor programs from different DNN models is challenging~\cite{xla_gnn}. Overall speaking, it introduces the following challenges

% \subsection{Metric Learning / Cross-domain prediction}
% Meta Learning xxx

%\subsection
\vspace{1mm}
\noindent\textbf{ML Benchmarking.}
% The first case is to estimate the performance of a set of DNN models on a specific device. 
%Although some 
Benchmark results~\cite{mattson2020mlperf, reddi2020mlperf, zhu2018benchmarking, cuda2020benchmarking} are available %, they are usually limited to
for some specific DNN models (e.g., ResNet50~\cite{he2016resnet}, BERT~\cite{devlin2018bert}) and common GPUs. Some non-GPU vendors may publish their benchmark results~\cite{mlperf}, but are also limited to some common DNN models. %Considering it requires substantial time and effort to manually test all DNN models, 
We propose a system to predict the performance of any given DNN model without extensively running it on a target device, as long as the model can be represented as TIRs. 
% The community has also published some benchmarks (e.g., MLPerf~\cite{mlperf}) to provide evaluations of training and inference performance for various DNN models on different hardware. However, those benchmarks are limited on some common devices (e.g., V100, T4) and DNN models (e.g., BERT, ResNet-50)

%% file: 0appendix.tex
% \newpage
\clearpage
\appendix
\input{appendix/appendix_cluster_proof}

\section{Auto-tuner Design} \label{appendix:auto-tune}
Table.~\ref{tab:auto_tune_rst} lists the detailed values found by our auto-tuner. The cost model has 13.8 million parameters in total.
\begin{table}[h]
\centering
\begin{tabular}{|c|l|c|}
\toprule
Taxonomy & Variable & Value \\
\midrule
\multirow{6}*{\makecell[c]{Model\\Architecture}} & batch size & 600 \\ \cline{2-3}
~ & \makecell[l]{output dimension of\\the input layer} & 716 \\ \cline{2-3}
~ & $\#$ of transformer layers & 11 \\ \cline{2-3}
~ & \makecell[l]{hidden dimension of linear\\layers in the encoder} & 985 $\times$ 3 layers \\ \cline{2-3}
~ & $\#$  of features for embeddings & 69 \\ \cline{2-3}
~ & \makecell[l]{hidden dimension of linear\\layers in the decoder} & 930 $\times$ 3 layers \\ \hline
\multirow{6}*{\makecell[c]{Hyper-\\parameters}}  & learning rate & 1.68e-05 \\ \cline{2-3}
~ & learning rate scheduler & CyclicLR \\ \cline{2-3}
~ & optimizer type & Adam \\ \cline{2-3}
~ & weight decay & 0.0013 \\ \cline{2-3}
~ & dtype & float32 \\ \cline{2-3}
~ & $\alpha$ & 1 \\
\bottomrule
\end{tabular}
\vspace{2mm}
\caption{Autotune Results}
\label{tab:auto_tune_rst}
% \vspace{-5mm}
\end{table}

\section{Replayer Design} \label{appendix:replay}
To achieve end-to-end performance evaluation, the replayer takes the TIR-based Data Flow Graph (DFG) as input and simulates the execution order of the nodes in the TIR-based DFG. The TIR-based DFG contains nodes, each of which represents a tensor program labeled with its execution time, and edges describing the dependencies between each pair of tensor programs. Algorithm~\ref{algo:simulate} shows the detailed procedure to perform topology-sorting on the TIR-based DFG and estimate the end-to-end performance. Given a set of devices $D$, we maintain a priority queue for each device, which stores TIR functions whose predecessors have all been executed. The replayed iteratively fetches a TIR function of each priority queue to execute, updates the timestamps as the completed time of this TIR function, and enqueues successors to the corresponding device queue if necessary. In general, our simulation assumes the execution on GPUs with a single device. However, in scenarios where multiple CUDA streams are utilized, we can allocate multiple devices, with each device representing one stream.

\begin{algorithm}[!th]
\caption{End-to-end Simulation Algorithm}
\label{algo:simulate}
\begin{algorithmic}[1]
\small
\State {\bfseries Input:}  global DFG: $\mathcal{G}(V, E)$, device set $D$
\State {\bfseries Output:}  Iteration Time
\For {$d$ in $D$}
	\State $d.deviceTime \leftarrow 0$ \Comment{initialize $deviceTime$ with 0}
	\State $d.queue \leftarrow []$ \Comment{initialize device frontier with a empty queue}
\EndFor
\For {$u$ in $V$}
	\State $u.ref \leftarrow u.indegree$
	\If {u.ref = 0}
		\State d.queue.enqueue(u)
	\EndIf
\EndFor
\While {True}
	\State d = select(D) \Comment{first device with non-empty d.queue}
	\If{d is None}
		\State stop simulation
	\EndIf
	\State $u \leftarrow d.queue.dequeue(0)$ \Comment{select the op with the smallest readyTime}
	\State $u.start \leftarrow max(d.deviceTime, u.readyTime)$ \Comment{decide the start time}
	\State $d.deviceTime \leftarrow u.start + u.duration + u.gap$ \Comment{update device time}
	\State \textbf{Re-sort $D$ by $deviceTime$} \Comment{$O(log(|D|)$}
	\For {$c \in u.successors$}
		\State $c.ref \leftarrow c.ref-1$
		\State $c.readyTime \leftarrow \max(c.readyTime, u.start + u.duration + u.gap)$
		\If {c.ref = 0}
			\State device(c).queue.enqueue(c) \Comment{Enqueue the successor to the corresponding priority queue.}
		\EndIf
	\EndFor
\EndWhile
\State Iteration Time = $\max([device(c).deviceTime]$ for $c$ in D)
\end{algorithmic}
\end{algorithm}

\section{Appendix: Supplement Experiments}

\subsection{Cross-model fine-tuning}
\begin{figure}[!t]
\subfigure[Cross-model learning on T4]{     
\centering
\includegraphics[width=0.42\textwidth, trim=8 0 8 0, clip]{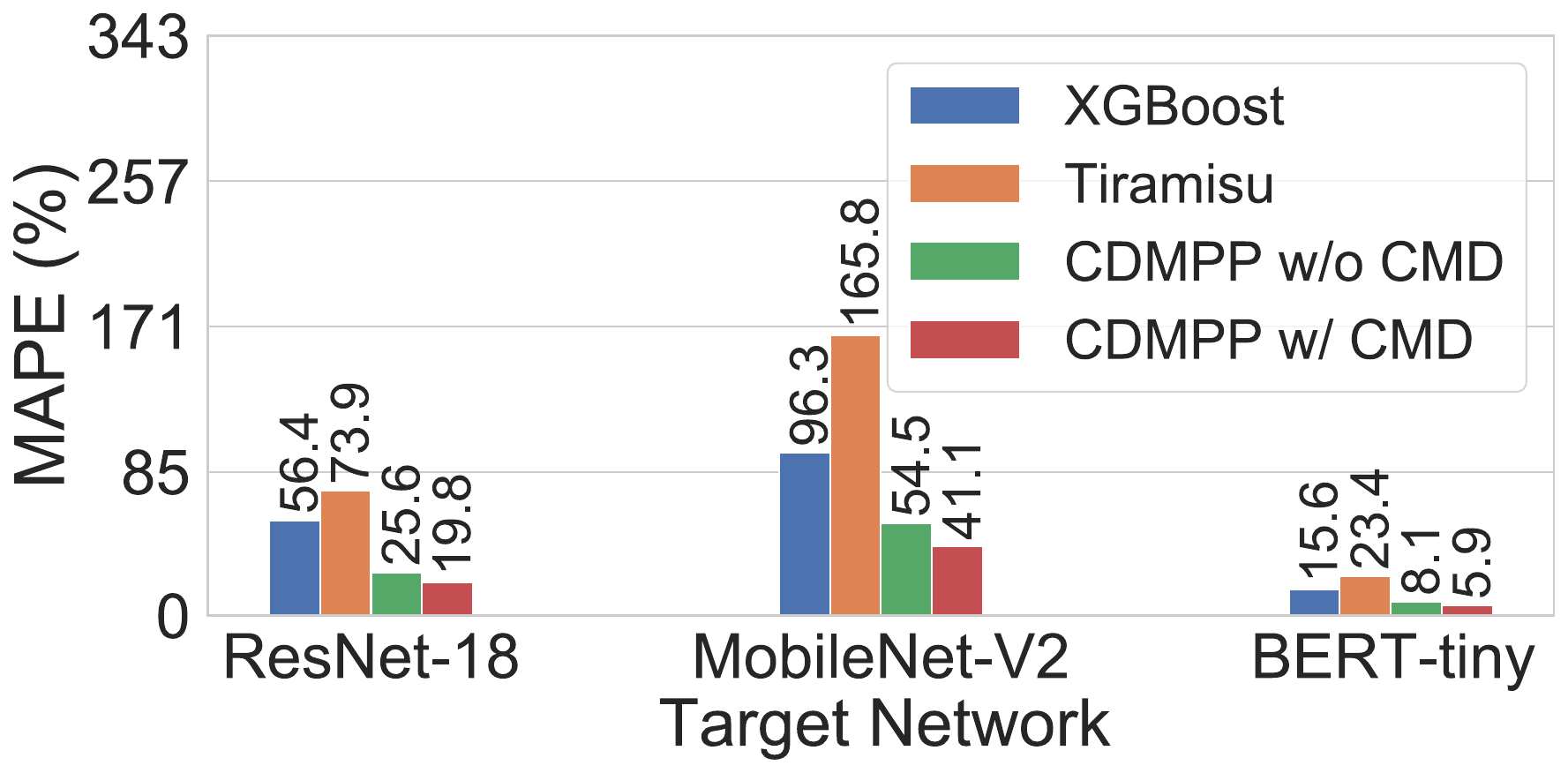}
}
\subfigure[Cross-model learning on EPYC]{     
\centering
\includegraphics[width=0.42\textwidth, trim=8 0 8 0, clip]{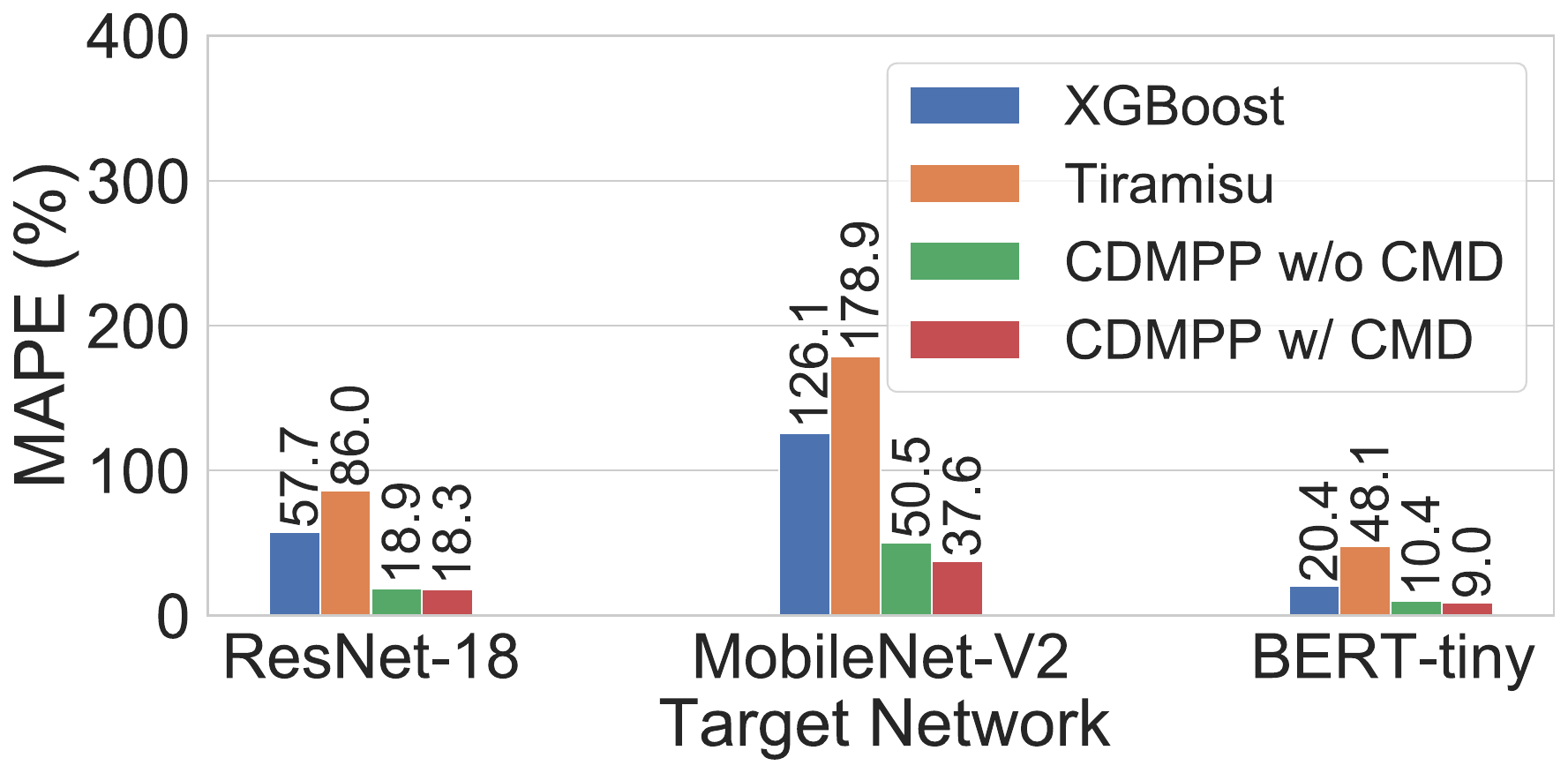}
}
\vspace{-2mm}
\caption{Comparison of cross-model prediction errors. The number on top of each bar is the exact MAPE value.}
\label{fig:cm-finetune-append}
% \vspace{-3mm}
\end{figure}

\begin{figure}[!t]
\subfigure[BERT-tiny, w/o CMD]{     
\centering
\includegraphics[width=0.225\textwidth, trim=8 0 8 0, clip]{fig/8evaluation/0cmpp_finetune/t4_1_200_bert_tiny_dist_visual.pdf}
\label{fig:cm-latent-analysis.a-append}
}
\subfigure[BERT-tiny, w/ CMD]{     
\centering
\includegraphics[width=0.225\textwidth, trim=8 0 8 0, clip]{fig/8evaluation/0cmpp_finetune/t4_1_200_bert_tiny_cmd_dist_visual.pdf}
\label{fig:cm-latent-analysis.b-append}
}
\subfigure[MobileNet-V2, w/o CMD]{     
\centering
\includegraphics[width=0.225\textwidth, trim=8 0 8 0, clip]{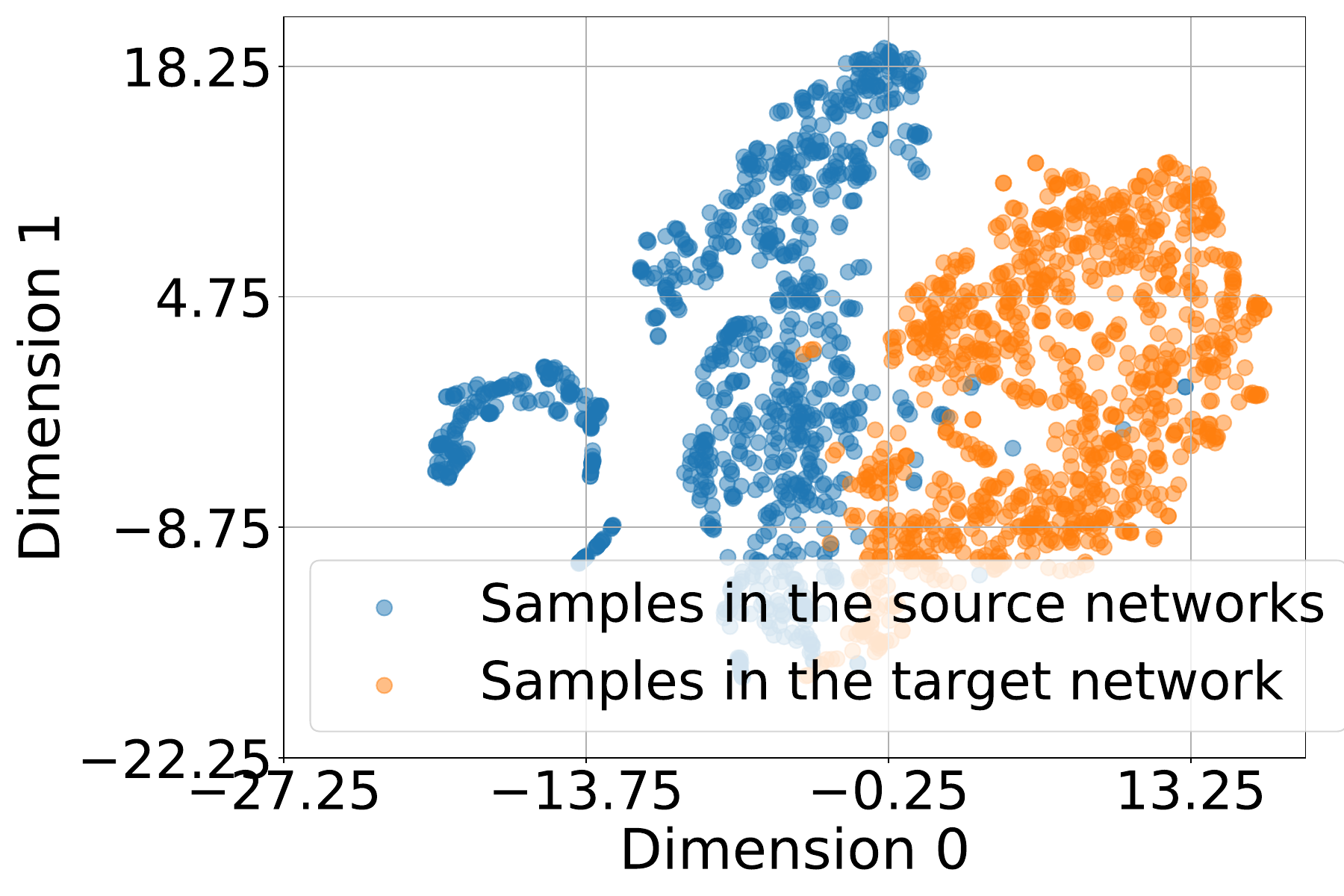}
\label{fig:cm-latent-analysis.c-append}
}
\subfigure[MobileNet-V2, w/ CMD]{     
\centering
\includegraphics[width=0.225\textwidth, trim=8 0 8 0, clip]{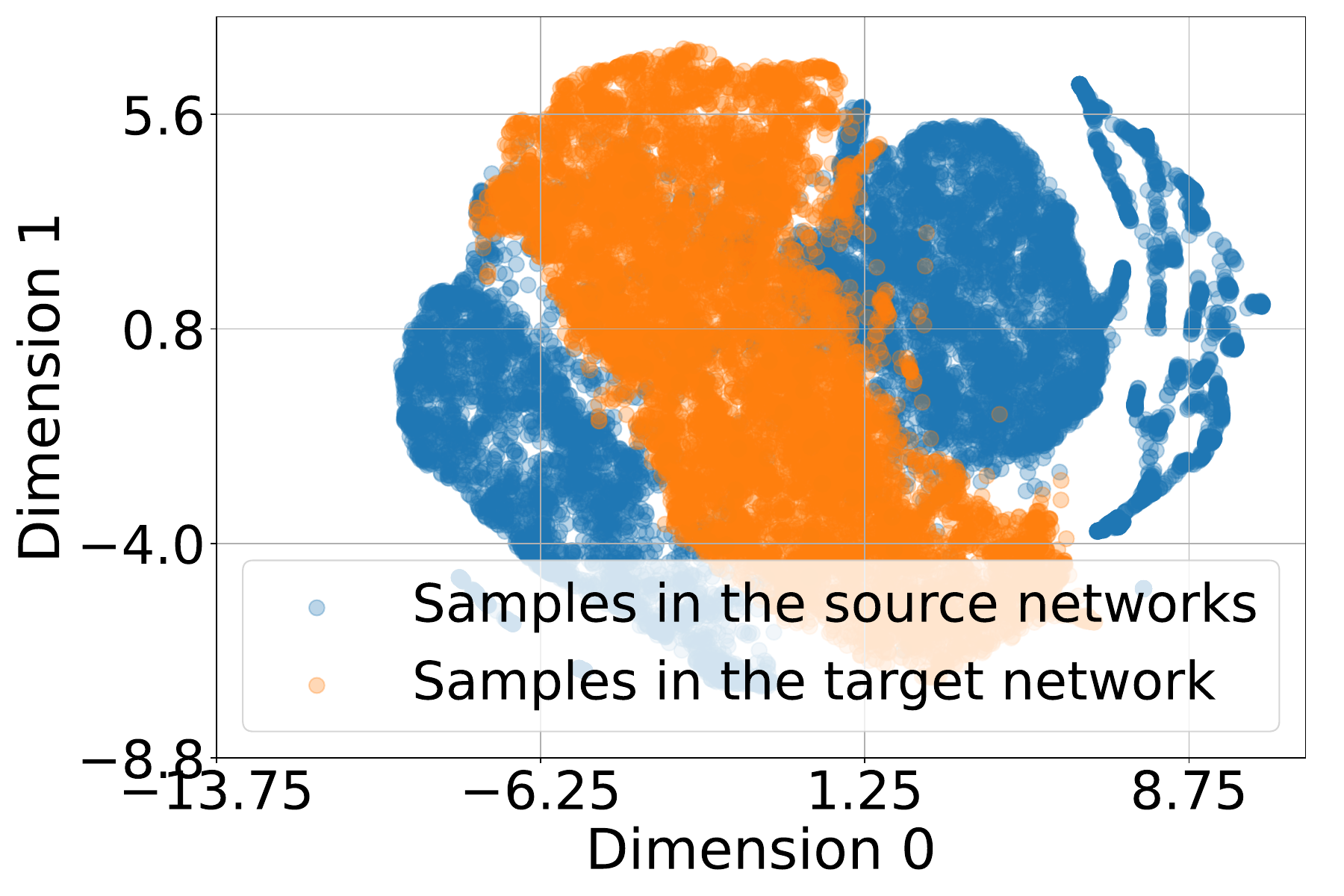}
\label{fig:cm-latent-analysis.d-append}
}
\vspace{-2mm}
\caption{Hidden representation comparison with t-SNE.}
\label{fig:cm-latent-analysis-append}
% \vspace{-3mm}
\end{figure}

Fig.~\ref{fig:cm-finetune-append} shows more results for cross-model fine-tuning. In most cases, {\sysname} can achieve a small prediciton error. However, in some cases, e.g., when taking MobileNet-V2 as the target network, the prediction error is large for all methods, because the distribution shift between the source and target domain is too large for domain adaption technique to mitigate, as shown in Fig.~\ref{fig:cm-latent-analysis.c-append}. In this case, we recommend augmenting the training set to cover more samples.

\subsection{End2end performance for CMPP}
\begin{figure}[!t]
\centering
\includegraphics[width=0.48\textwidth, trim=200 50 100 35, clip]{fig/8evaluation/end2end-cm/legend.pdf}
\subfigure[Predicted onto ResNet50, BS=1]{     
\centering
\includegraphics[width=0.225\textwidth, trim=8 0 8 0, clip]{fig/8evaluation/end2end-cm/ResNet-50_1.pdf}
}
\subfigure[Predicted onto ResNet50, BS=4]{     
\centering
\includegraphics[width=0.225\textwidth, trim=8 0 8 0, clip]{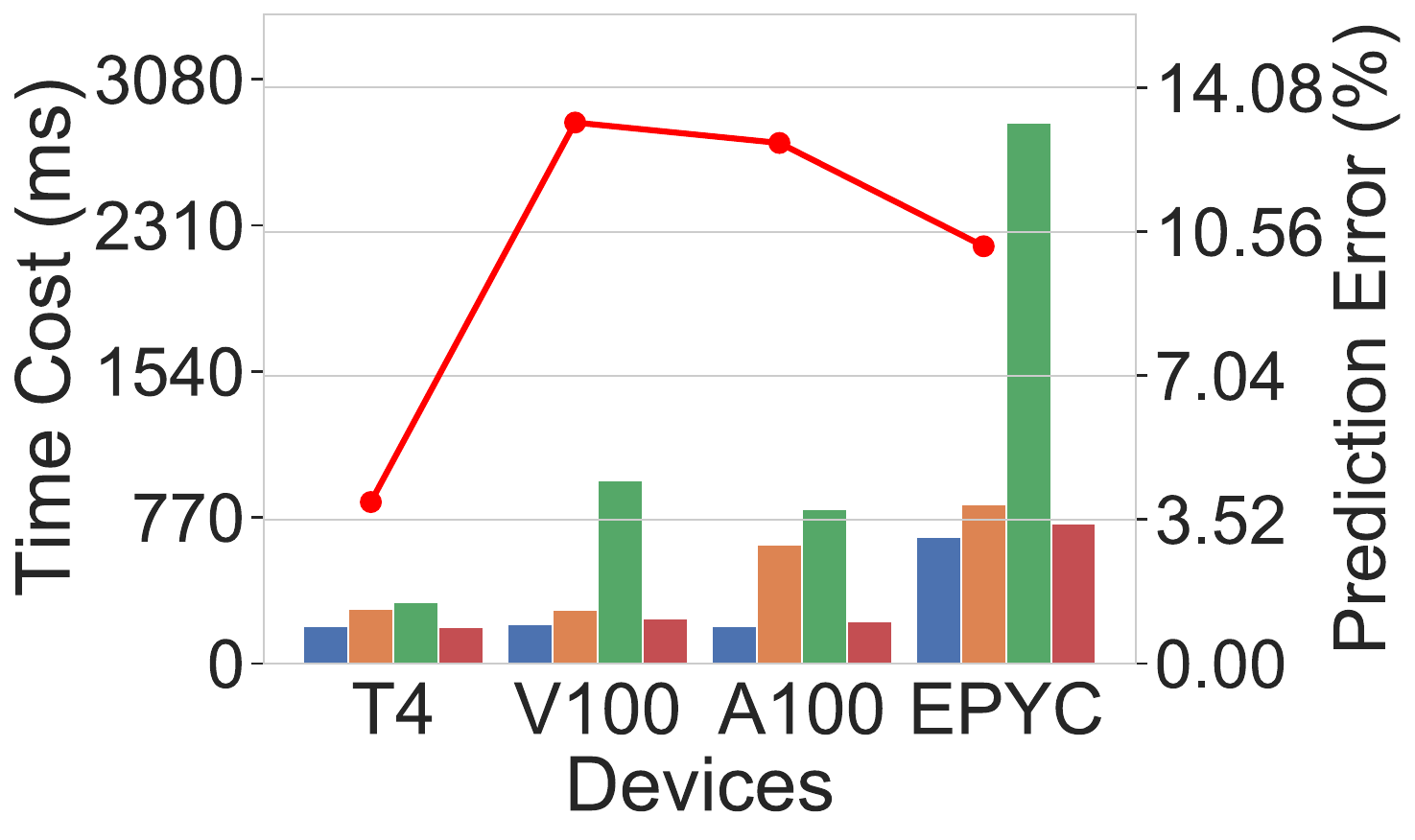}
}
\subfigure[Predicted onto ResNet50, BS=8]{     
\centering
\includegraphics[width=0.225\textwidth, trim=8 0 8 0, clip]{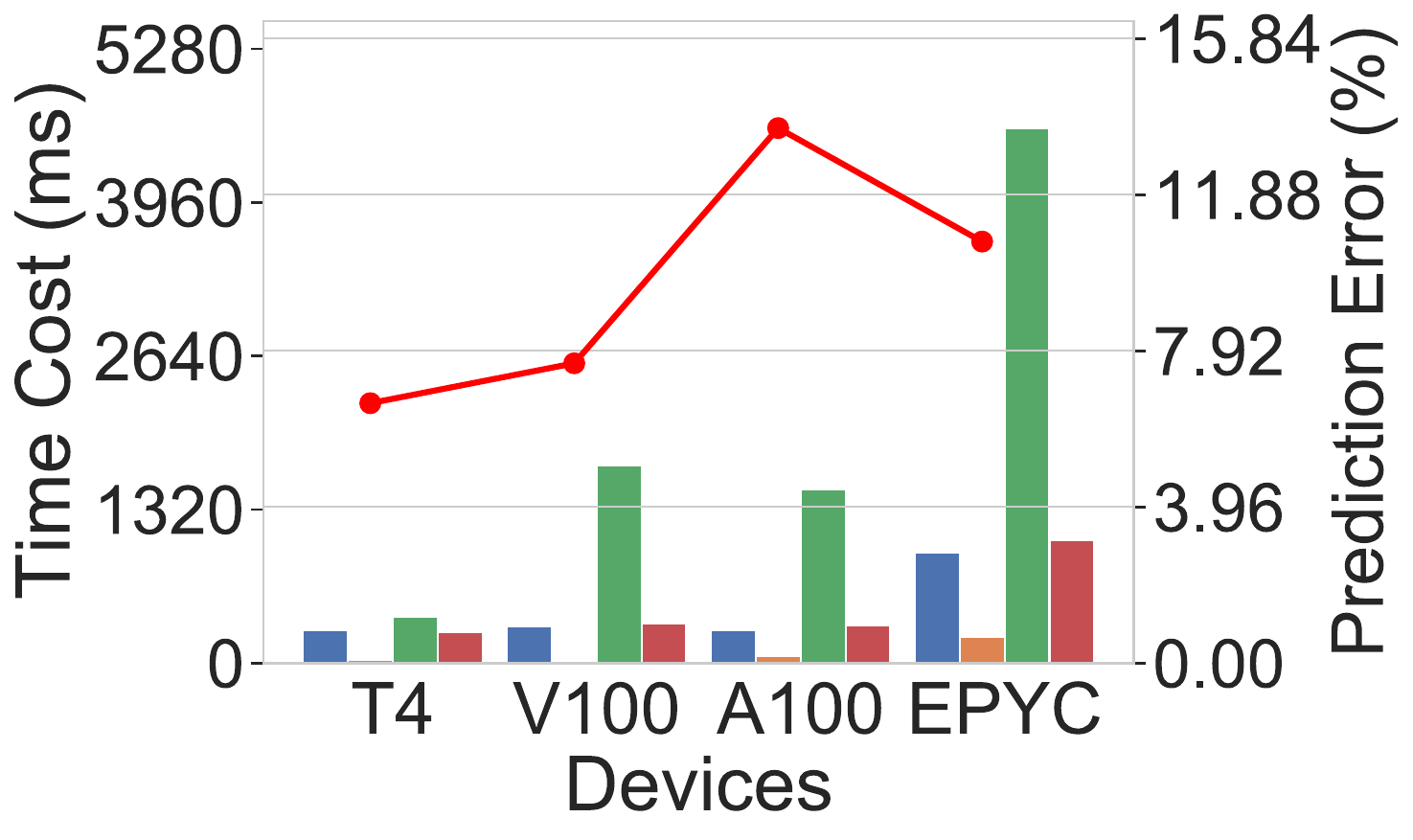}
}
\subfigure[Predicted onto InceptionV3, BS=1]{     
\centering
\includegraphics[width=0.225\textwidth, trim=8 0 8 0, clip]{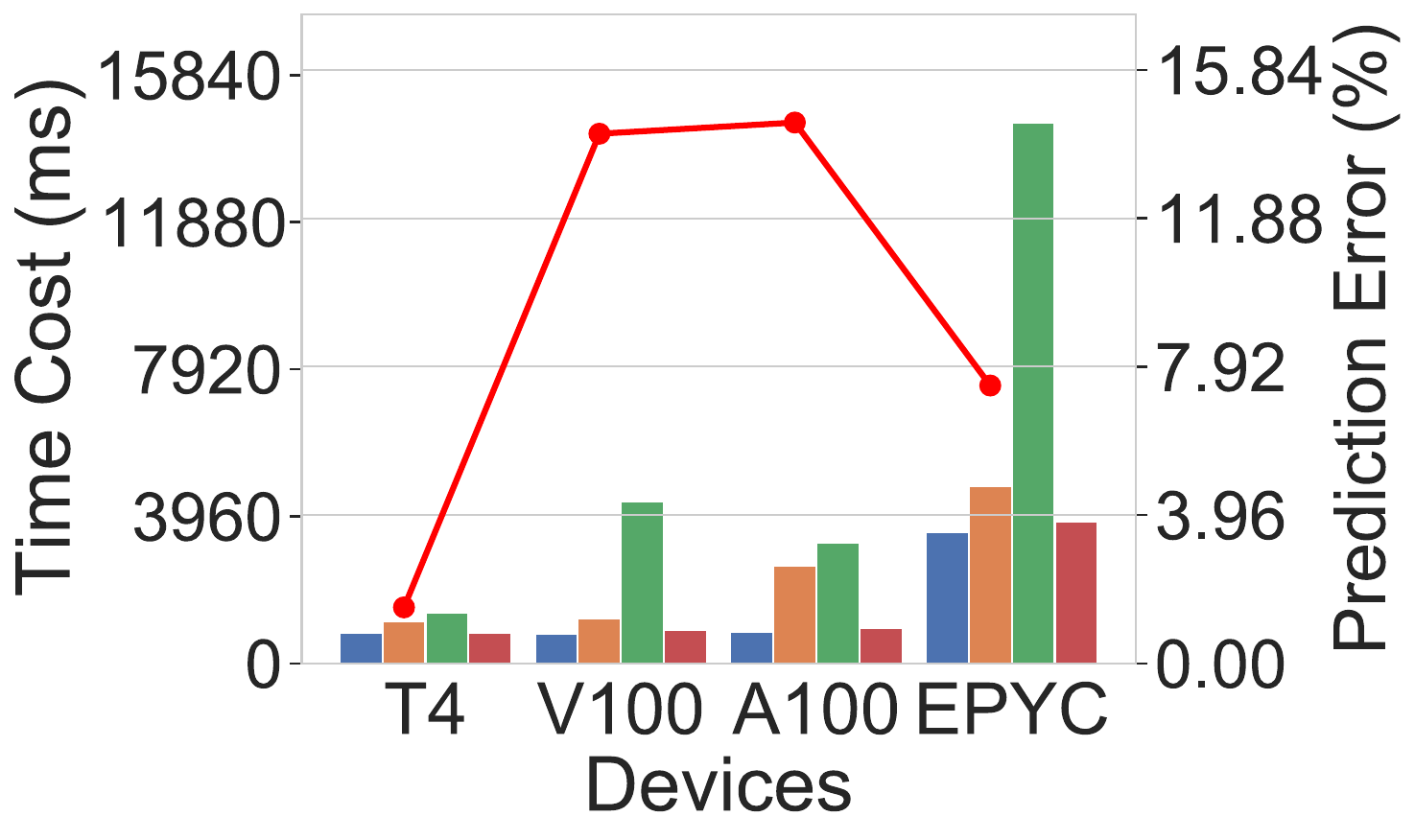}
}
\subfigure[Predicted onto BERT Base, BS=1]{     
\centering
\includegraphics[width=0.225\textwidth, trim=8 0 8 0, clip]{fig/8evaluation/end2end-cm/BERT_Base_1.pdf}
}
\subfigure[Predicted onto BERT Base, BS=4]{     
\centering
\includegraphics[width=0.225\textwidth, trim=8 0 8 0, clip]{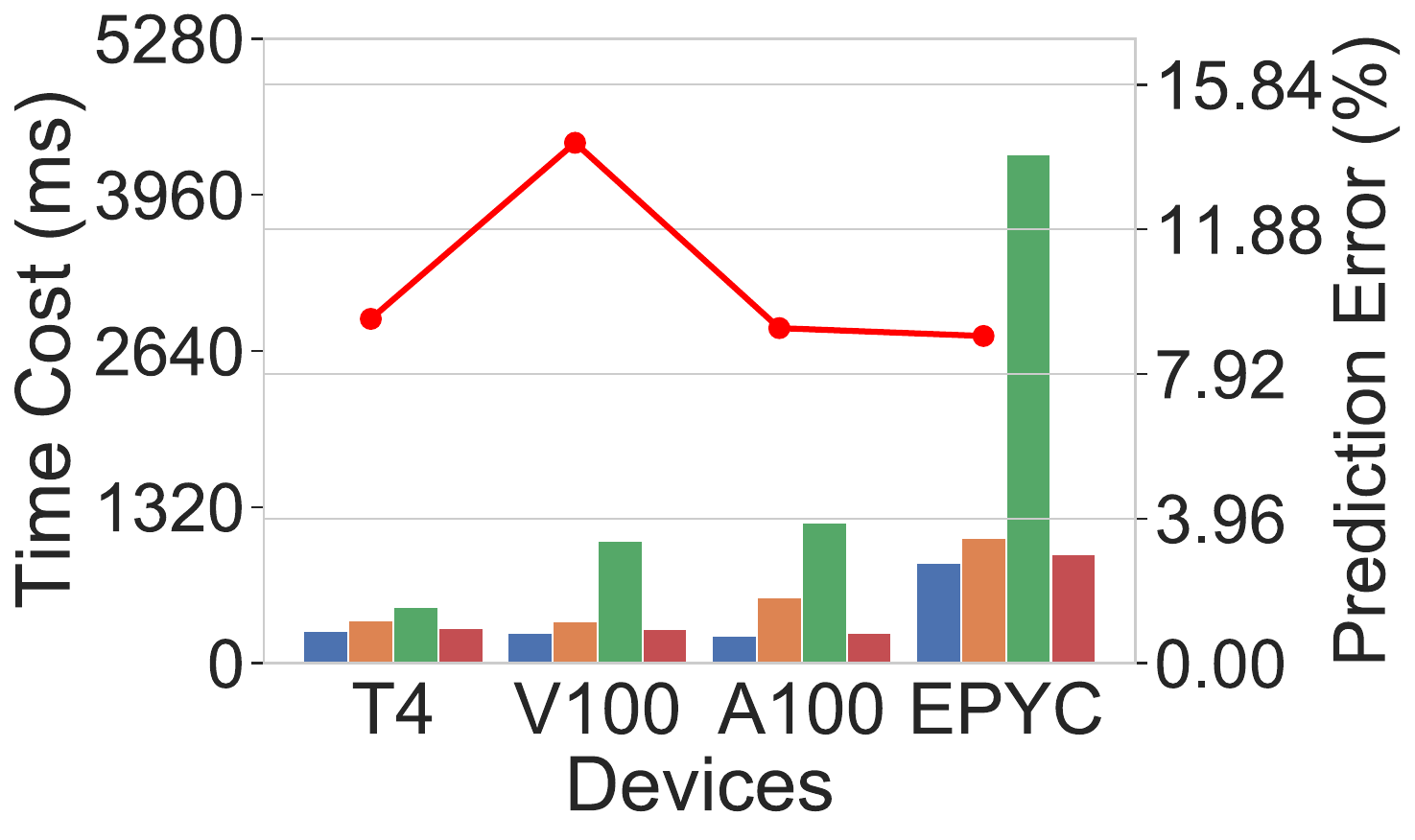}
}
\caption{End-to-end performance prediction for cross-model learning. 
% \Hu{Increase the font size}
}
\label{fig:end2end-cm-append}
\end{figure}

Fig.~\ref{fig:end2end-cm-append} evaluates the cross-model prediction error of end-to-end performance against the real performance on various devices. The results show that {\sysname} can accurately estimate the latency with an error at most $12.4\%$ and outperform all the baselines.

\subsection{Effect of distribution difference on model generalizability 
}
\begin{figure}[t]
\subfigure[]{     
\centering
\includegraphics[width=0.225\textwidth, trim=8 0 8 0, clip]{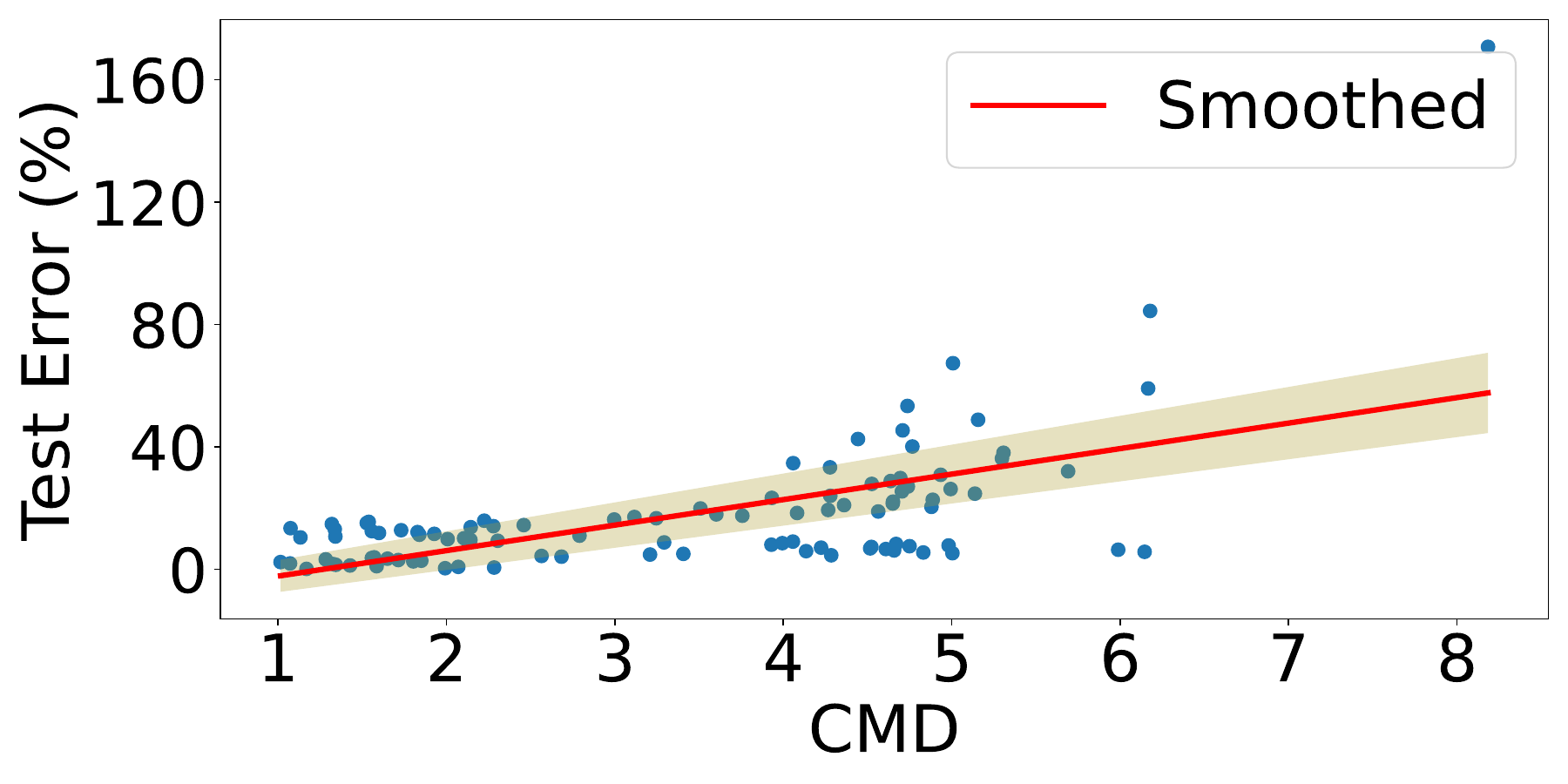}
\label{fig:cmd_effect_a-append}
}
\subfigure[]{
\centering
\includegraphics[width=0.225\textwidth, trim=8 0 8 0, clip]{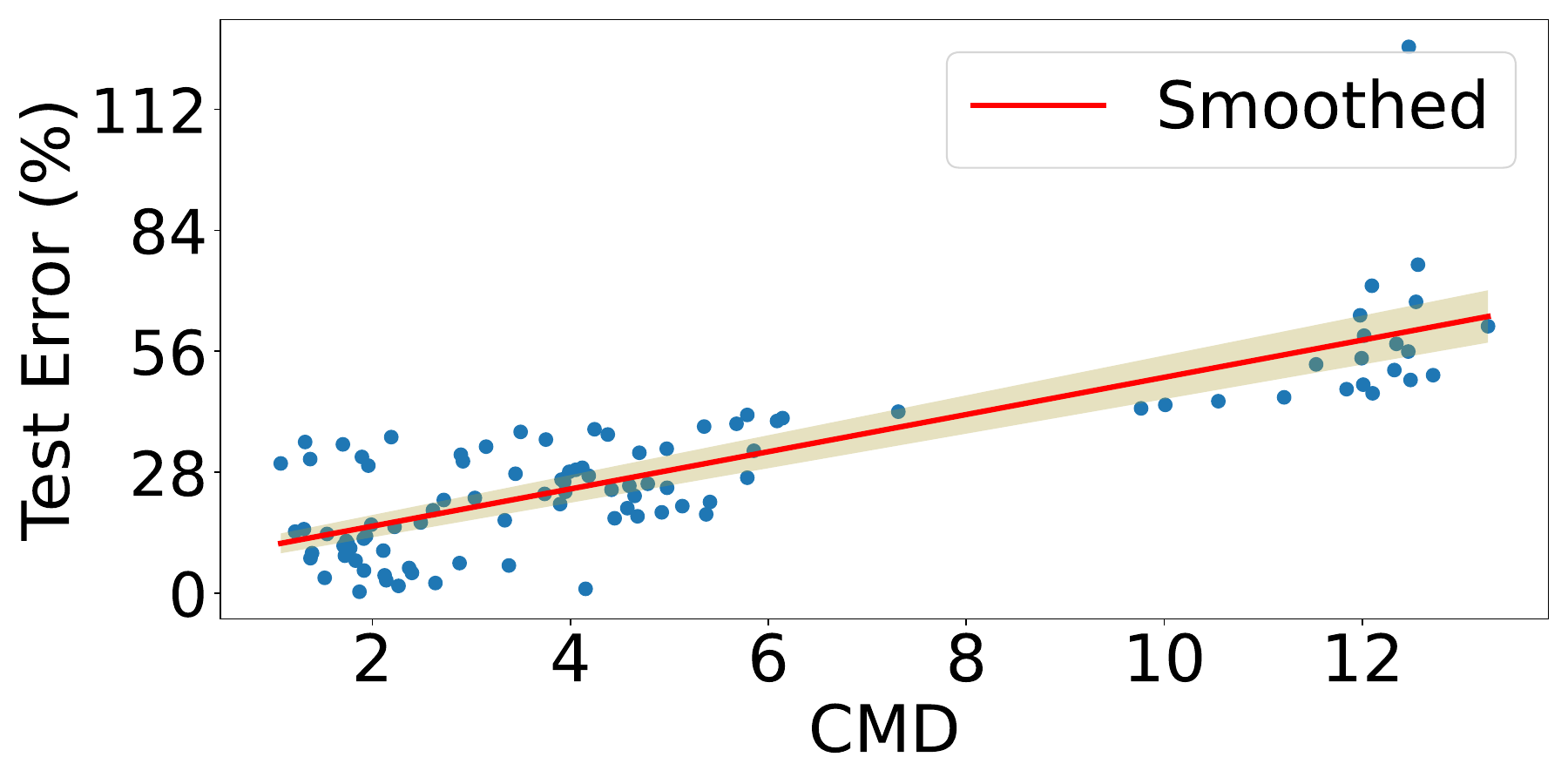}
\label{fig:cmd_effect_b-append}
}
\caption{\small Effect of distribution difference on model generalizability. x-axis: CMD between $X_{train}$ and $X_{test}$ subsets, y-axis: corresponding test error of the subset of $X_{test}$: a) Cross-model learning, where both $X_{train}$ and $X_{test}$ are collected from T4; b) Cross-device learning, where$X_{train}$ and $X_{test}$ are collected from different devices}
% from v100 to p100
\label{fig:cmd_effect-append}
\end{figure}

To illustrate the effects of distribution difference (measured as CMD) on prediction performance, we plot the CMD (x-axis) between two subsets that are randomly sampled from the training set and test set respectively, and the corresponding prediction error (y-axis), for both cross-model and cross-device training in Fig.~\ref{fig:cmd_effect-append}. %The results demonstrate that 
The performance of the predictor is positively related to prediction error, 
% \cwu{why? HHP: Intuitively, a cost model trained on a training set usually has a small error on the test set that is sampled from the same distribution as the training set.}, 
which indicates a predictor trained on one distribution can be well generalized to another distribution that has a small CMD difference from the training distribution. %This inspires us to minimize the CMD between the latent representations of source domains and that of target domains.

%% file: appendix/appendix_cluster_proof.tex
\section{Appendix: Proof of Eq.~\ref{eq:cluster_guarantee}} \label{appendix:cluster_guarantee}
In this appendix, we provide proof of Eq.~\ref{eq:cluster_guarantee}, namely,
\begin{equation}\begin{aligned}
     \mathcal{L}_{\mathrm{pre\_train}}(x) \leq \mathcal{L}_{\mathrm{pre\_train}}(c_x) + \mathcal{O}(\epsilon) .
\end{aligned}\end{equation}

\begin{proof}
    First, recall that $X$ is the feature space of all possible tensor programs and $x \in X$ is a given tensor program. $\mathcal{H}$ is the latent/embedding space, and we denote $h(x)$ as the latent representation of tensor $x$ in the latent space.  $C$ is our selected tensor program sample for fine-tuning.

    For the analysis, we make the standard assumption that the loss function $ \mathcal{L}_{\mathrm{pre\_train}}$ is $L_k$-Lipschitz continuous with respect to the representation in the latent space, namely, we have that 
    $$\|\mathcal{L}_{\mathrm{pre\_train}}(h(x))-\mathcal{L}_{\mathrm{pre\_train}}(h(x')) \| \leq L_k \Delta(h(x),h(x')).$$

    Next, recall that $c_x= \argmin_{c \in C}d_{\mathcal{H}}(h(x),h(c))$ is the closest $c \in C$ to $x$ in the latent space. Let's consider the difference between $x$ and $c_x$. First, note that if $\mathcal{L}_{\mathrm{pre\_train}}(x) > \mathcal{L}_{\mathrm{pre\_train}}(c_x)$,  Eq.~\ref{eq:cluster_guarantee} is automatically satisfied and no further derivation is needed, as 
    $$\epsilon \geq 0.$$
    
    Without loss of generality, we may assume that $\mathcal{L}_{\mathrm{pre\_train}}(x) > \mathcal{L}_{\mathrm{pre\_train}}(c_x)$.
    Then, combining the $L_k$-Lipschitz continuity of the loss function, we have,
    \begin{equation}\label{eq:analysis}
        \begin{split}
        \mathcal{L}_{\mathrm{pre\_train}}(x) & -\mathcal{L}_{\mathrm{pre\_train}}(c_x)  \\
           & =  \mathcal{L}_{\mathrm{pre\_train}}(h(x))-\mathcal{L}_{\mathrm{pre\_train}}(h(c_x))  \\
           & \leq L_k  \Delta(h(x),h(c_x))
        \end{split}
    \end{equation}
    Next, recall that $\epsilon := \max_{x \in X} \min_{c \in C}d_{\mathcal{H}}(h(x),h(c))$, which captures the maximal distance between any tensor program in the input space and the closest tensor program among the finetuning samples in the latent space.
    By the definition, we have that 
    $$\epsilon \geq  \Delta(h(x),h(c_x)), \quad \forall x, c_x $$
    Substitute this into Eq.~\ref{eq:analysis}, we have 
    \begin{equation}
        % \begin{split}
        \begin{aligned}
        \mathcal{L}_{\mathrm{pre\_train}}(x) & -\mathcal{L}_{\mathrm{pre\_train}}(c_x)  \\
           & \leq L_k  \Delta(h(x),h(c_x))
           & \leq L_k \epsilon
           & \leq \mathcal{O}(\epsilon)
        \end{aligned}
        % \end{split}
    \end{equation}
    Therefore, we arrive 
    $$\mathcal{L}_{\mathrm{pre\_train}}(x) \leq \mathcal{L}_{\mathrm{pre\_train}}(c_x) + \mathcal{O}(\epsilon) .$$
\end{proof}

%% file: appendix/EuroSys24_ArtifactAppendix_template.tex
%%%%%%%%%%%%%%%%%%%%%%%%%%%%%%%%%%%%%%%%%%%%%%%%%%%%
% Artifact Appendix Template for EuroSys'24 AE
%
% This document has a maximum length of 2 pages.
%%%%%%%%%%%%%%%%%%%%%%%%%%%%%%%%%%%%%%%%%%%%%%%%%%%%

\appendix
\section{Artifact Appendix}
% \textit{This artifact appendix is meant to be a self-contained document that describes a roadmap for the evaluation of your artifact. It should include a clear description of the hardware, software, and configuration requirements as well as the major claims made by your paper and how to reproduce each claim through your artifact. Linking the claims of your paper to the artifact is a necessary step that ultimately allows artifact evaluators to reproduce your results. Towards that end, you should explicitly list down items (e.g., results, plots, tables) from the paper and cross-reference those with the experiments to be reproduced with your artifact.}\\
% \textit{Please fill all the mandatory sections, keeping their titles and organization but removing the current illustrative content, and remove the optional sections \ref{sec:reuse} and \ref{sec:gnotes} where those do not apply to your artifact.}

%%%%%%%%%%%%%%%%%%%%%%%%%%%%%%%%%%%%%%%%%%%%%%%%%%%%%%%%%%%%%%%%%%%%%
\subsection{Abstract}
% {\em [Mandatory]} 
% {\em Provide a short description of your artifact.}
{\sysname} is a generic optimized framework for precise performance prediction of tensor programs across diverse DNN models and devices. We provide the source code of {\sysname} for artifact evaluation, which implements data preprocessing, cost model training, and model inference.

%%%%%%%%%%%%%%%%%%%%%%%%%%%%%%%%%%%%%%%%%%%%%%%%%%%%%%%%%%%%%%%%%%%%%
\subsection{Description \& Requirements}
% \textit{[Mandatory] This section should list all the information necessary to recreate the same experimental setup you have used to run your artifact. This includes at least a persistent link to a publicly accessible archival repository where all the artifact's main components (software, data sets, documentation, etc.) can be accessed and, where this applies, the minimal hardware and software requirements to run your artifact. It is also very good practice to list and describe in this section benchmarks where those are part of or simply have been used to produce results with, your artifact.}

\subsubsection{How to access}
% \textit{Describe here how to access your artifact. In the case of a public repository, you should provide a persistent link to it. In the case of a private repository, you should provide instructions on how to access it and where that access will be limited for the duration of this evaluation, that should be clearly indicated.\\}
% Note: This evaluation does not mandate the use of specific public repositories, so institutional repositories or open commercial repositories are acceptable. In any case, repositories used to archive the artifact should have a declared plan to enable permanent accessibility.

You can access the source code of {\sysname} at this \textbf{Github Repo}: \url{https://github.com/joapolarbear/cdmpp}.The dataset and source code is also available at DOI:\url{10.6084/m9.figshare.24156084}

\subsubsection{Hardware dependencies}
% \textit{[Simply write "None." where this does not apply to your artifact.]}
The current implementation of {\sysname} requires GPUs to run the cost model. 

\subsubsection{Software dependencies}
% \textit{[Simply write "None." where this does not apply to your artifact.]}
\begin{itemize}
\item customized TVM: \url{https://github.com/joapolarbear/tvm}
\item dPRO: \url{https://github.com/joapolarbear/dpro}
\item CUDA driver version >=450.80.02 (Linux) / 452.39 (Windows)
\end{itemize}

\subsubsection{Benchmarks} 
% \textit{Describe here any data (e.g., data sets, models, workloads, etc.) required by the experiments with this artifact reported in your paper.} \textit{[Simply write "None." where this does not apply to your artifact.]}
 
\textbf{Dataset: }
You can access the dataset through the following links: 
\begin{itemize}
\item $dataset\_cpu\_v3.3.zip$: \url{https://drive.google.com/file/d/1JQwGEe8jCpuhZPnUxO0Sb1CJJ06uevy6/view}
\item $dataset\_gpu\_v3.3.zip$: \url{https://drive.google.com/file/d/1jqHbmvXUrLPDCIqJIaPee_atsPc0ZFFK/view}
\end{itemize}

You can choose to use either the CPU part or the GPU part. Please follow the instructions in [Tenset Dataset](\url{https://github.com/tlc-pack/tenset/blob/main/docs/get_started_with_cost_model_experiments.md}) to download the dataset accordingly. Our profiled dataset for A100, V100 and P100 will be available at [figshare](\url{10.6084/m9.figshare.24156084})

%%%%%%%%%%%%%%%%%%%%%%%%%%%%%%%%%%%%%%%%%%%%%%%%%%%%%%%%%%%%%%%%%%
\subsection{Set-up}

% {\em [Mandatory]} \textit{This section should include all the installation and configuration steps required to prepare the environment to be used for the evaluation of your artifact.}

\subsubsection{Software Environment}
Pull the docker image
\begin{lstlisting}
docker pull haaanpeng/cdmpp:eurosys_ae
\end{lstlisting}

Launch the container
\begin{lstlisting}
docker run -it --runtime=nvidia --shm-size 32768m --name hphu-test haaanpeng/cdmpp:eurosys_ae /bin/bash
\end{lstlisting}

Download the source code and install dependencies.
\begin{lstlisting}
cd && git clone --recursive https://github.com/joapolarbear/cdmpp && cd cdmpp && bash setup.sh
\end{lstlisting}

\subsection{Prepare Dataset}

\subsubsection{Download and unzip}
You can choose to use either the CPU part or the GPU part. See [Tenset Dataset](\url{https://github.com/tlc-pack/tenset/blob/main/docs/get_started_with_cost_model_experiments.md}) to download the dataset accordingly. Our profiled dataset for A100, V100 and P100 will be available at [DOI:\url{10.6084/m9.figshare.24156084}](\url{https://figshare.com/articles/dataset/cdmpp-data/24156084})

\vspace{1mm}
\noindent\textbf{An example of T4 GPU. }
Here we show an example of downloading the dataset of NVIDIA T4.
\begin{enumerate}
    \item Change directory to \url{<cdmpp_root_directory>/3rdparty/tenset/scripts/}
    \item Download. You can download it from Google Drive with the link [$dataset\_gpu\_v3.3.zip$](\url{https://drive.google.com/file/d/1jqHbmvXUrLPDCIqJIaPee_atsPc0ZFFK/view?usp=sharing}). Or you can use the command line
    \begin{lstlisting}
    pip3 install gdown
    gdown https://drive.google.com/uc?
        id=1jqHbmvXUrLPDCIqJIaPee_atsPc0ZFFK
    \end{lstlisting}
    \item Unzip. Put \url{dataset_gpu_v3.3.zip} under \url{<cdmpp_root_directory>/3rdparty/tenset/scripts/} and run \url{unzip} \url{dataset_gpu_v3.3.zip}. A new folder \url{<dataset\_gpu>} will appear in \url{<cdmpp_root_directory>/3rdparty/tenset/scripts/}. Make \url{dataset} as a soft link to it by the following command
  \begin{lstlisting}
  ln -s <cdmpp_root_directory>/3rdparty/tenset/
    scripts/dataset_gpu dataset
  \end{lstlisting}
\end{enumerate}

\noindent\textbf{An example of AMD EPYC 7452 CPU. }
Here we show an example to download the dataset of AMD EPYC 7452 CPU.
\begin{enumerate}
    \item Change directory to \url{<cdmpp_root_directory>/3rdparty/tenset/scripts/}
    \item Download. You can download it from Google Drive with the link [\url{dataset_cpu_v3.3.zip}](\url{https://drive.google.com/file/d/1JQwGEe8jCpuhZPnUxO0Sb1CJJ06uevy6/view?usp=sharing}). Or you can use the command line
    \begin{lstlisting}
    pip3 install gdown
    gdown https://drive.google.com/uc?
        id=1JQwGEe8jCpuhZPnUxO0Sb1CJJ06uevy6
    \end{lstlisting}
    \item Unzip. Put \url{dataset_cpu_v3.3.zip} under \url{<cdmpp_root_directory>/3rdparty/tenset/scripts/} and run \url{unzip} \url{dataset_cpu_v3.3.zip}. A new folder \url{<dataset_cpu>} will appear in \url{<cdmpp_root_directory>/3rdparty/tenset/scripts/}. Make \url{dataset} as a soft link to it by the following command
  \begin{lstlisting}
  ln -s <cdmpp_root_directory>/3rdparty/tenset/
    scripts/dataset_cpu dataset
  \end{lstlisting}
\end{enumerate}
In the above process, if \textbf{dataset} already exists, just run 
  \begin{lstlisting}
  mv dataset_cpu/measure_records/* dataset/measure_records/
  \end{lstlisting}
After the above processes, you will see several directories under \url{<cdmpp_root_directory>/3rdparty/tenset/scripts/dataset/measure_records} as follows
\begin{lstlisting}
measure_records
  |-t4
  |-k80
\end{lstlisting}

Note that each directory name represents a specific device and we will use those device names as flags to specify which device we will use to extract features or run training.

\subsubsection{Feature Extraction.} After downloading the dataset and putting it on the right path, we will extract features for the dataset of each device. Make sure that you have put the profiled dataset under \url{3rdparty/tenset/scripts/dataset/measure_records/<DEVICE\_MODEL>}, where \url{<DEVICE\_MODEL>} is the device whose dataset you want to extract from. Then, you can run the following commands to extract features.
\begin{lstlisting}
cd && cd cdmpp
bash scripts/dataset/gen_raw_feature_all.sh
\end{lstlisting}
By default, the extracted features will be stored at \url{workspace/ast\_ansor/<DEVICE\_MODEL>}.
The process of extracting features and data preprocessing may take around $10\sim 20$ minutes for the dataset of each device.

\subsubsection{Data Preprocessing [Optional].} 
Run the following commands to preprocess the dataset
\begin{lstlisting}
bash scripts/dataset/make_dataset.sh
\end{lstlisting}
The preprocessed data will be stored under the \textit{tmp/} directory. You can also skip this process since this can be done automatically before training starts, i.e., when the preprocessed dataset is required to be used for the first time. This step takes around 5 minutes for the dataset of each device.

%%%%%%%%%%%%%%%%%%%%%%%%%%%%%%%%%%%%%%%%%%%%%%%%%%%%%%%%%%%%%%%%%%%%%
\subsection{Evaluation workflow}
% {\em [Mandatory]} \textit{This section should include all the operational steps and experiments which must be performed to evaluate your artifact is functional and to validate your paper's key results and claims. For that purpose, we ask you to use the two following subsections and cross-reference the items therein as explained next.}
We\footnote{Submission, reviewing, and badging methodology followed for the evaluation of this artifact can be found at \url{https://sysartifacts.github.io/eurosys2024/}.} mainly shows an example of the process to evaluate cross-model performance prediction, with the dataset collected from T4.

\subsubsection{Major Claims}
% \textit{Enumerate here the major claims (Cx) made in your paper. Follows an example:}\\

\begin{itemize}
    \item \textit{(C1): {\sysname} can achieve a prediction error around $19\%$ [refer to Fig.~\ref{fig:cm-finetune}-(a) in the paper].}
    % \item \textit{(C2): System\_name has been used to uncover new bugs in the Y software. This is proven by the experiments (E2) and (E3) in [ibid].}
\end{itemize}

\vspace{1mm}
\subsubsection{Experiments}
% \textit{Link explicitly the description of your experiments to the items you have provided in the previous subsection about Major Claims. We also highly encourage you to provide your estimates of human- and compute-time for each of the listed experiments. Follows an example:}
~\\

\textit{Experiment (E1): [CMPP] [5 human-minutes + $3\sim 5$ GPU compute-hour]: cross model cost model training on the dataset from T4}\\\\
\textit{[How to]}\\
% \textit{Describe thoroughly the steps to perform the experiment and to collect and organize the results as expected from your paper. We encourage you to use the following structure with three main blocks for the description of your experiment.} \\
Please follow the steps to perform this experiments
~\\
\textit{[Preparation]}
% \textit{Describe in this block the steps required to prepare and configure the environment for this experiment.}\\
We will use the configuration file \url{tmp/search\_trial\_20221119_1575.yaml}, which contains hyper-parameters found by our auto-tuner, to run the following experiments. You can also change the hyper-parameters in the configuration file according to your requirements.
~\\
\textit{[Execution]}
% \textit{Describe in this block the steps to run this experiment.}\\
Run the following commands
\begin{lstlisting}
    bash scripts/exp/cross_model.sh none
\end{lstlisting}
~\\
\textit{[Results]}
% \textit{Describe in this block the steps required to collect and interpret the results for this experiment.}\\
You will see training logs like this
\begin{lstlisting}
[2023-09-30 13:53:55] [base_learner.py:303] INFO - Time 1240.564 s - Epoch 126 step 27000 bs 600 - loss_train=17.410308837891, {'mape': 0.2110657768101716, 'rmse': 0.00045307391267025536, '20%accuracy': 0.6356699751861042, '10%accuracy': 0.3441997518610422, '5%accuracy': 0.17478287841191067} 
\end{lstlisting}

After training converges, the MAPE should be around 0.19, indicating $19\%$ test error.
~\\
% \textit{In all of the above blocks, we also recommend you to provide precise indications about the expected outcome for each of the steps.}

~\\

%% file: 0main.bbl
\begin{thebibliography}{10}

\bibitem{abadi2016tensorflow}
Mart{\'\i}n Abadi, Paul Barham, Jianmin Chen, Zhifeng Chen, Andy Davis, Jeffrey Dean, Matthieu Devin, Sanjay Ghemawat, Geoffrey Irving, Michael Isard, et~al.
\newblock {Tensorflow: A System for Large-scale Machine Learning}.
\newblock In {\em Proceedings of the 12th USENIX Symposium on Operating Systems Design and Implementation}, 2016.

\bibitem{adams2019learning}
Andrew Adams, Karima Ma, Luke Anderson, Riyadh Baghdadi, Tzu-Mao Li, Micha{\"e}l Gharbi, Benoit Steiner, Steven Johnson, Kayvon Fatahalian, Fr{\'e}do Durand, et~al.
\newblock Learning to optimize halide with tree search and random programs.
\newblock {\em ACM Transactions on Graphics (TOG)}, 2019.

\bibitem{optuna_2019}
Takuya Akiba, Shotaro Sano, Toshihiko Yanase, Takeru Ohta, and Masanori Koyama.
\newblock {Optuna: A Next-generation Hyperparameter Optimization Framework}.
\newblock In {\em Proceedings of the 25th {ACM} {SIGKDD} International Conference on Knowledge Discovery and Data Mining}, 2019.

\bibitem{amd2019epyc}
AMD.
\newblock {AMD EPYC}, 2019.
\newblock \url{https://www.amd.com/en/products/cpu/amd-epyc-7452}.

\bibitem{baghdadi2021deep_tiramisu}
Riyadh Baghdadi, Massinissa Merouani, Mohamed-Hicham Leghettas, Kamel Abdous, Taha Arbaoui, Karima Benatchba, et~al.
\newblock A deep learning based cost model for automatic code optimization.
\newblock In {\em Proceedings of Machine Learning and Systems}, 2021.

\bibitem{belkina2019automated}
Anna~C Belkina, Christopher~O Ciccolella, Rina Anno, Richard Halpert, Josef Spidlen, and Jennifer~E Snyder-Cappione.
\newblock {Automated optimized parameters for T-distributed stochastic neighbor embedding improve visualization and analysis of large datasets}.
\newblock {\em Nature communications}, 2019.

\bibitem{box1964analysis}
George~EP Box and David~R Cox.
\newblock An analysis of transformations.
\newblock {\em Journal of the Royal Statistical Society: Series B (Methodological)}, 1964.

\bibitem{sklearn_api}
Lars Buitinck, Gilles Louppe, Mathieu Blondel, Fabian Pedregosa, Andreas Mueller, Olivier Grisel, Vlad Niculae, Peter Prettenhofer, Alexandre Gramfort, Jaques Grobler, Robert Layton, Jake VanderPlas, Arnaud Joly, Brian Holt, and Ga{\"{e}}l Varoquaux.
\newblock {API design for machine learning software: experiences from the scikit-learn project}.
\newblock In {\em ECML PKDD Workshop: Languages for Data Mining and Machine Learning}, 2013.

\bibitem{cai2018proxylessnas}
Han Cai, Ligeng Zhu, and Song Han.
\newblock {Proxylessnas: Direct neural architecture search on target task and hardware}.
\newblock {\em arXiv preprint arXiv:1812.00332}, 2018.

\bibitem{chen2015xgboost}
Tianqi Chen, Tong He, Michael Benesty, Vadim Khotilovich, Yuan Tang, Hyunsu Cho, Kailong Chen, et~al.
\newblock Xgboost: extreme gradient boosting.
\newblock {\em R package version 0.4-2}, 2015.

\bibitem{chen2018tvm}
Tianqi Chen, Thierry Moreau, Ziheng Jiang, Lianmin Zheng, Eddie Yan, Haichen Shen, Meghan Cowan, Leyuan Wang, Yuwei Hu, Luis Ceze, et~al.
\newblock {TVM: An Automated End-to-End Optimizing Compiler for Deep Learning}.
\newblock In {\em Proceedings of the 13th USENIX Symposium on Operating Systems Design and Implementation}, 2018.

\bibitem{chen2018learning}
Tianqi Chen, Lianmin Zheng, Eddie Yan, Ziheng Jiang, Thierry Moreau, Luis Ceze, Carlos Guestrin, and Arvind Krishnamurthy.
\newblock Learning to optimize tensor programs.
\newblock In {\em Proceedings of Advances in Neural Information Processing Systems}, 2018.

\bibitem{nvidia2014K80}
NVIDIA Corporation.
\newblock {NVIDIA K80}, 2014.
\newblock \url{https://www.nvidia.com/en-gb/data-center/tesla-k80/}.

\bibitem{nvidia2016P100}
NVIDIA Corporation.
\newblock {NVIDIA P100}, 2016.
\newblock \url{https://www.nvidia.com/en-sg/data-center/tesla-p100/}.

\bibitem{nvidia2017V100}
NVIDIA Corporation.
\newblock {NVIDIA V100}, 2017.
\newblock \url{https://www.nvidia.com/en-gb/data-center/tesla-v100/}.

\bibitem{nvidia20182080Ti}
NVIDIA Corporation.
\newblock {NVIDIA 2080Ti}, 2018.
\newblock \url{https://www.nvidia.cn/geforce/graphics-cards/rtx-2080-ti/}.

\bibitem{nvidia2018T4}
NVIDIA Corporation.
\newblock {NVIDIA T4}, 2018.
\newblock \url{https://www.nvidia.com/en-in/data-center/tesla-t4/}.

\bibitem{nvidia2020A100}
NVIDIA Corporation.
\newblock {NVIDIA A100}, 2020.
\newblock \url{https://www.nvidia.com/en-sg/data-center/a100/}.

\bibitem{cuda2020benchmarking}
NVIDIA Corporation.
\newblock {NVIDIA Data Center Deep Learning Product Performance}, 2020.
\newblock \url{https://developer.nvidia.com/deep-learning-performance-training-inference}.

\bibitem{dai2019transformer}
Zihang Dai, Zhilin Yang, Yiming Yang, Jaime Carbonell, Quoc~V Le, and Ruslan Salakhutdinov.
\newblock {Transformer-xl: Attentive language models beyond a fixed-length context}.
\newblock {\em arXiv preprint arXiv:1901.02860}, 2019.

\bibitem{dasgupta2011probability}
Anirban DasGupta.
\newblock {\em Probability for statistics and machine learning: fundamentals and advanced topics}.
\newblock Springer, 2011.

\bibitem{devlin2018bert}
Jacob Devlin, Ming-Wei Chang, Kenton Lee, and Kristina Toutanova.
\newblock Bert: Pre-training of deep bidirectional transformers for language understanding.
\newblock {\em arXiv preprint arXiv:1810.04805}, 2018.

\bibitem{dudziak2020brp}
Lukasz Dudziak, Thomas Chau, Mohamed Abdelfattah, Royson Lee, Hyeji Kim, and Nicholas Lane.
\newblock {BRP-NAS: Prediction-based NAS using GCNS}.
\newblock In {\em Proceedings of Advances in Neural Information Processing Systems}, 2020.

\bibitem{amazon2010graviton}
Amazon EC2.
\newblock {AWS Graviton Processor}, 2019.
\newblock \url{https://aws.amazon.com/cn/ec2/graviton/}.

\bibitem{geoffrey2021habitat}
X~Yu Geoffrey, Yubo Gao, Pavel Golikov, and Gennady Pekhimenko.
\newblock {Habitat: A Runtime-Based Computational Performance Predictor for Deep Neural Network Training}.
\newblock In {\em Proceedings of the 2021 USENIX Annual Technical Conference}, 2021.

\bibitem{google2016TPU}
Google.
\newblock {Tensor Processing Unit}, 2016.
\newblock \url{https://cloud.google.com/tpu/docs/tpus}.

\bibitem{he2016resnet}
Kaiming He, Xiangyu Zhang, Shaoqing Ren, and Jian Sun.
\newblock {Deep Residual Learning for Image Recognition}.
\newblock In {\em Proceedings of the IEEE conference on Computer Vision and Pattern Recognition}, 2016.

\bibitem{hsieh2020non}
Kevin Hsieh, Amar Phanishayee, Onur Mutlu, and Phillip Gibbons.
\newblock The non-iid data quagmire of decentralized machine learning.
\newblock In {\em Proceedings of International Conference on Machine Learning}, 2020.

\bibitem{hu2022dpro}
Hanpeng Hu, Chenyu Jiang, Yuchen Zhong, Yanghua Peng, Chuan Wu, Yibo Zhu, Haibin Lin, and Chuanxiong Guo.
\newblock {dPRO: A Generic Profiling and Optimization System for Expediting Distributed DNN Training}.
\newblock In {\em Proceedings of Machine Learning and Systems}, 2022.

\bibitem{huawei2018shengteng}
HUAWEI.
\newblock {HUAWEI Ascend}, 2018.
\newblock \url{https://e.huawei.com/cn/products/servers/ascend}.

\bibitem{intel2016e5}
Intel.
\newblock {Intel® Xeon® Processor E5-2673 v4}, 2016.
\newblock \url{https://ark.intel.com/content/www/us/en/ark/products/series/91287/intel-xeon-processor-e5-v4-family.html}.

\bibitem{intel2019platinum}
Intel.
\newblock {Intel Platinum}, 2019.
\newblock \url{https://www.intel.sg/content/www/xa/en/products/details/processors/xeon/scalable/platinum.html}.

\bibitem{jia2019taso}
Zhihao Jia, Oded Padon, James Thomas, Todd Warszawski, Matei Zaharia, and Alex Aiken.
\newblock {TASO: Optimizing Deep Learning Computation with Automatic Generation of Graph Substitutions}.
\newblock In {\em Proceedings of the 27th ACM Symposium on Operating Systems Principles}, 2019.

\bibitem{jia2019optimizing}
Zhihao Jia, James Thomas, Todd Warszawski, Mingyu Gao, Matei Zaharia, and Alex Aiken.
\newblock {Optimizing DNN computation with relaxed graph substitutions}.
\newblock In {\em Proceedings of Machine Learning and Systems}, 2019.

\bibitem{jia2019beyond}
Zhihao Jia, Matei Zaharia, and Alex Aiken.
\newblock {Beyond Data and Model Parallelism for Deep Neural Networks}.
\newblock In {\em Proceedings of Machine Learning and Systems}, 2019.

\bibitem{abhinav2020domain}
Abhinav~Ramesh Kashyap, Devamanyu Hazarika, Min-Yen Kan, and Roger Zimmermann.
\newblock {Domain Divergences: a Survey and Empirical Analysis}, 2020.

\bibitem{kaufman2021learned}
Sam Kaufman, Phitchaya Phothilimthana, Yanqi Zhou, Charith Mendis, Sudip Roy, Amit Sabne, and Mike Burrows.
\newblock A learned performance model for tensor processing units.
\newblock In {\em Proceedings of Machine Learning and Systems}, 2021.

\bibitem{kaufman2019learned}
Samuel Kaufman, Phitchaya~Mangpo Phothilimthana, and Mike Burrows.
\newblock {Learned TPU cost model for XLA tensor programs}.
\newblock In {\em Proc. Workshop ML Syst. NeurIPS}, 2019.

\bibitem{kosaian2021boosting}
Jack Kosaian, Amar Phanishayee, Matthai Philipose, Debadeepta Dey, and Rashmi Vinayak.
\newblock Boosting the throughput and accelerator utilization of specialized cnn inference beyond increasing batch size.
\newblock In {\em Proceedings of International Conference on Machine Learning}, 2021.

\bibitem{habana2019goya}
Habana Labs.
\newblock {Habana Goya}, 2019.
\newblock \url{https://habana.ai/}.

\bibitem{lattner2020mlir}
Chris Lattner, Mehdi Amini, Uday Bondhugula, Albert Cohen, Andy Davis, Jacques Pienaar, River Riddle, Tatiana Shpeisman, Nicolas Vasilache, and Oleksandr Zinenko.
\newblock {MLIR: A compiler infrastructure for the end of Moore's law}.
\newblock {\em arXiv preprint arXiv:2002.11054}, 2020.

\bibitem{lin2020mcunet}
Ji~Lin, Wei-Ming Chen, Yujun Lin, Chuang Gan, Song Han, et~al.
\newblock {Mcunet: Tiny deep learning on iot devices}.
\newblock In {\em Proceedings of Advances in Neural Information Processing Systems}, 2020.

\bibitem{liu2022nnlqp}
Liang Liu, Mingzhu Shen, Ruihao Gong, Fengwei Yu, and Hailong Yang.
\newblock {NNLQP: A Multi-Platform Neural Network Latency Query and Prediction System with An Evolving Database}.
\newblock In {\em Proceedings of the 51st International Conference on Parallel Processing}, 2022.

\bibitem{long2016cmd}
Mingsheng Long, Han Zhu, Jianmin Wang, and Michael~I. Jordan.
\newblock {Deep Transfer Learning with Joint Adaptation Networks}, 2016.

\bibitem{mattson2020mlperf}
Peter Mattson, Christine Cheng, Gregory Diamos, Cody Coleman, Paulius Micikevicius, David Patterson, Hanlin Tang, Gu-Yeon Wei, Peter Bailis, Victor Bittorf, et~al.
\newblock Mlperf training benchmark.
\newblock In {\em Proceedings of Machine Learning and Systems}, 2020.

\bibitem{meltzer2019pinet}
Peter Meltzer, Marcelo Daniel~Gutierrez Mallea, and Peter~J Bentley.
\newblock {Pinet: A permutation invariant graph neural network for graph classification}.
\newblock {\em arXiv preprint arXiv:1905.03046}, 2019.

\bibitem{mlperf}
MLPerf.
\newblock {MLPerf}, 2020.
\newblock \url{https://mlcommons.org/}.

\bibitem{mohammadi2018deep}
Mehdi Mohammadi, Ala Al-Fuqaha, Sameh Sorour, and Mohsen Guizani.
\newblock {Deep learning for IoT big data and streaming analytics: A survey}.
\newblock {\em IEEE Communications Surveys \& Tutorials}, 2018.

\bibitem{cuda}
NVIDIA.
\newblock {CUDA Toolkit Release Notes}, 2020.
\newblock \url{https://docs.nvidia.com/cuda/archive/10.2/cuda-toolkit-release-notes/index.html}.

\bibitem{cudnn}
NVIDIA.
\newblock {cuDNN Documentation}, 2021.
\newblock \url{https://docs.nvidia.com/deeplearning/cudnn/developer-guide/index.html}.

\bibitem{paszke2017automatic}
Adam Paszke, Sam Gross, Soumith Chintala, Gregory Chanan, Edward Yang, Zachary DeVito, Zeming Lin, Alban Desmaison, Luca Antiga, and Adam Lerer.
\newblock {Automatic differentiation in PyTorch}.
\newblock In {\em NIPS-W}, 2017.

\bibitem{paszke2019pytorch}
Adam Paszke, Sam Gross, Francisco Massa, Adam Lerer, James Bradbury, Gregory Chanan, Trevor Killeen, Zeming Lin, Natalia Gimelshein, Luca Antiga, et~al.
\newblock {PyTorch: An Imperative Style, High-performance Deep Learning Library}.
\newblock In {\em Proceedings of Advances in Neural Information Processing Systems}, 2019.

\bibitem{petrini2022learning}
Leonardo Petrini, Francesco Cagnetta, Eric Vanden-Eijnden, and Matthieu Wyart.
\newblock Learning sparse features can lead to overfitting in neural networks.
\newblock {\em arXiv preprint arXiv:2206.12314}, 2022.

\bibitem{pham2018efficient}
Hieu Pham, Melody Guan, Barret Zoph, Quoc Le, and Jeff Dean.
\newblock Efficient neural architecture search via parameters sharing.
\newblock In {\em Proceedings of International Conference on Machine Learning}, 2018.

\bibitem{reddi2020mlperf}
Vijay~Janapa Reddi, Christine Cheng, David Kanter, Peter Mattson, Guenther Schmuelling, Carole-Jean Wu, Brian Anderson, Maximilien Breughe, Mark Charlebois, William Chou, et~al.
\newblock Mlperf inference benchmark.
\newblock In {\em 2020 ACM/IEEE 47th Annual International Symposium on Computer Architecture (ISCA)}, 2020.

\bibitem{redko2020survey}
Ievgen Redko, Emilie Morvant, Amaury Habrard, Marc Sebban, and Youn{\`e}s Bennani.
\newblock A survey on domain adaptation theory: learning bounds and theoretical guarantees.
\newblock {\em arXiv preprint arXiv:2004.11829}, 2020.

\bibitem{ryu2021metatune}
Jaehun Ryu and Hyojin Sung.
\newblock {Metatune: Meta-learning based cost model for fast and efficient auto-tuning frameworks}.
\newblock {\em arXiv preprint arXiv:2102.04199}, 2021.

\bibitem{simonyan2014very}
Karen Simonyan and Andrew Zisserman.
\newblock {Very Deep Convolutional Networks for Large-Scale Image Recognition}.
\newblock In {\em Proceedings of International Conference on Learning Representations}, 2015.

\bibitem{steiner2021value}
Benoit Steiner, Chris Cummins, Horace He, and Hugh Leather.
\newblock Value learning for throughput optimization of deep learning workloads.
\newblock In {\em Proceedings of Machine Learning and Systems}, 2021.

\bibitem{stojanov2021domain}
Petar Stojanov, Zijian Li, Mingming Gong, Ruichu Cai, Jaime~G. Carbonell, and Kun Zhang.
\newblock {Domain Adaptation with Invariant Representation Learning: What Transformations to Learn?}
\newblock In {\em Proceedings of Advances in Neural Information Processing Systems}, 2021.

\bibitem{tang2017enabling}
Jie Tang, Dawei Sun, Shaoshan Liu, and Jean-Luc Gaudiot.
\newblock {Enabling deep learning on IoT devices}.
\newblock {\em IEEE Computer}, 2017.

\bibitem{attention2017}
Ashish Vaswani, Noam Shazeer, Niki Parmar, Jakob Uszkoreit, Llion Jones, Aidan~N Gomez, \L~ukasz Kaiser, and Illia Polosukhin.
\newblock Attention is all you need.
\newblock In {\em Proceedings of Advances in Neural Information Processing Systems}, 2017.

\bibitem{wang2023margin}
Yidong Wang, Bowen Zhang, Wenxin Hou, Zhen Wu, Jindong Wang, and Takahiro Shinozaki.
\newblock Margin calibration for long-tailed visual recognition.
\newblock In {\em Proceedings of Asian Conference on Machine Learning}, 2023.

\bibitem{weisberg2001yeo}
Sanford Weisberg.
\newblock {Yeo-Johnson power transformations}.
\newblock {\em Department of Applied Statistics, University of Minnesota. Retrieved June}, 2001.

\bibitem{williams2009roofline}
Samuel Williams, Andrew Waterman, and David Patterson.
\newblock Roofline: an insightful visual performance model for multicore architectures.
\newblock {\em Communications of the ACM}, 2009.

\bibitem{you2019fast_tpu}
Yang You, Zhao Zhang, Cho-Jui Hsieh, James Demmel, and Kurt Keutzer.
\newblock {Fast deep neural network training on distributed systems and cloud TPUs}.
\newblock {\em IEEE Transactions on Parallel and Distributed Systems}, 2019.

\bibitem{zellinger2019cmd}
Werner Zellinger, Bernhard~A Moser, Thomas Grubinger, Edwin Lughofer, Thomas Natschl{\"a}ger, and Susanne Saminger-Platz.
\newblock Robust unsupervised domain adaptation for neural networks via moment alignment.
\newblock {\em Information Sciences}, 2019.

\bibitem{zhai2023tlp}
Yi~Zhai, Yu~Zhang, Shuo Liu, Xiaomeng Chu, Jie Peng, Jianmin Ji, and Yanyong Zhang.
\newblock {TLP: A Deep Learning-Based Cost Model for Tensor Program Tuning}.
\newblock In {\em Proceedings of the 28th ACM International Conference on Architectural Support for Programming Languages and Operating Systems}, 2023.

\bibitem{zhang2021nn_Meter}
Li~Lyna Zhang, Shihao Han, Jianyu Wei, Ningxin Zheng, Ting Cao, Yuqing Yang, and Yunxin Liu.
\newblock {nn-Meter: towards accurate latency prediction of deep-learning model inference on diverse edge devices}.
\newblock In {\em Proceedings of the 19th Annual International Conference on Mobile Systems, Applications, and Services}, 2021.

\bibitem{zhao2022moses}
Zhihe Zhao, Xian Shuai, Yang Bai, Neiwen Ling, Nan Guan, Zhenyu Yan, and Guoliang Xing.
\newblock {Moses: Efficient Exploitation of Cross-device Transferable Features for Tensor Program Optimization}.
\newblock {\em arXiv preprint arXiv:2201.05752}, 2022.

\bibitem{zheng2020ansor}
Lianmin Zheng, Chengfan Jia, Minmin Sun, Zhao Wu, Cody~Hao Yu, Ameer Haj-Ali, Yida Wang, Jun Yang, Danyang Zhuo, Koushik Sen, et~al.
\newblock {Ansor: Generating High-Performance Tensor Programs for Deep Learning}.
\newblock In {\em Proceedings of the 14th USENIX Symposium on Operating Systems Design and Implementation}, 2020.

\bibitem{zheng2021tenset}
Lianmin Zheng, Ruochen Liu, Junru Shao, Tianqi Chen, Joseph~E Gonzalez, Ion Stoica, and Ameer~Haj Ali.
\newblock {Tenset: A large-scale program performance dataset for learned tensor compilers}.
\newblock In {\em Proceedings of the Thirty-fifth Conference on Neural Information Processing Systems Datasets and Benchmarks Track}, 2021.

\bibitem{zhu2018benchmarking}
Hongyu Zhu, Mohamed Akrout, Bojian Zheng, Andrew Pelegris, Anand Jayarajan, Amar Phanishayee, Bianca Schroeder, and Gennady Pekhimenko.
\newblock Benchmarking and analyzing deep neural network training.
\newblock In {\em 2018 IEEE International Symposium on Workload Characterization (IISWC)}, 2018.

\bibitem{qi2021gnncmd}
Qi~Zhu, Natalia Ponomareva, Jiawei Han, and Bryan Perozzi.
\newblock {Shift-Robust GNNs: Overcoming the Limitations of Localized Graph Training Data}, 2021.

\end{thebibliography}
